\newif\ifanonymous
\anonymoustrue
\newif\ifnamedauthors
\namedauthorstrue
\newif\ifreport
\reporttrue
\documentclass[11pt,logo,onecolumn,copyright]{nvidiatechreport}

\usepackage{microtype}
\microtypesetup{tracking=false}
\usepackage{graphicx}
\usepackage{booktabs}
\usepackage{multirow}
\usepackage{array}
\usepackage{amsmath,amssymb,amsthm}
\usepackage{xcolor}
\usepackage{colortbl}
\usepackage{tikz}
\usetikzlibrary{arrows.meta,positioning,fit,calc,backgrounds}
\usepackage{hyperref}
\usepackage{xspace}

\usepackage{algorithm,algorithmic}
\usepackage{natbib}
\setcitestyle{authoryear,round,citesep={;},aysep={,},yysep={;}}
\renewcommand{\cite}[1]{\citep{#1}}

\hypersetup{pdftitle={Shallow Encoding Emits Deep Memories},pdfsubject={Shallow Encoding Emits Deep Memories}}
\title{Shallow Encoding Emits Deep Memories}
\definecolor{seedleaf}{HTML}{76B900}
\definecolor{seedleafdark}{HTML}{4E7A00}
\definecolor{seedbrown}{HTML}{8A5A2B}
\newcommand{\seedicon}[1][1]{\begin{tikzpicture}[x=#1em,y=#1em,baseline=0.08em]
  \fill[seedbrown] (0.5,0.22) ellipse (0.30 and 0.20);
  \fill[white,opacity=.35] (0.42,0.27) ellipse (0.09 and 0.05);
  \draw[seedleafdark,line width=0.07em,line cap=round] (0.5,0.38) .. controls (0.5,0.58) .. (0.52,0.74);
  \fill[seedleaf] (0.52,0.72) .. controls (0.68,0.98) and (0.98,0.92) .. (1.00,0.80) .. controls (0.86,0.66) and (0.64,0.64) .. (0.52,0.72) -- cycle;
  \fill[seedleafdark] (0.52,0.72) .. controls (0.38,0.96) and (0.08,0.90) .. (0.04,0.78) .. controls (0.18,0.64) and (0.40,0.62) .. (0.52,0.72) -- cycle;
\end{tikzpicture}}
\newcommand{\projectpage}{https://latent-voyager-seed.static.hf.space}
\ifnamedauthors
\author{Mingyuan Ma\textsuperscript{*}, Yangmin Li\textsuperscript{*}, Harris Nover, Zheyu Fu, Po-Han Huang, Alex~Settle, Ashwin~Nanjappa\textsuperscript{\dag}\\{\Affilfont NVIDIA}\\{\rule{0pt}{1.45em}\normalfont\fontsize{9}{11}\selectfont\href{\projectpage}{\seedicon[1.25]\ \textcolor{seedleafdark}{\bfseries Project page}}}}
\correspondingauthor{Ashwin Nanjappa\textsuperscript{\dag}. \textsuperscript{*}Equal contribution: Mingyuan Ma and Yangmin Li co-led the project, and their order was decided at random.}
\hypersetup{pdfauthor={Mingyuan Ma, Yangmin Li, Harris Nover, Zheyu Fu, Po-Han Huang, Alex Settle, Ashwin Nanjappa}}
\else
\author{Anonymous Authors}
\affil{Anonymous Institution}
\correspondingauthor{Anonymous Authors (\texttt{anon@anon.org})}
\hypersetup{pdfauthor={Anonymous Authors}}
\fi
\date{2026-11-03}
\renewcommand{\today}{2026-11-03}
\keywords{LLM inference, hybrid models, linear attention, recurrent state, prefill, algorithm--system co-design}
\hypersetup{pdfkeywords={LLM inference, hybrid models, linear attention, recurrent state, prefill, algorithm--system co-design}}

\newcommand{\model}{Qwen3.8-Flash-Next\xspace}
\newcommand{\lightning}{Nemotron-3.5-Lightning\xspace}
\newcommand{\kimi}{Kimi-Linear\xspace}
\newcommand{\glm}{GLM-5.3-Flash\xspace}
\newcommand{\super}{Nemotron-3.5-Super\xspace}
\newcommand{\modelnv}{Qwen3.8-Flash-Next-NVFP4\xspace}
\newcommand{\modelFull}{Qwen3.8-Flash-Next\xspace}\newcommand{\modelnvFull}{Qwen3.8-Flash-Next-NVFP4\xspace}\newcommand{\lightningFull}{Nemotron-3.5-Lightning\xspace}\newcommand{\superFull}{Nemotron-3.5-Super\xspace}
\newcommand{\duet}{SEED\xspace}   %

\newcommand{\vbar}{\bar v}

\newcommand{\Rr}{\mathbb{R}}

\newcommand{\rms}{\mathrm{RMSNorm}}

\newcommand{\fnbfpending}[1]{\ifmmode\text{\color{red}#1}\else{\color{red}#1}\fi}
\newcommand{\pendingcell}[1]{\ifmmode\text{\color{red}#1}\else{\color{red}#1}\fi}

\providecommand{\anonph}[1]{\ifnamedauthors[#1 at release]\else[anonymized #1]\fi}
\ifanonymous

\newcommand{\hubRepo}[1]{\texttt{\anonph{Hub repository}}}

\newcommand{\relDir}[1]{[release directory]}
\newcommand{\arcDir}[1]{[release directory]}
\newcommand{\relName}[1]{[release name]}
\newcommand{\runName}[1]{[run]}
\newcommand{\specName}[1]{[specification]}

\newcommand{\commitid}[1]{[hash]}

\else
\newcommand{\hubRepo}[1]{\begingroup\def\UrlBreaks{\do\/\do\-}\nolinkurl{#1}\endgroup}
\newcommand{\commitid}[1]{#1}

\newcommand{\relDir}[1]{release/#1}
\newcommand{\arcDir}[1]{release/archive/#1}
\newcommand{\relName}[1]{#1}
\newcommand{\runName}[1]{#1}
\newcommand{\specName}[1]{#1}

\fi
\newtheorem{proposition}{Proposition}
\newtheorem{lemma}[proposition]{Lemma}
\newtheorem{corollary}[proposition]{Corollary}
\definecolor{ok}{RGB}{0,110,60}
\definecolor{bad}{RGB}{170,30,30}
\definecolor{boxfill}{RGB}{245,245,245}
\definecolor{nvgreen}{HTML}{76B900}
\definecolor{nvdark}{HTML}{1A1A1A}
\definecolor{nvgrey}{HTML}{5E5E5E}
\definecolor{nvlight}{HTML}{B5E36B}
\definecolor{nvpale}{HTML}{E6E6E6}

\renewcommand{\dbltopfraction}{0.9}
\renewcommand{\topfraction}{0.9}
\renewcommand{\textfraction}{0.07}
\ifreport\else\fi
\ifreport\newcommand{\colfit}[1]{#1}\else\newcommand{\colfit}[1]{\resizebox{\columnwidth}{!}{#1}}\fi
\ifreport\newcommand{\mainfigwidth}{\textwidth}\else\newcommand{\mainfigwidth}{0.94\textwidth}\fi
\newsavebox{\fitboxbox}
\ifreport
\makeatletter\renewcommand{\scriptsize}{\@setfontsize\scriptsize\@viipt\@viiipt}\makeatother
\ifdefined{font=small,skip=6pt}\else
\makeatletter\long\def\@makecaption#1#2{\vskip\abovecaptionskip{\small#1: #2\par}\vskip\belowcaptionskip}\makeatother\fi
\else
\usepackage[T1]{fontenc}
\usepackage{caption}
\fi
\let\fitboxtabular\tabular \let\endfitboxtabular\endtabular
\renewenvironment{tabular}{\begin{lrbox}{\fitboxbox}\fitboxtabular}%
  {\endfitboxtabular\end{lrbox}\ifdim\wd\fitboxbox>\linewidth\resizebox{\linewidth}{!}{\usebox{\fitboxbox}}\else\usebox{\fitboxbox}\fi}

\usepackage{placeins}
\usepackage{cuted}
\usepackage{etoolbox}
\DeclareCaptionFont{appendixcompact}{\fontsize{11}{12}\selectfont}

\newcommand{\stateOrigMB}{113.2}
\newcommand{\stateNewMB}{29.2}

\newcommand{\ratioStateMin}{2.0}
\newcommand{\ratioStateMax}{3.9}
\newcommand{\ratioStateMeanMin}{1.3}
\newcommand{\ratioStateMeanMax}{2.6}

\newcommand{\lstateOrigMB}{48.2}
\newcommand{\lstateNewMB}{18.8}

\newcommand{\kstateOrigMB}{41.9}
\newcommand{\kstateNewMB}{21.3}

\newcommand{\gstateOrigMB}{142.6}
\newcommand{\gstateNewMB}{72.4}
\newcommand{\gdeepB}{4,228}

\newcommand{\gdsaB}{1,057}

\newcommand{\prefillOrig}{12.1}%
\newcommand{\prefillNew}{9.1}%

\newcommand{\prefillLayerPctMin}{63}
\newcommand{\prefillLayerPctMax}{89}%
\newcommand{\prefillFlopPctMin}{71}

\newcommand{\latentKB}{5.4}
\newcommand{\deepKB}{10.6}

\newcommand{\fnR}{16}
\newcommand{\fnW}{16}
\newcommand{\fnCode}{8192{+}256}
\newcommand{\fnRank}{8192}
\newcommand{\fnSpikes}{256}

\newcommand{\trainableKM}{0.11B}
\newcommand{\trainableGL}{0.71B}

\newcommand{\trainMinutes}{160}
\newcommand{\trainNodes}{16}

\newcommand{\trainTwoMinutes}{111}

\newcommand{\examFNMATHPairTwo}{$-0.53\pm0.15$}

\newcommand{\examFNIFEvalPair}{$-0.6\pm0.3$}

\newcommand{\examFNReadMathPair}{$-0.33\pm0.11$}

\newcommand{\examFNReadIFPair}{$-0.18\pm0.27$}

\newcommand{\examNVMATHPairTwo}{$-0.30\pm0.16$}

\newcommand{\examNVMATHNormPairTwo}{$-0.20\pm0.14$}

\newcommand{\examLNLCBPair}{$-0.6\pm0.6$}

\newcommand{\examLNRulerSixtyFourPair}{$-0.3\pm0.2$}

\newcommand{\examBelowOneSEN}{19}%

\newcommand{\examBelowTwoSEN}{3}

\newcommand{\examCountWord}{eleven}
\newcommand{\examCellsTotal}{66}%
\newcommand{\examWithinTwoSEN}{63}%

\newcommand{\heldKMVThreeAll}{2.6}

\newcommand{\heldKMPrompts}{820}
\newcommand{\heldKMVThreeFiveAll}{2.3}

\newcommand{\heldNVVThreeAll}{10.8}

\newcommand{\heldNVPrompts}{1{,}013}
\newcommand{\heldNVVThreeFiveAll}{9.9}

\newcommand{\ruleFNGapK}{0.074}

\newcommand{\ruleFNTemplateErr}{0.180}
\newcommand{\ruleFNContentErr}{0.153}

\newcommand{\ruleLNGapK}{0.075}

\newcommand{\ruleKMGapK}{0.031}

\newcommand{\ruleKMBudget}{0.0077}

\newcommand{\ruleKMDriftSixteen}{0.0096}

\newcommand{\ruleGLPrevGap}{0.045}

\newcommand{\ruleGLBudget}{0.0087}

\newcommand{\ruleGLTemplateErr}{0.488}
\newcommand{\ruleGLContentErr}{0.384}

\newcommand{\ruleGLCutA}{269}
\newcommand{\ruleGLCutB}{530}
\newcommand{\ruleGLBase}{595}

\newcommand{\ruleGLPTwo}{528}
\newcommand{\ruleGLPOne}{124}

\newcommand{\ruleGLPOneP}{3\times10^{-118}}

\newcommand{\ruleGLFloor}{0.018}
\newcommand{\ruleGLCopyCapMax}{2}

\newcommand{\ruleFNFourKCost}{+0.0103}
\newcommand{\ruleFNFourKVar}{0.853}
\newcommand{\ruleFNFourKLost}{227}
\newcommand{\ruleFNCutCopies}{290}

\newcommand{\ruleGLCopyN}{600}

\newcommand{\ruleAddLo}{0.85}
\newcommand{\ruleAddHi}{0.88}
\newcommand{\ruleAddN}{4}

\newcommand{\ruleFNBeyond}{0.063}

\newcommand{\ruleKMBeyond}{0.065}

\newcommand{\ruleGLBeyond}{0.057}
\newcommand{\ruleKMBeyondTT}{0.019}%
\newcommand{\ruleGLBeyondTT}{0.015}%

\newcommand{\fairLNLCBWall}{$-0.22\pm0.52$}

\newcommand{\fairLNRulerQA}{$-0.09\pm0.18$}

\newcommand{\shareOwnFN}{59}

\newcommand{\shareOwnLN}{68}
\newcommand{\shareOwnKM}{37}
\newcommand{\shareOwnGL}{58}
\newcommand{\shareOwnSU}{68}%

\newcommand{\mainKimiDepth}{19}
\newcommand{\mainStateNewMB}{\stateNewMB}
\newcommand{\mainLstateNewMB}{\lstateNewMB}
\newcommand{\mainKstateNewMB}{\kstateNewMB}

\newcommand{\mainPrefillOrig}{\prefillOrig}
\newcommand{\mainPrefillNew}{\prefillNew}

\newcommand{\mainMemoryLabel}{tab:bytes}
\newcommand{\appendixMemoryTable}{}

\newcommand{\sinkPooledMinPct}{69}
\newcommand{\sinkPooledMaxPct}{99.7}
\newcommand{\sinkSUMedPct}{70}
\newcommand{\sinkKMMedPct}{34}
\newcommand{\sinkGLMedPct}{24}
\newcommand{\sinkKMPoolPct}{82}
\newcommand{\sinkGLPoolPct}{86}
\newcommand{\sinkKMValPct}{16}
\newcommand{\sinkGLValPct}{13}
\newcommand{\sinkFNValPct}{81}
\newcommand{\sinkDomainMaxChange}{0.05}
\newcommand{\skipSUCheaper}{18}
\newcommand{\skipSUSampled}{22}
\newcommand{\skipSURatioAbove}{6.5--14}
\newcommand{\skipGLFromLayer}{24}

\newcommand{\skipGLSampledFrom}{7}
\newcommand{\skipGLRatioAbove}{11--19}
\newcommand{\skipLNSampled}{17}
\newcommand{\skipLNCheaper}{16}

\newcommand{\skipLNRatioAbove}{5.5--38}

\newcommand{\lensSUKL}{6.8}

\newcommand{\clGrowthMin}{1.2}
\newcommand{\clGrowthMax}{1.4}

\newcommand{\clComprMin}{512}
\newcommand{\clComprMax}{1{,}024}
\newcommand{\clStepsMax}{16K}
\newcommand{\precFPSixteenMaxChange}{0.0006}
\newcommand{\precGainMin}{2.0}
\newcommand{\precGainMax}{3.9}

\newcommand{\precFPEightGainOthersMin}{1.8}
\newcommand{\precFPEightGainOthersMax}{4.7}
\newcommand{\precFPEightDenseMax}{0.075}

\newcommand{\precPplSeedMax}{0.6}
\newcommand{\precSeedDriftMin}{0.002}
\newcommand{\precSeedDriftMax}{0.007}
\newcommand{\nvRulerRatioMin}{1.9}
\newcommand{\nvRulerRatioMax}{3.5}

\newcommand{\nvCheckMaxDiff}{0.001}
\newcommand{\nvPplMin}{1.2}
\newcommand{\nvPplMax}{2.1}

\newcommand{\nvSUEffectOurs}{0.012}
\newcommand{\nvSUEffectSgl}{0.010}
\newcommand{\nvRulerKL}{0.0105}

\newcommand{\flipSeedPct}{7.4}
\newcommand{\flipOrigPct}{7.3}
\newcommand{\flipCells}{36}

\newcommand{\lenRatio}{0.94}
\newcommand{\lenCells}{26}
\newcommand{\lenPairs}{171{,}317}
\newcommand{\lenShorter}{14}
\newcommand{\lenLonger}{2}
\newcommand{\lenCappedOrig}{338}
\newcommand{\lenCappedSeed}{286}
\newcommand{\lenThinkOrig}{353}
\newcommand{\lenThinkSeed}{300}
\newcommand{\curveFracMinPct}{89}
\newcommand{\curveFracTokens}{150M}
\newcommand{\relServedMin}{0.006}
\newcommand{\relServedMax}{0.039}
\newcommand{\relExactMin}{0.004}
\newcommand{\relExactMax}{0.032}
\newcommand{\relBypassMinPct}{13}
\newcommand{\relBypassSets}{5}

\newcommand{\relCacheBeyondMin}{2}
\newcommand{\relCacheBeyondMax}{79}

\begin{document}

\begin{abstract}
Hybrid language models interleave a few softmax-attention layers with many linear-attention or state-space layers, so every layer keeps a memory of earlier tokens: a key--value cache or a fixed-size recurrent state.
Serving them spends more compute and high-bandwidth memory (HBM) on these memories than the output requires. The prefill runs every layer for every prompt token, although only the last token's prediction is used, and every decode step loads and stores whole recurrent states, which on \model take as many bytes per request as the attention cache of thousands of tokens.
We find two properties of trained hybrids that locate this excess.
First, writes are easier than reads: writing a deep layer's memory of a token needs little of the deep computation that predicting the next token needs. Trained copies of the deep layers' write paths, which we call \emph{emitters}, therefore write a prompt's deep memories from the residual stream of a shallower \emph{anchor layer}, and all prompt tokens but the last run as few as \prefillLayerPctMin\% of the layers.
Second, recurrent states have a \emph{state sink}: most of their energy lies in a rank-one term along a fixed direction of each head, and the rest is a low-rank working memory.
\duet (Shallow Encoding Emits Deep Memories) builds on both: it makes the prefill shallow in depth and the decode state shallow in representation. Each recurrent state is stored as its sink plus low-rank factors, which the Gated DeltaNet, Kimi Delta Attention and Mamba-2 updates extend exactly; we prove that the open-loop error of the periodic compressions back to low rank is bounded independently of their number.
\duet keeps the original weights frozen, compresses no attention cache, and runs all layers for every generated token.
On six checkpoints of five production hybrids of 30B to over 300B parameters, it is nearly lossless on \examCountWord{} official exams. Served with SGLang on NVIDIA GB300 GPUs on an agentic-coding trace, it raises the per-GPU throughput of \model 1.2--1.4$\times$.
Neither property was designed into these models.

\end{abstract}

\maketitle

\section{Introduction}
\label{sec:intro}

Hybrid language models keep a few softmax-attention layers and make the other token-mixing layers recurrent~\cite{mamba2_2024,gateddeltanet2025,kimilinear2025}, and production models now take this form~\cite{qwen3next2025,nvidia2025nemotronh,minimax2025minimax01}.
Every layer keeps a \emph{memory} of earlier tokens. An attention layer keeps a key--value (KV) cache, one key and one value per token. A linear-attention or state-space layer folds each token into a fixed-size \emph{recurrent state}, one $d_k\times d_v$ matrix per head.
Serving runs in two phases with different bottlenecks~\cite{splitwise2024,distserve2024,mooncake2025}. The \emph{prefill} processes all $N$ prompt tokens together and is compute-bound. The \emph{decode} generates one token per request per step; each step reads the weights and the memories of every resident request, so it is bound by high-bandwidth memory (HBM) bandwidth and by how many requests fit next to the weights (\S\ref{sec:setting}).

Both phases spend more compute and HBM on these memories than the output requires.
In the prefill, only the last prompt token predicts the first reply token; the other $N-1$ run all $L$ layers of the model only to write their memories, and agentic workloads make these prompts long~\cite{deepseekv41}.
In the decode, every recurrent head loads its whole state from HBM and stores it back at every step, and fixed-size does not mean small: on \model, the states of one request take \stateOrigMB\,MB, as much as the attention cache of about 4.5K tokens (Table~\ref{\mainMemoryLabel}).

We show that this excess is structural. Trained hybrids have two properties, one per phase, that make their memories far simpler than the computation and storage that serve them. We find both in all five production models we study, \model~\cite{qiu2026qwen38next}, \lightning~\cite{nemotron3_2026}, \kimi~\cite{kimilinear2025}, \glm~\cite{zai2026glm53flash,glm5team2026glm5} and \super~\cite{nemotron35super_2026}, whose recurrent layers are Gated DeltaNet, Kimi Delta Attention and Mamba-2.

\begin{figure*}[t]
\centering
\includegraphics[width=\mainfigwidth]{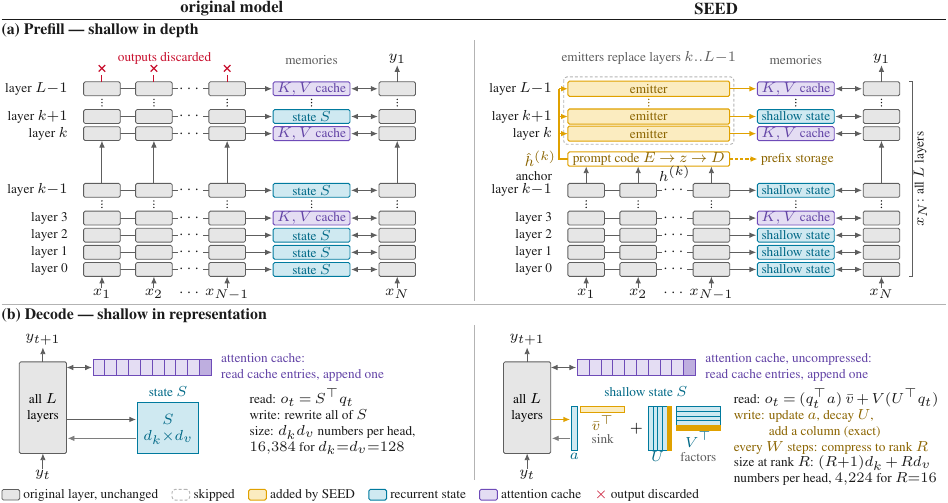}
\caption{\duet keeps a shallow encoding of a hybrid model's memories in each phase. Left: original model; right: \duet{}; layer stacks as in \model, with three recurrent layers per attention layer. (a)~Prefill, shallow in depth: $x_1..x_{N-1}$ run layers $0..k{-}1$ only. Emitters, trained copies of the deep layers' write paths, write every deeper memory from $h^{(k)}$, the output of anchor layer $k{-}1$, directly or through an optional prompt code. At the end of the prefill, the states left by $x_1..x_{N-1}$ are compressed to the shallow form of~(b); $x_N$ runs all $L$ layers. (b)~Decode, shallow in representation: each stored recurrent state is its sink $a\vbar^{\top}$ plus rank-$R$ factors $UV^{\top}$. A step updates the sink's coefficient $a$, decays $U$ and appends one column to each factor, which keeps the update exact; every $W$ steps, the factors are compressed back to rank $R$. Every generated token runs all $L$ layers, and no attention cache is compressed.}
\label{fig:arch}
\end{figure*}

\begin{table}[t]
\centering\footnotesize
\setlength{\tabcolsep}{5pt}
\renewcommand{\arraystretch}{1.35}
\caption{\duet in the two serving phases. In each phase, a property of trained hybrids lets \duet keep an encoding of the memories that is shallow along the phase's bottleneck, compute in the prefill and HBM in the decode, and emit from it what the model reads.}
\label{tab:title}
\arrayrulecolor{black!25}
\begin{tabular}{@{}>{\raggedright\arraybackslash}p{0.145\linewidth}>{\raggedright\arraybackslash}p{0.38\linewidth}!{\color{black!25}\vrule}>{\raggedright\arraybackslash}p{0.38\linewidth}@{}}
\arrayrulecolor{black}\specialrule{0.8pt}{0pt}{0pt}
 & \textbf{Prefill} {\color{nvgrey}(compute-bound)} & \textbf{Decode} {\color{nvgrey}(HBM-bound)} \\
\arrayrulecolor{black}\specialrule{0.5pt}{0pt}{0pt}\arrayrulecolor{black!25}
Property & writes are easier than reads: a deep layer's write needs little of the deep computation (\S\ref{sec:depth}) & the state sink: a rank-one term along a fixed direction of each head holds most of the states' energy; the rest is low-rank (\S\ref{sec:sink}) \\ \hline
Model reads & caches and states of layers $k..L{-}1$ for each prompt token & read-out and update of each $d_k{\times}d_v$ recurrent state \\ \hline
Shallow encoding & \emph{in depth}: prompt tokens $x_1..x_{N-1}$ run only layers $0..k{-}1$; their output $h^{(k)}$ stands in for layers $k..L{-}1$ & \emph{in representation}: the shallow state, the sink $a\vbar^{\top}$ plus rank-$R$ factors $U$, $V$ \\ \hline
Emitted by & emitters, trained copies of the deep layers' write paths, write every deep cache and state from $h^{(k)}$ & the factors, which give the read-out and the exact update; compressed to rank $R$ every $W$ steps \\ \hline
Saves & prompt tokens but the last run as few as \prefillLayerPctMin\% of the layers and \prefillFlopPctMin\% of the parameter FLOPs & the state is \ratioStateMin--\ratioStateMax$\times$ smaller after a compression, \ratioStateMeanMin--\ratioStateMeanMax$\times$ on average \\ \hline
States & \multicolumn{2}{>{\raggedright\arraybackslash}p{0.78\linewidth}@{}}{every stored recurrent state is shallow: the states that $x_1..x_{N-1}$ leave are compressed to the sink plus rank-$R$ factors at the end of the prefill, and the decode keeps this form} \\ \hline
Unchanged & \multicolumn{2}{>{\raggedright\arraybackslash}p{0.78\linewidth}@{}}{the weights; no attention cache is compressed or quantized; the last prompt token and every generated token run all $L$ layers} \\
\arrayrulecolor{black}\specialrule{0.8pt}{0pt}{0pt}
\end{tabular}
\arrayrulecolor{black}
\end{table}

\paragraph{Writes are easier than reads.}
A layer writes its memory of a token through its \emph{write path}, which reads only the residual stream that enters the layer. Predicting the next token, a \emph{read}, needs the attention and feed-forward blocks of every deep layer.
This \emph{write--read asymmetry} is large.
We call $k$ the \emph{anchor depth} and layer $k-1$, at \prefillLayerPctMin--\prefillLayerPctMax\% of the depth, the \emph{anchor layer}; its output is the residual stream $h^{(k)}$~\cite{elhage2021mathematical}. Fed into the deep layers' own write paths without any training, $h^{(k)}$ moves the next-token distribution by a Kullback--Leibler (KL) divergence below 0.1 nats per token.
An untrained early exit at the same depth, the output head applied to $h^{(k)}$ (the logit lens; \citealp{logitlens2020}), is off by \lensSUKL--11.1 nats per token (Figure~\ref{fig:prefill-cut}).
We trace the gap to the two roles of a deep layer. At a prompt token, its only lasting output is the memory it writes; the rest of its computation reaches the reply only through the inputs of later writes (\S\ref{sec:depth}).

\paragraph{Recurrent states have a sink.}
In the median Gated DeltaNet head of \model, one direction holds 85\% of the state's energy, its squared Frobenius norm.
This direction is close to the head's mean value $\vbar$. The values' component along $\vbar$ carries most of their energy, and the linear recurrence accumulates it into a rank-one term $a\vbar^{\top}$, the \emph{state sink}, whose coefficient follows the head's own recurrence (Lemma~\ref{prop:decomp}).
Pooled over heads, the sink holds \sinkPooledMinPct--\sinkPooledMaxPct\% of the state energy on all five models.
Like the attention sink of softmax layers~\cite{streamingllm2024}, it forms regardless of the prompt, on \model even on random tokens; unlike it, it lies along a direction of the state, not at a position in the sequence.
The sink holds the energy but not the prompt. After the head's normalization, it acts as a bias and as the gain of the rest, and replacing it by its mean over other prompts costs only 0.002 nats per token. What the head has read lies in the orthogonal \emph{content}, a short working memory: on \model, a median of 6 directions hold 90\% of its energy (\S\ref{sec:sink}).

\paragraph{The model's own update preserves the form.}
A compact state is useful only if it survives every update without growing back to $d_k\times d_v$.
A sink plus low-rank factors does: the form $a\vbar^{\top}+UV^{\top}$ is closed under the Gated DeltaNet, Kimi Delta Attention and Mamba-2 updates, each step updating the sink's coefficient and appending one column to each factor, an exact rank-one update of the content (Lemma~\ref{prop:closed}).
The only approximation is a compression of the factors back to rank $R$, once at the end of the prefill and then every $W$ steps; it never touches the sink. In the open loop, where the compressed and the exact recurrence see the same inputs, its errors do not compound: the accumulated error stays within the largest single compression error times the head's decay horizon, however many compressions occur (Corollary~\ref{cor:periodic}). In the closed loop, where errors also change later inputs, the measured drift grows only \clGrowthMin--\clGrowthMax$\times$ from the first to the last 4K of \clStepsMax{} decode steps while the number of compressions grows 4$\times$, on the four models followed this long (\S\ref{sec:eval-drift}).

\paragraph{\duet.}
\duet, \emph{Shallow Encoding Emits Deep Memories}, builds one part on each property (Figure~\ref{fig:arch}, Table~\ref{tab:title}).
In the prefill, its encoding is shallow in depth. The first $N-1$ prompt tokens run only layers $0..k{-}1$, and \emph{emitters}, trained copies of the deep layers' write paths, write every deep attention cache and recurrent state from $h^{(k)}$. An optional compact \emph{prompt code} of $h^{(k)}$, read by the emitters, can stand in for the deep memories in a prefix cache.
In the decode, its encoding is shallow in representation. From the end of the prefill on, every recurrent state is stored as its sink plus rank-$R$ factors instead of a dense $d_k\times d_v$ matrix, and every step computes its read-out and its exact update from them.
Each encoding is shallow along the bottleneck of its phase: the prefill's saves FLOPs, and the decode's saves HBM bytes. All prompt tokens but the last run as few as \prefillLayerPctMin\% of the layers and \prefillFlopPctMin\% of the parameter FLOPs, and the state is \ratioStateMin--\ratioStateMax$\times$ smaller after each compression.
The original weights stay frozen, no attention cache is compressed, and the last prompt token and every generated token run all $L$ layers.
Every setting follows from one set of criteria, applied to inexpensive measurements of the original model (\S\ref{sec:depth-k}).

\paragraph{Evidence.}
We apply \duet to six checkpoints of five production hybrids of 30B to over 300B parameters: \model in bf16 and NVFP4, \lightning, \kimi, \glm and \super.
Each is compared with its original model question by question, with the same sampling seeds.
Under the official graders, \duet is nearly lossless: \examWithinTwoSEN{} of the \examCellsTotal{} checkpoint--exam cells of \examCountWord{} exams lie within two standard errors of the original model or above it, and the repetitions on which no design choice rested agree with them (\S\ref{sec:eval}). Its answers change about as often as those of a second sample of the original model.
The next-token distributions stay close. With the exact prefill, the shallow state drifts \precSeedDriftMin--\precSeedDriftMax{} nats per token; NVIDIA's official NVFP4 checkpoints drift \nvRulerRatioMin--\nvRulerRatioMax$\times$ as much on the three models that have one (\S\ref{sec:eval-drift}).
Served with SGLang on NVIDIA GB300 GPUs on an agentic-coding trace, \model with \duet delivers 1.2--1.4$\times$ the original model's throughput per GPU, both peak to peak and at an equal per-user rate of 40 tokens/s: shallow prefill relieves the prefill instance in disaggregated serving, and shallow states relieve the recurrent-state pool that limits the batch in aggregated serving (\S\ref{sec:perf}).

Neither property was designed into these models.

\paragraph{Contributions.}
\begin{itemize}
\item Two properties of trained hybrid models: the write--read asymmetry, for recurrent states as well as attention caches, and the state sink, each measured on five production models (\S\S\ref{sec:depth}--\ref{sec:sink}).
\item \duet, which turns both into a serving method for frozen checkpoints, with a proof that the Gated DeltaNet, Kimi Delta Attention and Mamba-2 updates preserve its state form exactly and an open-loop error bound that does not grow with the number of compressions (\S\ref{sec:design}).
\item An evaluation of six checkpoints against their original models on official exams, decode drift and serving throughput: \duet matches the original models' exam scores, and \model with \duet delivers 1.2--1.4$\times$ the throughput per GPU (\S\S\ref{sec:eval}--\ref{sec:perf}). We will release the code, the component sets and their training data.
\end{itemize}

\section{Setting}
\label{sec:setting}

\paragraph{Cost model.}
We count the cost of each phase with a roofline model. A prefill chunk of $n$ tokens reads the $M_{\mathrm w}$ bytes of weights once and spends $F$ FLOPs per token, two per parameter the token meets; it is compute-bound once its arithmetic intensity $nF/M_{\mathrm w}$ exceeds the GPU's ridge point, its FLOP rate divided by its HBM bandwidth. $F$ omits attention-score FLOPs, which \duet skips in the deep attention layers, and recurrence-scan FLOPs, which it keeps, so on long prompts $F$ understates its saving. For a chunk without the last prompt token, \duet lowers both $F$ and the weight bytes read. A decode GPU with $M_{\mathrm{HBM}}$ bytes of HBM holds the weights and $B$ resident requests; at context length $T$, each occupies $M_{\mathrm r}(T)$ bytes of caches and states, and each decode step moves $Q_{\mathrm r}(T)$ bytes for it: its recurrent states, read and written in full, and the cache entries the step reads. These are all entries under dense attention; under the sparse attention of \model and \glm, they are 2{,}048 selected tokens per layer plus the indexer keys of all tokens, pooled over blocks of four. A step also reads the weights the batch touches, $M_{\mathrm w}^{\mathrm{step}}\le M_{\mathrm w}$ (on a mixture-of-experts model, only the experts the batch is routed to). Hence
\begin{equation}
t_{\mathrm{step}}\;\gtrsim\;\frac{M_{\mathrm w}^{\mathrm{step}}+B\,Q_{\mathrm r}(T)}{\text{HBM bandwidth}},\qquad B\;\le\;\frac{M_{\mathrm{HBM}}-M_{\mathrm w}}{M_{\mathrm r}(T)},
\label{eq:cost}
\end{equation}
and the tokens per second per GPU, $B/t_{\mathrm{step}}$, depend on a request's memories twice: through capacity, as their size caps the batch, and through bandwidth, as their traffic adds to every step. In both, the attention cache grows with $T$; the recurrent state, which \duet shrinks, does not. A prefix cache, which keeps the memories of cached prompts for reuse, competes for the same HBM.

\paragraph{Models.}
The five models have 30B to 320B parameters (Appendix~\ref{app:details}) and differ in their recurrent layers, attention and weight precision. \model~\cite{qiu2026qwen38next} is a 48-layer mixture-of-experts model in which every fourth layer is block-sparse softmax attention and the other three are Gated DeltaNet layers~\cite{gateddeltanet2025} of 48 value heads. Each head keeps an fp32 state $S_t$ of $d_k\times d_v$ (key by value dimension, $d_k=d_v=128$), updated by the gated delta rule
\begin{equation}
\begin{split}
S_t&=g_t\,(I-\beta_t \kappa_t \kappa_t^{\top})\,S_{t-1}+\beta_t\,\kappa_t v_t^{\top},\\
o_t&=S_t^{\top}q_t,\qquad y_t=\rms(o_t)\odot\operatorname{sigmoid}(\zeta_t),
\end{split}
\label{eq:gdn}
\end{equation}
with keys $\kappa_t$ of unit norm and queries $q_t$ of norm $d_k^{-1/2}$ in $\Rr^{d_k}$, values $v_t$ and read-outs $o_t$ in $\Rr^{d_v}$, a write strength $\beta_t\in(0,1)$, a decay $g_t\in(0,1)$, a per-head RMSNorm~\cite{zhang2019rmsnorm} and, in this model, a sigmoid output gate with input $\zeta_t$. \model's sparse attention caches 2{,}112 bytes per token and layer, and its residual stream holds 10{,}240 numbers per token (four parallel streams).
\lightning~\cite{nemotron3_2026} has 52 layers: 23 Mamba-2 layers~\cite{mamba2_2024} of 64 heads with a $64\times128$ state each, 6 attention layers and 23 stateless expert layers. \super~\cite{nemotron35super_2026} has 88 layers of the same kinds: 40 Mamba-2 layers of 128 heads with $64\times128$ states, 8 attention layers and 40 expert layers whose 512 routed experts compute in a 1{,}024-number latent space; we use its language model without the vision encoder.
\kimi~\cite{kimilinear2025} has 27 layers: 20 Kimi Delta Attention layers of 32 heads, each with a $128\times128$ state under a gated delta rule with per-channel decay, and 7 multi-head latent attention (MLA) layers~\cite{deepseek2024v2} that cache 576 numbers per token. \glm~\cite{zai2026glm53flash,glm5team2026glm5} has 34 Kimi Delta Attention layers~\cite{kimilinear2025} of 64 heads with $128\times128$ states and 11 DeepSeek sparse attention layers~\cite{deepseek2025v32}, each caching a 512-number latent and a pooled indexer key (\gdsaB{} bytes per token and layer) and attending to 2{,}048 selected tokens. Its residual stream is four streams of 4{,}096 mixed by manifold-constrained hyper-connections~\cite{zhu2025hyperconnections,xie2025mhc}. Its expert and dense feed-forward weights and most sparse-attention projections are fp8; its Kimi Delta Attention layers stay bf16.
Each checkpoint gets one \emph{component set}: the emitters, prompt code and state settings that \duet adds to it. We treat NVIDIA's NVFP4 release of \model, \modelnv{}, which stores its routed experts in the four-bit NVFP4 format~\cite{nvfp4_2025}, as a separate checkpoint with its own component set. Appendix~\ref{app:details} gives the remaining dimensions.

\paragraph{Memory per request and accounting rules.}
Table~\ref{\mainMemoryLabel} gives the bytes of one request at the precision of each model's reference implementation; the recurrent-state bytes do not depend on the context length. Three rules fix what we count. First, the original weights are frozen, and every original component (weights, attention caches, arithmetic) keeps its precision: \duet compresses no attention cache, its prompt code is the only quantized component it adds, and the exams use no speculative decoding. Second, only the emitters and the prompt code are trained, on public text, synthetic retrieval documents, public prompts and the frozen model's own responses to public prompts, the kinds of data that published distillation recipes use~\cite{puzzle2024,mambainllama2024} (Appendix~\ref{app:training}). Third, accuracy is measured on official exams, paired with the original model (\S\ref{sec:eval}).

\begin{table*}[t]
\centering\footnotesize
\setlength{\tabcolsep}{2pt}
\caption{\duet's recurrent state is \ratioStateMin--\ratioStateMax$\times$ smaller right after a compression, and its prefill runs fewer layers; attention caches stay uncompressed. Bytes of one request (fp32 states, bf16 attention caches; MB $=10^6$ bytes), for the representation, not an engine's allocation. \duet's state, the factors plus the explicit sink, is at rank $R$ after a compression, $R{+}W/2$ on average (the midpoint of a compression window; resident and traffic rows) and $R{+}W$ at the peak, just before a compression. 4K$+$4K: a 4{,}096-token prompt after 4{,}096 generated tokens; 32K$+$8K: a 32{,}768-token prompt after 8{,}192. Traffic counts every state as read and written in full at every step, \duet's factors included, although the update of \S\ref{sec:design-state} rescales $U$, updates the sink's coefficient and leaves $V$ unchanged apart from the appended column. \model and \glm read 2048 selected tokens per sparse layer and step, plus the indexer keys of all tokens, pooled over blocks of four (\glm: in the model's own fp8); the NVFP4 \model checkpoint has the same caches and states. Short-convolution windows (0.8--5.0\,MB per request, equal in both columns) are not included. FLOPs: 2 per parameter that a token meets; compression per generated token in fp64 and in fp32 GFLOP (Appendix~\ref{app:perf-model}).}
\label{tab:bytes}
\begin{tabular*}{\textwidth}{@{\extracolsep{\fill}}lrr@{\hspace{8pt}}rr@{\hspace{8pt}}rr@{\hspace{8pt}}rr@{\hspace{8pt}}rr@{}}
\toprule
 & \multicolumn{2}{c}{\ifreport\shortstack{Qwen3.8-\\Flash-Next}\fi} & \multicolumn{2}{c}{\ifreport\shortstack{Nemotron-3.5-\\Lightning}\fi} & \multicolumn{2}{c}{\ifreport\shortstack{Kimi-\\Linear}\fi} & \multicolumn{2}{c}{\ifreport\shortstack{GLM-5.3-\\Flash}\fi} & \multicolumn{2}{c}{\ifreport\shortstack{Nemotron-3.5-\\Super}\fi} \\
\cmidrule(lr){2-3}\cmidrule(lr){4-5}\cmidrule(lr){6-7}\cmidrule(lr){8-9}\cmidrule(l){10-11}
 & Orig. & \duet & Orig. & \duet & Orig. & \duet & Orig. & \duet & Orig. & \duet \\
\midrule
State at $R$ (MB) & 113.2 & 29.2 & 48.2 & 18.8 & 41.9 & 21.3 & 142.6 & 72.4 & 167.8 & 65.5 \\
\quad mean, $R{+}W/2$ (MB) & 113.2 & 43.4 & 48.2 & 27.9 & 41.9 & 31.8 & 142.6 & 108.1 & 167.8 & 97.0 \\
\quad peak, $R{+}W$ (MB) & 113.2 & 57.5 & 48.2 & 36.9 & 41.9 & 42.3 & 142.6 & 143.7 & 167.8 & 128.5 \\
\addlinespace
Attn.\ cache / token (B) & 25{,}344 & 25{,}344 & 6{,}144 & 6{,}144 & 8{,}064 & 8{,}064 & 11{,}627 & 11{,}627 & 8{,}192 & 8{,}192 \\
Resident, 4K$+$4K (MB) & 321 & 251 & 99 & 78 & 108 & 98 & 238 & 203 & 235 & 164 \\
Resident, 32K$+$8K (MB) & 1{,}151 & 1{,}081 & 300 & 280 & 372 & 362 & 619 & 584 & 503 & 433 \\
\addlinespace
State traffic / step (MB) & 226 & 87 & 96 & 56 & 84 & 64 & 285 & 216 & 336 & 194 \\
Step traffic, 4K$+$4K (MB) & 283 & 143 & 147 & 106 & 150 & 130 & 311 & 242 & 403 & 261 \\
Step traffic, 32K$+$8K (MB) & 308 & 168 & 348 & 307 & 414 & 394 & 323 & 254 & 671 & 530 \\
\addlinespace
Prefill layers run & 48 & 31 & 52 & 33 & 27 & 19 & 45 & 31 & 88 & 78 \\
Prefill / prompt token (GFLOP) & 12.1 & 9.1 & 5.8 & 4.1 & 5.5 & 4.1 & 32.2 & 23.5 & 23.4 & 21.3 \\
Compression, fp64 (GFLOP) & -- & 0.18 & -- & 0.11 & -- & 0.10 & -- & 0.35 & -- & 0.40 \\
\quad fp32 (GFLOP) & -- & 0.49 & -- & 0.21 & -- & 0.16 & -- & 0.55 & -- & 0.74 \\
\bottomrule
\end{tabular*}
\end{table*}

\label{endsec:setting}

\section{Shallow in Depth: The Write--Read Asymmetry}
\label{sec:depth}

Writing a prompt token's deep memories needs far less of the deep computation than predicting the next token. Fed the anchor layer's output $h^{(k)}$, at \prefillLayerPctMin--\prefillLayerPctMax\% of the depth, the deep layers' own write paths move the next-token distribution by a KL below 0.1 nats per token without any training, while an untrained early exit, the output head applied to $h^{(k)}$, is off by \lensSUKL--11.1 nats per token. \duet's prefill rests on this asymmetry; its risk is retrieval, which a copy probe guards against (\S\ref{sec:depth-k}).

\paragraph{Shallow prefill.}
A prefill runs every layer for every prompt token but uses only the last token's output. For the other tokens, the deep layers' only lasting result is the \emph{deep memories}: the keys and values that each deep attention layer caches and the update that each deep recurrent layer folds into its state. \duet keeps these memories and replaces the rest of the deep computation. The first $N-1$ prompt tokens run layers $0..k{-}1$, up to the anchor layer, and skip the full computation of layers $k..L{-}1$, whose memories emitters write from the anchor layer's output $h^{(k)}$; the criteria of Section~\ref{sec:depth-k} set the anchor depth $k$ per model. The last prompt token and every generated token run all $L$ layers: the last prompt token reads the emitted memories, writes its own with the original layers and gives the distribution of the first reply token.

\begin{figure*}[t]
\centering
\includegraphics[width=\textwidth]{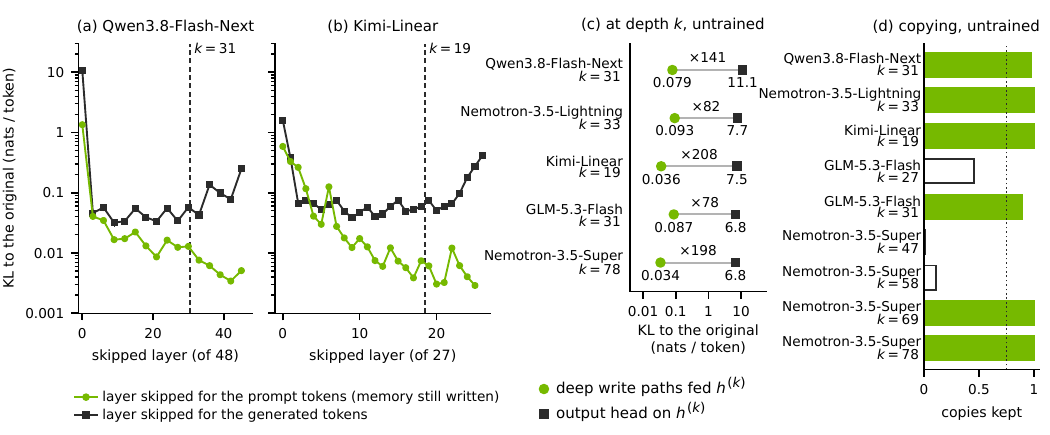}
\caption{Writes are easier than reads in the hybrids we serve. (a, b) Per-token KL from the original model's next-token distribution when one layer is skipped only for the prompt tokens, its memory still written (green), or only for the generated tokens (black); the deepest layers matter most for the generated tokens because they produce the next-token distribution. Dashed: the anchor depth. (c) KL at the anchor depth without training: the deep layers' write paths fed $h^{(k)}$ (green) against an untrained early exit, the output head applied to $h^{(k)}$ (black), with the ratio of the two KLs. (d) The risk is retrieval: identifiers copied by the untrained shallow prefill, as a share of the original model's; hollow: below three quarters. Protocols: Appendix~\ref{app:depth}.}
\label{fig:prefill-cut}
\end{figure*}

\paragraph{Every deep memory is a function of a shallow residual stream.}
Layer $\ell$ writes the memory of token $t$ from the residual stream $h^{(\ell)}_t$ that enters it. An attention layer projects this input to a key and a value. A linear-attention layer projects it to a key, a value, a write strength and a decay, convolves the key and value over the last few tokens, and folds them into its state. Through the attention and feed-forward blocks of layers $k..\ell{-}1$, $h^{(\ell)}_{1..T}$ is a function of the residual stream $h^{(k)}_{1..T}$ at any earlier depth $k$:
\begin{equation}
\mathrm{mem}_\ell(h^{(\ell)}_{1..T})=\mathrm{mem}_\ell\big(f_{k\to\ell}(h^{(k)}_{1..T})\big).
\label{eq:suff}
\end{equation}
The anchor layer's output thus determines every deeper memory. The question is how much of $f_{k\to\ell}$ a write needs.

\paragraph{Emitters.}
A write needs very little of $f_{k\to\ell}$. An \emph{emitter} for layer $\ell$ is an architectural copy of the layer's write path: its read of the residual stream (on \model and \glm, the input-dependent weights that mix the four streams into the layer's input), its normalization and its key and value projections, plus a sparse-attention layer's indexer key and a linear-attention layer's write-strength and decay projections, short convolution and scan. The emitter starts from the layer's own weights but reads $h^{(k)}$ of the same tokens instead of $h^{(\ell)}$, or its reconstruction $\hat h^{(k)}$ when a prompt code is used.

\paragraph{The untrained gap.}
Without training, the emitters form the \emph{untrained shallow prefill}: POP's training-free fill~\cite{pop2026} extended to recurrent layers, and our baseline. Its increase in next-token cross-entropy is the \emph{untrained gap}: \ruleFNGapK{} nats per token on \model at $k=31$, and \ruleKMGapK--\ruleLNGapK{} across the five models at the anchor depths we use. Its KL is 0.03--0.09 nats per token on all five models, against \lensSUKL--11.1 for the early exit at the same depth (Figure~\ref{fig:prefill-cut}c; Appendix~\ref{app:depth}). This contrast is the write--read asymmetry. We then train the emitters together with the prompt code against the frozen model (\S\ref{sec:design-depth}).

\paragraph{The asymmetry, layer by layer.}
Skipping one layer's output for the prompt tokens, while still writing its memory from the residual stream that enters it, costs less than skipping it for the generated tokens (Figure~\ref{fig:prefill-cut}a, b; Figure~\ref{fig:skip-more} in Appendix~\ref{app:depth}). In KL, this holds in all 16 sampled layers of \model, \skipLNCheaper{} of \skipLNSampled{} sampled layers of \lightning, \skipSUCheaper{} of \skipSUSampled{} of \super and 23 of the 26 layers of \kimi below its last; in next-token loss, it holds in 34 of the 35 layers of the transformer Qwen3-8B-Base~\cite{qwen3_2025} below its last (Figure~\ref{fig:prefill-ladders}). (Skipping the last layer for the prompt tokens changes no memory.) On these models, all exceptions lie in the first quarter of the layers. On \glm, the asymmetry holds in all \skipGLSampledFrom{} sampled layers from layer \skipGLFromLayer{} of 45 on, including every sampled layer above its anchor depth, and reverses below, where \duet keeps the original layers. At and above the first emitted layer, the generated-token skip costs 5.6--49 times as much as the prompt-token skip on \model, \skipLNRatioAbove{} on \lightning, 5.4--96 on \kimi, \skipSURatioAbove{} on \super and \skipGLRatioAbove{} on \glm.

\paragraph{Why writes are easier than reads.}
We attribute the asymmetry to the two roles of a deep layer. At a prompt token, only its write matters, and a skipped layer still writes its own memory, so the skip reaches the reply only through the inputs of later layers' writes. At a generated token, the deep layers must also produce the next-token distribution, a read that needs the attention and feed-forward blocks of every deep layer. In free-running generation, an error there also feeds back through the sampled tokens, whereas prompt tokens are given. Both reasons confine the shortcut to the prefill. They also explain the two curves of Figure~\ref{fig:prefill-cut}a, b in the deepest layers: the generated-token skip rises again because these layers produce the next-token distribution, while the prompt-token skip keeps falling because at a prompt token their output is never read.

\paragraph{The writes need not be exact.}
Beyond the first emitted layer, whose write path reads $h^{(k)}$ itself, the released emitters' attention caches differ from the original model's by \relCacheBeyondMin--\relCacheBeyondMax\% relative squared error per layer and tensor, and their recurrent states, pooled over each layer's heads, by 20--52\% on average over the emitted layers (Tables~\ref{tab:released} and~\ref{tab:staterr}). Yet the six released sets stay within a KL of \relExactMin--\relExactMax{} nats per token of the original model with exact states, and \relServedMin--\relServedMax{} with the states compressed every $W$ steps, as served (above a second run of the original model; Table~\ref{tab:released}). The later layers tolerate inexact memories, so the emitters are trained mainly on the output KL rather than on a memory-reconstruction loss (\S\ref{sec:design-depth}).

\paragraph{The prompt code.}
Since $h^{(k)}$ determines every deep memory, one vector per token suffices to restore a prompt's deep memories for a prefix cache or a transfer between nodes. The prompt code is a compact, lossy encoding of this vector for storage; a running request keeps the emitted caches and states, which its decode steps read. On \model, $h^{(31)}$ holds 10{,}240 numbers per token, about twice the bf16 keys and values of its five attention layers above the anchor layer. \duet codes the offset $\Delta=h^{(k)}-h^{(0)}$ from the token's input embedding $h^{(0)}$, which is looked up from the token id. With an encoder $E$, a decoder $D$ and the mean offset $\mu$, the code keeps $r$ coordinates $z=E(\Delta-\mu)$ in NVFP4, plus the exact bf16 values, at gap-coded one-byte indices, of the $m$ largest entries of the remainder $\Delta-\mu-Dz$. These entries are outliers, and a fixed rank-$r$ code could capture each only by spending one of its directions on it (Appendix~\ref{app:rule}). The emitters do not need the code, but every set with a code reads the decoded $\hat h^{(k)}$, $h^{(0)}+\mu+Dz$ plus the exact values, in training and in every exam, so its exams score the memories that a stored code would restore.

\paragraph{How large the code must be.}
Copy fidelity, not the spectrum of the offset, sets the code's rank. \emph{Closed-form} codes, the untrained principal-component codes that initialize training, of rank 4{,}096 and 6{,}144 fail on \model to copy identifiers that the shallow prefill without a code (\emph{uncoded}) copies, although rank 4{,}096 holds \ruleFNFourKVar{} of the offset's variance (Table~\ref{tab:depth-rule}, Appendix~\ref{app:rule}). Both \model sets store $r=\fnRank$ coordinates plus $m=\fnSpikes$ exact values, \latentKB\,KB per token. The \emph{deep cache}, the attention cache of the layers above the anchor layer, takes \deepKB\,KB in bf16 and about as much as the code in fp8. With $m=128$ exact values each, \lightning's code keeps $r=2{,}048$ coordinates of a residual stream of 2{,}688 numbers, \super's $r=512$ of 4{,}096, and \kimi's all $r=2{,}304$. The code is trained end to end with the emitters, which learn to read it: on all \relBypassSets{} sets with a code, feeding the released emitters the exact residual stream instead of $\hat h^{(k)}$ raises their KL by at least \relBypassMinPct\% (Table~\ref{tab:released}).

\paragraph{Where a code does not pay.}
\glm's residual stream is 16{,}384 numbers wide, while its four sparse attention layers at and above $k=31$ cache only \gdeepB{} bytes per token. No probed closed-form code within the bytes of its bf16 deep cache copies identifiers: the best copies at most \ruleGLCopyCapMax{} of \ruleGLCopyN{}, where the uncoded shallow prefill copies \ruleGLCutB. \glm therefore runs without a code, and its emitters read the exact $h^{(31)}$.

\subsection{Choosing the Settings}
\label{sec:depth-k}
One set of criteria and thresholds fixes the anchor depth, the prompt code and the state rank of every set. It uses only inexpensive measurements of the original model, its untrained shallow prefill and closed-form codes; no exam score enters (exact statements and reasons: Appendix~\ref{app:rule}). The main measurement is the untrained gap $\mathrm{gap}(k)$ on 48 held-out 4K windows with 256 teacher-forced continuation tokens; it falls towards the deep layers (Figure~\ref{fig:gap}, Appendix~\ref{app:depth}). Each criterion guards against one failure:
\begin{itemize}
\item \emph{Depth.} $k$ is the shallowest anchor depth whose (i)~first emitted layer (index $k$) is a softmax-attention layer; (ii)~untrained gap is at most 0.075 nats per token, three quarters of a budget of 0.1 nats for the untrained shallow prefill and shallow state together; and (iii)~untrained shallow prefill copies at least three quarters of the identifiers that the original model copies on a copy probe, which hides one record with a universally unique identifier (UUID) in 4K--16K tokens of web text (Appendix~\ref{app:rule}). Copying needs its own criterion: an identifier is copied only if all its tokens are, so a per-token average loss hardly registers a failed copy.
\item \emph{Code.} The code keeps the full offset if that takes no more bytes than even the fp8 deep cache. Otherwise, it uses the smallest closed-form rank that takes no more bytes than the bf16 deep cache, which a stored code replaces, and shows no significant copy loss against the uncoded shallow prefill (exact McNemar test, $p<0.10$); failing that, there is no code. It keeps $m=256$ exact values on models whose chat-template tokens reconstruct worse than their other tokens, and $m=128$ otherwise.
\item \emph{State.} The \emph{state drift} is the per-token KL between the original model with exact states and with shallow states: the prompt-final state compressed once, then again every $W=R$ steps. Both runs are teacher-forced on the original model's own replies to 48 held-out prompts, and the KL is averaged over decode steps 961--1{,}024. The state rank $R\in\{8,16,32\}$ is the smallest whose state drift is at most a quarter of $\mathrm{gap}(k)$, else $R=32$; the compression period is $W=R$.
\end{itemize}
Table~\ref{tab:depth-rule} (Appendix~\ref{app:rule}) lists the measurements and resulting settings. The copy criterion is binding on \glm: at depth 27 its untrained gap is small (\ruleGLPrevGap), but its untrained shallow prefill copies only \ruleGLCutA{} of the original model's \ruleGLBase{} identifiers, so its anchor depth is 31, where it copies \ruleGLCutB. \kimi gets $R=32$ because its gap is small (\ruleKMGapK), not because its states compress worse.
\label{endsec:depth}

\section{Shallow in Representation: The State Sink}
\label{sec:sink}

Pooled over heads, most of the energy of a hybrid model's recurrent states lies in one rank-one term per head, along a fixed direction that does not depend on the prompt: the \emph{state sink}. The rest, the \emph{content}, is a low-rank working memory. The sink forms because the values share a common component that the linear recurrence accumulates; after the head's normalization, it acts as a bias and a gain, while what the head has read about the prompt lies mainly in the content. \duet keeps every layer in the decode (\S\ref{sec:depth}), so its decode saving must come from how the memories are represented, and this structure supplies it. We analyse it on all $36\times48=1{,}728$ Gated DeltaNet heads of \model, at the end of six held-out web prompts of 3{,}584 tokens, and score one-time interventions on these states by the loss on the next 512 tokens.

\paragraph{One direction carries the state.}
\label{sec:sink-obs}
Let $s_\star u_\star v_\star^{\top}$ be the top term of the singular value decomposition (SVD) of a head's state $S_T$. It carries a median 85\% of $\|S_T\|_F^2$, the second term about 5\%, and the median share of the top term lies between 71\% and 99.8\% in every layer (Figure~\ref{fig:sink-heat}, Appendix~\ref{app:details}). Its output-side vector $v_\star$ is nearly the same across unrelated prompts (median $|\cos|$ 0.92), on a shuffled prompt (0.90) and on random token ids (0.87; Figure~\ref{fig:sink-heat}c). The attention sink of softmax layers also forms regardless of the tokens' meaning and on random tokens~\cite{streamingllm2024,gu2024attentionsink}, but it is a position in the sequence, often the initial tokens, whereas the state sink is a direction of the state.

\paragraph{The direction is the state's component along the mean value.}
\label{sec:sink-identity}
Once the keys, decays and write strengths are fixed, the recurrence~\eqref{eq:gdn} is linear in the values, so splitting each value along a fixed direction splits the state in the same way.
\begin{lemma}[sink decomposition]
\label{prop:decomp}
Let $S_t=\Phi_tS_{t-1}+\beta_t\kappa_tv_t^{\top}$ with any $d_k\times d_k$ matrices $\Phi_t$ (in~\eqref{eq:gdn}, $\Phi_t=g_t(I-\beta_t\kappa_t\kappa_t^{\top})$). Fix $\vbar\neq0$ and split each value and the state along it:
\[
v_t=\omega_t\vbar+\varepsilon_t\ \ \text{with}\ \ \varepsilon_t^{\top}\vbar=0,\qquad
a_t=\frac{S_t\vbar}{\|\vbar\|^2},\qquad C_t=S_t-a_t\vbar^{\top},
\]
so that $C_t\vbar=0$. Then the sink coefficient $a_t$ and the content $C_t$ each follow the head's own recurrence,
\begin{equation}
a_t=\Phi_ta_{t-1}+\beta_t\omega_t\kappa_t,\qquad C_t=\Phi_tC_{t-1}+\beta_t\kappa_t\varepsilon_t^{\top},
\label{eq:decomp}
\end{equation}
and their energies add: for every $z\in\Rr^{d_k}$,
\[
\|S_t-z\vbar^{\top}\|_F^2=\|a_t-z\|^2\,\|\vbar\|^2+\|C_t\|_F^2 .
\]
In particular, $a_t\vbar^{\top}$ is the matrix of the form $z\vbar^{\top}$ closest to $S_t$.
\end{lemma}

The identity holds for any direction. We fix $\vbar$ as the head's mean value over generic text, independent of the prompt, because that direction carries most of the state (as measured below). We call the projection $a_T\vbar^{\top}$ the \emph{state sink} and $C_T$ the \emph{content}. The sink's coefficient $a_T$ follows the head's own recurrence, driven by $\omega_t$, the values' component along their mean, which averages one over generic text by construction. The component $\omega_t\vbar$ carries 60--99\% of the values' energy, $\mathbb{E}\|\omega_t\vbar\|^2/\mathbb{E}\|v_t\|^2$ (layer medians; 64--94\% from layer 8 on). If every value equals $\vbar$ and $S_0=0$, then $\omega_t=1$, $C_T=0$ and $a_T=\sum_t\Phi_T\cdots\Phi_{t+1}\beta_t\kappa_t$ depends on the keys, decays and write strengths alone; \citet{sinkrec2026} derive this case for one repeated key with constant decay and write strength.

\begin{figure*}[t]
\centering
\includegraphics[width=\mainfigwidth]{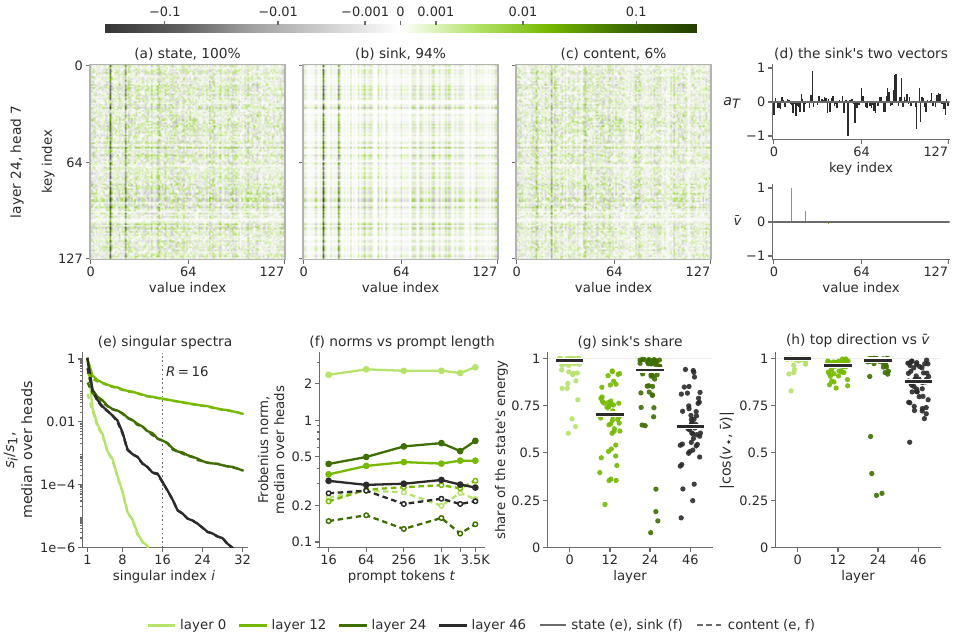}
\caption{A rank-one sink along the head's mean value holds most of a recurrent state's energy, and the content left over lies in few directions (\model; the first 3{,}584 tokens of one of the six held-out web prompts of Appendix~\ref{app:details}). (a) The $128\times128$ state of head 7 of layer 24, a middle head of the layer by the sink's share of the state's energy. (b) Its sink $a_T\vbar^{\top}$, the state's projection on the head's mean value $\vbar$ (estimated on other text). (c) The content $S_T-a_T\vbar^{\top}$. Titles: shares of $\|S_T\|_F^2$, which add to one. (a)--(c) share one colour scale, $\log(1+|x|/10^{-3})$ in the magnitude, nothing clipped. (d) $a_T$ and $\vbar$, each divided by its largest magnitude. (e) Singular spectra of the state (solid) and the content (dashed), divided by the head's top singular value; per-index medians over the layer's 48 heads. Dotted: the rank \duet keeps. (f) Frobenius norms of the sink (solid) and the content (dashed) against the prompt length; medians over heads. (g) The sink's share of the state's energy and (h) $|\cos|$ between $v_\star$ and $\vbar$, per head; bars: layer medians.}
\label{fig:sink-matrix}
\end{figure*}

\paragraph{The measurements match the decomposition.}
Over heads, the state's top singular term has a median (Frobenius) cosine similarity of 0.95 with the sink, and $u_\star$ a median $|\cos|$ of 1.00 with $a_T$ (Figure~\ref{fig:sink-matrix} shows one prompt). The sink holds a median 75\% of the state's energy, and more than half in the median head of every layer (53--98\%). In the four layers of Figure~\ref{fig:sink-matrix}f, the median norms of the sink and the content each stay within 17\% of their mean from 256 prompt tokens on, and the sink is the larger in every one. The content lies in few directions: a median of 6 hold 90\% of its energy and 17 hold 99\%, close to the rank $R=\fnR$ that \duet keeps on \model. Because the content is the state projected off one direction, $C_T=S_T(I-\hat v\hat v^{\top})$ with $\hat v=\vbar/\|\vbar\|$, its spectrum interlaces the state's, $s_{i+1}(S_T)\le s_i(C_T)\le s_i(S_T)$ (Figure~\ref{fig:sink-matrix}e). Low effective ranks have been reported for such states~\cite{nazari2026rank}, and in Qwen3-Next for one of two groups of heads, while the other approaches the rank bound~\cite{staterank2026}.

\paragraph{The other models.}
The Mamba-2 heads of \lightning decompose in the same way along the head's mean input. Over all $23\times64$ heads on six held-out prompts of 3{,}584 tokens, the sink holds a median 64\% of a head's energy, and the content reaches 90\% of its energy in a median of one direction and 99\% in four. The Kimi Delta Attention heads of \kimi and \glm decay each key channel at its own rate, and their values lie less along their mean: \sinkKMValPct\% and \sinkGLValPct\% of their energy in the median head, against \sinkFNValPct\% on \model. In the median head, their sink holds \sinkKMMedPct\% and \sinkGLMedPct\% of the energy, but it concentrates in the heads that hold the most energy and so holds \sinkKMPoolPct\% and \sinkGLPoolPct\% of the state energy pooled over all heads. Over the five models, the sinks hold \sinkPooledMinPct--\sinkPooledMaxPct\% of the pooled state energy (Appendix~\ref{app:details} gives every model's median and pooled shares). The content of \kimi and \glm is as low-rank as on \model: pooled over heads and layers by energy, rank 16 leaves \ruleKMBeyond{} and \ruleGLBeyond{} of it (\model: \ruleFNBeyond), and rank 32, the rank they keep (\S\ref{sec:depth-k}), leaves \ruleKMBeyondTT{} and \ruleGLBeyondTT{} (Appendix~\ref{app:rule}).

\paragraph{After normalization, the sink is a bias and a gain.}
\label{sec:sink-function}
Every gated-delta head normalizes its read-out before gating it. Write the read-out as $o_t=b_t+r_t$, with the sink's part $b_t=(q_t^{\top}a_t)\,\vbar$ and the content's part $r_t=C_t^{\top}q_t$. As $r_t\perp\vbar$, a unit-gain RMSNorm without its stabilizing constant returns $\sqrt{d_v}\,(b_t+r_t)/(\|b_t\|^2+\|r_t\|^2)^{1/2}$. The sink's part thus sets a fixed output direction, up to the sign of $q_t^{\top}a_t$, and through the normalizer scales the content, as the sink logit does in a softmax denominator. When $\|b_t\|\gg\|r_t\|$, the normalized read-out is $\sqrt{d_v}$ times this direction plus the content scaled by $\sqrt{d_v}/\|b_t\|$, up to an error of at most $\|r_t\|^2/(2\|b_t\|^2)$ relative to its norm $\sqrt{d_v}$ (Appendix~\ref{app:proofs}, Eq.~\eqref{eq:gain}). In most heads the sink's part is the larger: over the scored tokens, the ratio of the mean norms of $b_t$ and $r_t$ has a median of 2.3 over the heads and exceeds one in 82\% of them.

\paragraph{The sink holds the energy, the content holds the prompt.}
Eviction shows that both parts are needed but that the sink's dependence on the prompt matters little (Appendix~\ref{app:details}). Removing the sink while keeping the content costs $+0.073$ nats per token, and keeping only the sink $+0.067$. Replacing each head's sink by its mean over the other prompts costs only $+0.002$, and taking it from a prompt of random tokens $+0.017$. The sink's coefficient depends on the prompt, but what the head has read about the prompt lies mainly in the content.

\paragraph{The content is a short working memory.}
\label{sec:sink-content}
The content matters more than these evictions suggest. They act once, on the state at the end of the prefill (before the last prompt token, as in \duet), after which the last prompt token and the 512 scored tokens update it exactly. When the state is instead compressed after \emph{every} teacher-forced continuation token, keeping only the sink costs $+1.66$ nats per token over a base of 1.76 (Appendix~\ref{app:extra}). The content is dominated by recent writes. From $S_0=0$, $C_T=\sum_t\Phi_T\cdots\Phi_{t+1}\beta_t\kappa_t\varepsilon_t^{\top}$ is a sum of rank-one writes, each damped by the decays applied after it, so its $r$ most recent writes approximate it within $\sum_{t\le T-r}\big(\prod_{u=t+1}^{T}g_u\big)\beta_t\|\varepsilon_t\|$. This is small in heads whose typical decays are well below one. On \model, with each head's mean decay, the time scale $1/(1-\text{mean decay})$ is about 17 tokens in the median head, at most ten tokens in 41\% of the heads, and above 100 tokens in 28\%. The error bounds of \S\ref{sec:design-state} use instead the head's \emph{decay horizon} $1/(1-\bar g)$, set by its largest decay $\bar g$, a worst case.

\paragraph{What this means for \duet.}
The sink plus a rank-8 content costs $+0.023$ nats per token, against $+0.025$ for a plain rank-8 truncation, whose top term lies close to the sink. Keeping the sink apart also gives the stored form its structure: its direction is shared by all requests, it is never truncated, and it keeps the content orthogonal to $\vbar$. \duet thus stores a head's state, a bias and gain plus a low-rank working memory, as the sink plus rank-$R$ factors (\S\ref{sec:design-state}).
\label{endsec:sink}

\section{\duet}
\label{sec:design}

\duet follows from the two findings. Every deep memory is a function of a shallow residual stream (\S\ref{sec:depth}), so the prefill runs only layers $0..k{-}1$ on all but the last prompt token and emitters write the deeper memories. The recurrent states hold most of their energy, pooled over heads, in a sink along a fixed direction of each head, and the rest, the content, is low-rank (\S\ref{sec:sink}), so the decode keeps the sink exactly and the content at low rank. Section~\ref{sec:design-state} gives the exact update of this factored state and the bound on its only approximation; Section~\ref{sec:design-depth} trains the emitters and the prompt code against the frozen model. Each model has four settings (Table~\ref{tab:depth-rule}, Appendix~\ref{app:rule}): the anchor depth $k$, the prompt code ($r$ coded coordinates plus $m$ exact values; \glm has none), the state rank $R$ and the compression period $W$.

\subsection{The Shallow State}
\label{sec:design-state}
In the decode, where bytes are the bottleneck, \duet keeps each recurrent state as its sink plus a rank-$R$ content, $S=a\vbar^{\top}+UV^{\top}$. The mean value $\vbar\in\Rr^{d_v}$ is a per-head constant of the component set, shared by all requests. The factors $a\in\Rr^{d_k}$, $U\in\Rr^{d_k\times R}$ and $V\in\Rr^{d_v\times R}$ belong to the request, and the columns of $V$ are orthogonal to $\vbar$.

\paragraph{An exact update.}
Each step of the update~\eqref{eq:gdn} splits its value as in Lemma~\ref{prop:decomp}, $\omega=v^{\top}\vbar/\|\vbar\|^2$ and $\varepsilon=v-\omega\vbar$. It gives the part along $\vbar$ to the sink's coefficient and appends the rest as one new content column:
\begin{equation}
\begin{split}
a^{+}&=g\,a+\beta\,(\omega-g\,\kappa^{\top}a)\,\kappa,\\
U^{+}&=\big[\,gU,\;\kappa\,\big],\qquad V^{+}=\big[\,V,\;\beta\,(\varepsilon-gVU^{\top}\kappa)\,\big].
\end{split}
\label{eq:append}
\end{equation}
The scalar $\omega-g\,\kappa^{\top}a$ and the vector $\varepsilon-gVU^{\top}\kappa$ are the delta rule's own prediction error $v-gS^{\top}\kappa$, split along $\vbar$ and orthogonal to it, so between compressions the factored update is exact. Since $\varepsilon$ and the columns of $V$ are orthogonal to $\vbar$, so is the appended column: the content never gains a component along $\vbar$, so the split of Lemma~\ref{prop:decomp} is invariant under the update and is never recomputed. Between compressions, the factors thus hold a rank-$R$ summary and, much like a short attention cache, the most recent writes (keys in $U$, delta-rule corrections in $V$). The same exact update holds for Mamba-2, with the sink along the head's mean input, and for the per-channel decay of \kimi and \glm (Lemma~\ref{prop:closed}).

\paragraph{Read-out and bytes.}
At $w$ steps after the last compression ($1\le w\le W$; a step reads after its update), the read-out $o=(q^{\top}a)\,\vbar+V(U^{\top}q)$ uses all $R+w$ content columns and costs $(d_k+d_v)(R+w+1)$ multiply-adds instead of $d_kd_v$: $2\cdot128\,(R+w+1)$ instead of $128^2$ on \model. The sink is its own term rather than one more factor column because $\vbar$ is shared by all requests; it then costs $d_k$ numbers per request instead of $d_k+d_v$, and no compression spends a rank on it. A head holds $(R{+}1)d_k+Rd_v$ numbers right after a compression and $(R{+}W{+}1)d_k+(R{+}W)d_v$ just before the next, against $d_kd_v$ for a dense head: 16.9\,KB, 33.3\,KB and 65.5\,KB in fp32 on \model (Table~\ref{\mainMemoryLabel} gives the bytes per request).

\paragraph{Periodic compression and its bound.}
Every $W$ steps, a warm-started subspace iteration truncates the $(R+W)$-column content back to rank $R$. This truncation is the shallow state's only approximation. It leaves the sink untouched, and the content stays orthogonal to $\vbar$ because the truncation only projects it within its own column or row space (Lemma~\ref{prop:closed}). In the open loop, the compressed state $\tilde S_t$ and the exact recurrence $S_t$ see the same keys, values, decays and write strengths. If every decay is at most $\bar g<1$ and $\bar\epsilon$ is the largest Frobenius norm of a single compression error, then for every step $t$ (Corollary~\ref{cor:periodic}, Appendix~\ref{app:proofs})
\[
\|\tilde S_t-S_t\|_F\;\le\;\frac{\bar\epsilon}{1-\bar g^{\,W}}\;\le\;\frac{\bar\epsilon}{1-\bar g}.
\]
Repeated compressions thus do not compound in the open loop: the error stays within one compression error times the head's decay horizon $1/(1-\bar g)$, however many compressions occur. The bound also holds with $\bar g$ replaced by the largest mean decay over the $W$ steps between two compressions, and the sink coefficient stays exact. In the closed loop, a compression error also changes later keys and values; Section~\ref{sec:eval-drift} follows this loop over \clStepsMax{} decode steps.

The rule of \S\ref{sec:depth-k} gives $R=W=\fnR$ on both \model sets, \lightning and \super and $R=W=32$ on \kimi and \glm (Table~\ref{tab:depth-rule}). The shallow state also composes with 16-bit storage of the factors (\S\ref{sec:eval-drift}).

\paragraph{Serving.} \duet runs in SGLang~\cite{sglang2024} as two opt-in settings, one per part, so an operator enables either or both per workload. With shallow prefill, the first $N-1$ prompt tokens run layers $0..k{-}1$ and the emitters, and the last prompt token runs all $L$ layers. The state left by the first $N-1$ prompt tokens is compressed once, and $x_N$ runs as the first decode step. In disaggregated serving, a prefill node ships only the sink coefficients and the factors of each recurrent state to the decode node. Each request counts its own steps from the end of its prefill, and the compressions due in a step are batched over all heads of those requests.

The engine's recurrent-state pool holds each \model state as its sink coefficient and factors; for \lightning, \super, \kimi and \glm, the exam engines keep the state dense between compressions and factor it at each compression, which gives the same state in exact arithmetic (Lemma~\ref{prop:closed}).

\subsection{Training}
\label{sec:design-depth}\label{sec:design-recover}
\begin{figure}[t]
\centering
\includegraphics[width=\textwidth]{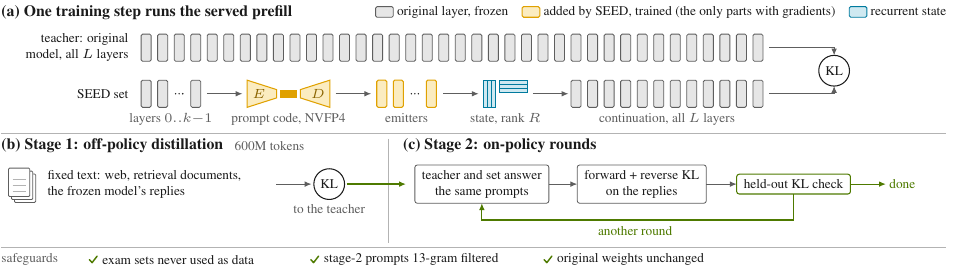}
\caption{One recipe trains every component set. Only \duet's components are trained, with the frozen model as the only teacher, in a forward pass that runs the prefill as served. Stage~1 distils on fixed text; stage~2 is on-policy, on the replies of the teacher and of the current set to the same prompts. Per-model terms: Appendix~\ref{app:training}.}
\label{fig:training}
\end{figure}
One recipe trains every component set against the frozen model, which stays the only teacher. We train all new parameters jointly: the emitters and the prompt code's encoder $E$ and decoder $D$, \trainableKM--\trainableGL{} exported parameters per set (Table~\ref{tab:training-details}). The main objective is the KL divergence to the frozen checkpoint's next-token distribution (Figure~\ref{fig:training}; Appendix~\ref{app:training} lists the auxiliary terms). The emitters start as copies of the frozen layers' write paths, $E$ and $D$ as the top principal directions of the offset's covariance, and $\mu$ is the mean offset, a fixed buffer.

\paragraph{The served prefill in every training step.}
The training forward pass runs the prefill as inference does: the prompt code passes through its storage quantizer (straight-through~\cite{bengio2013estimating}; the exact values are selected without gradient), and every prompt-final state, the state left by the first $N-1$ prompt tokens, is compressed at the decode rank (on \glm, every emitted layer's). Every component thus learns to write memories that survive their stored form. Training departs from inference only in truncating by an exact eigendecomposition instead of a subspace iteration and in running the continuation without the periodic compression (Appendix~\ref{app:training}).

\paragraph{Two stages.}
\emph{Stage 1} distils the frozen model in 4K and 16K windows of public text~\cite{fineweb2024,smollm2_2025}, public reasoning and chat corpora, synthetic retrieval documents and the frozen model's own responses to public prompts; part of the windows are aligned to the start of an answer (Table~\ref{tab:training-data} gives every set's data). \emph{Stage 2} runs on-policy rounds in which the frozen model and the current set answer public prompts. The forward KL applies to the frozen model's answers, and both the forward and the reverse KL to the set's own answers; the reverse KL penalizes the set for tokens that it writes and the frozen model would not~\cite{agarwal2024gkd,gu2024minillm,ko2025distillm2}. Stage 2 runs one or two rounds; after each, we measure the held-out KL and the divergence of the set's own replies (Appendix~\ref{app:training}). Every model releases its stage-2 set except \super, whose stage-1 set is released.

\paragraph{Model-specific terms.}
\lightning, \glm and \super add a term that aligns the emitted prompt-final states with the frozen model's. Without it, the emitted states of \lightning and \glm are the furthest from the frozen model's as the layers read them (Table~\ref{tab:staterr}, Appendix~\ref{app:recipe}); \super follows \lightning's recipe. \glm adds a second term for the indexer keys that its emitters write.

\begin{figure}[t]
\centering
\includegraphics[width=\textwidth]{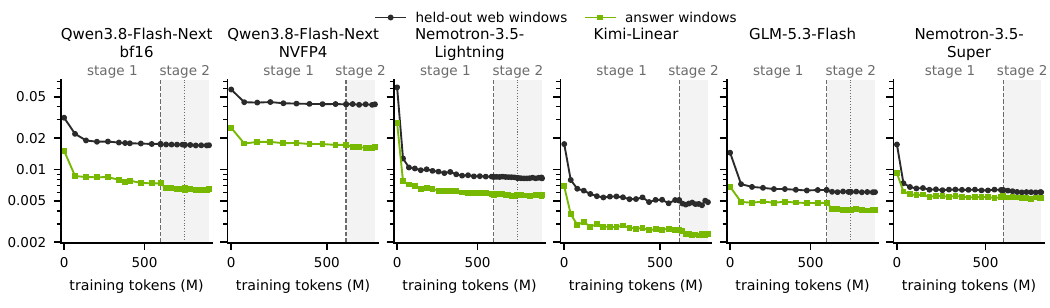}
\caption{Every set converges the same way: most of the KL reduction comes early in stage~1. Teacher-forced KL to the frozen model on held-out web and answer windows against training tokens (dashed: start of stage~2; dotted: start of a further round).}
\label{fig:curves}
\end{figure}
\paragraph{Convergence and cost.}
On every set (Figure~\ref{fig:curves}), at least \curveFracMinPct\% of stage~1's reduction of the held-out KL happens within its first \curveFracTokens{} tokens, a quarter of the stage; afterwards the KL changes slowly. Stage~2 changes the teacher-forced KL on held-out web windows by less than 0.0006 nats per token: it acts on the set's own on-policy replies, which a teacher-forced KL does not see (Appendix~\ref{app:training-served}). Training is a one-time cost per checkpoint and changes no original weight. For the bf16 \model set, stage 1 takes \trainMinutes{} minutes on \trainNodes{} nodes of four GPUs, and stage 2 another \trainTwoMinutes{} minutes plus the time to serve its generations (Table~\ref{tab:train}). Appendix~\ref{app:training} gives the settings, the data and the costs (Table~\ref{tab:train}).
\label{endsec:design}

\section{Accuracy}
\label{sec:eval}

\begin{figure}[t]
\centering
\includegraphics[width=\textwidth]{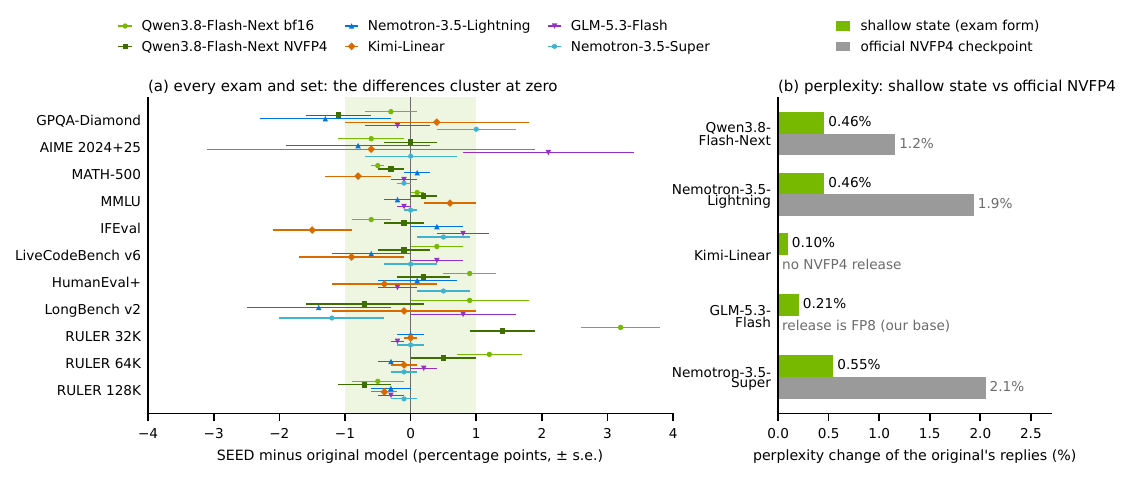}
\caption{\duet matches the original model's scores on every model, and its shallow state moves the next-token distribution less than an official NVFP4 checkpoint does. (a) The \examCellsTotal{} paired differences of Table~\ref{tab:exam-main} ($\pm$ one standard error; shaded: $\pm1$ point). (b) The perplexity of the original model's own replies changes by less than \precPplSeedMax\% under the shallow state of the exams, less than under NVIDIA's official NVFP4 checkpoints (Table~\ref{tab:precision-main}; \glm above its full-rank factored path).}
\label{fig:glance}
\end{figure}

\duet is nearly lossless on every model. On \examCountWord{} official exams, \examWithinTwoSEN{} of the \examCellsTotal{} checkpoint--exam cells of the six component sets lie within two standard errors of the original model or above it (Table~\ref{tab:exam-main}, Figure~\ref{fig:glance}a). Its shallow state keeps the next-token distributions closer to the original model's than NVIDIA's official NVFP4 checkpoints do (Figure~\ref{fig:glance}b), and over \clStepsMax{} decode steps its drift grows far more slowly than the number of compressions.

\begin{table*}[t]
\centering\scriptsize
\caption{\duet matches the original model's scores across the \examCountWord{} exams: \examWithinTwoSEN{} of the \examCellsTotal{} cells lie within two standard errors of the original model or above it. Each component set (on \super, its stage-1 set; \S\ref{sec:design-depth}) is paired with its original model on the same questions and per-request seeds. Rows: original score, the set's score and their paired difference (mean $\pm$ standard error over questions). Eight repetitions per exam except LongBench v2 (four), MMLU and RULER (one each).}
\label{tab:exam-main}
\ifreport\setlength{\tabcolsep}{1.5pt}\def\tabexamopen{\begin{tabular*}{\textwidth}{@{\extracolsep{\fill}}lrrrrrrrrrrr@{}}}\def\tabexamclose{\end{tabular*}}%
\fi
\renewcommand{\arraystretch}{1.08}
\tabexamopen
\toprule
 & & & & & & & & & \multicolumn{3}{c@{}}{RULER} \\
\cmidrule(l){10-12}
 & \begin{tabular}[b]{@{}r@{}}GPQA\\Diamond\end{tabular} & \begin{tabular}[b]{@{}r@{}}AIME\\2024+25\end{tabular} & MATH-500 & MMLU & IFEval & \begin{tabular}[b]{@{}r@{}}LiveCode\\Bench v6\end{tabular} & HumanEval+ & \begin{tabular}[b]{@{}r@{}}LongBench\\v2\end{tabular} & 32K & 64K & 128K \\
\midrule
\multicolumn{12}{@{}l}{\textit{\modelFull{} (bf16)}} \\
original & 91.9 & 99.4 & 99.3 & 92.5 & 91.7 & 88.5 & 93.9 & 64.3 & 93.2 & 94.2 & 95.5 \\
\duet & 91.5 & 98.8 & 98.8 & 92.6 & 91.1 & 88.9 & 94.8 & 65.3 & 96.4 & 95.5 & 95.0 \\
difference & $-0.3\!\pm\!0.4$ & $-0.6\!\pm\!0.5$ & $-0.5\!\pm\!0.1$ & $+0.1\!\pm\!0.1$ & $-0.6\!\pm\!0.3$ & $+0.4\!\pm\!0.4$ & $+0.9\!\pm\!0.4$ & $+0.9\!\pm\!0.9$ & $+3.2\!\pm\!0.6$ & $+1.2\!\pm\!0.5$ & $-0.5\!\pm\!0.4$ \\
\midrule
\multicolumn{12}{@{}l}{\textit{\modelnvFull{}}} \\
original & 92.4 & 99.6 & 99.2 & 92.3 & 91.6 & 88.9 & 94.4 & 65.3 & 94.7 & 94.9 & 95.3 \\
\duet & 91.2 & 99.6 & 98.9 & 92.5 & 91.5 & 88.8 & 94.6 & 64.6 & 96.1 & 95.4 & 94.6 \\
difference & $-1.1\!\pm\!0.5$ & $0.0\!\pm\!0.4$ & $-0.3\!\pm\!0.2$ & $+0.2\!\pm\!0.2$ & $-0.1\!\pm\!0.3$ & $-0.1\!\pm\!0.4$ & $+0.2\!\pm\!0.4$ & $-0.7\!\pm\!0.9$ & $+1.4\!\pm\!0.5$ & $+0.5\!\pm\!0.5$ & $-0.7\!\pm\!0.4$ \\
\midrule
\multicolumn{12}{@{}l}{\textit{\lightningFull{}}} \\
original & 76.3 & 89.6 & 98.2 & 88.1 & 91.5 & 78.2 & 91.9 & 47.1 & 96.5 & 96.4 & 95.9 \\
\duet & 75.0 & 88.8 & 98.3 & 88.0 & 92.0 & 77.6 & 92.0 & 45.7 & 96.4 & 96.0 & 95.6 \\
difference & $-1.3\!\pm\!1.0$ & $-0.8\!\pm\!1.1$ & $+0.1\!\pm\!0.2$ & $-0.2\!\pm\!0.2$ & $+0.4\!\pm\!0.4$ & $-0.6\!\pm\!0.6$ & $+0.1\!\pm\!0.6$ & $-1.4\!\pm\!1.1$ & $0.0\!\pm\!0.2$ & $-0.3\!\pm\!0.2$ & $-0.3\!\pm\!0.3$ \\
\midrule
\multicolumn{12}{@{}l}{\textit{\kimi{}}} \\
original & 67.6 & 67.5 & 93.2 & 77.7 & 73.2 & 44.3 & 78.7 & 34.7 & 95.0 & 94.0 & 94.5 \\
\duet & 68.0 & 66.9 & 92.4 & 78.3 & 71.7 & 43.4 & 78.3 & 34.6 & 95.0 & 93.9 & 94.1 \\
difference & $+0.4\!\pm\!1.4$ & $-0.6\!\pm\!2.5$ & $-0.8\!\pm\!0.5$ & $+0.6\!\pm\!0.4$ & $-1.5\!\pm\!0.6$ & $-0.9\!\pm\!0.8$ & $-0.4\!\pm\!0.8$ & $-0.1\!\pm\!1.1$ & $0.0\!\pm\!0.1$ & $-0.1\!\pm\!0.2$ & $-0.4\!\pm\!0.2$ \\
\midrule
\multicolumn{12}{@{}l}{\textit{\glm{}}} \\
original & 91.3 & 96.0 & 98.7 & 93.0 & 89.9 & 87.8 & 94.8 & 63.2 & 96.7 & 96.3 & 96.4 \\
\duet & 91.1 & 98.1 & 98.6 & 92.8 & 90.7 & 88.2 & 94.7 & 64.1 & 96.5 & 96.5 & 96.1 \\
difference & $-0.2\!\pm\!0.5$ & $+2.1\!\pm\!1.3$ & $-0.1\!\pm\!0.2$ & $-0.1\!\pm\!0.1$ & $+0.8\!\pm\!0.4$ & $+0.4\!\pm\!0.4$ & $-0.2\!\pm\!0.3$ & $+0.8\!\pm\!0.8$ & $-0.2\!\pm\!0.1$ & $+0.2\!\pm\!0.2$ & $-0.3\!\pm\!0.2$ \\
\midrule
\multicolumn{12}{@{}l}{\textit{\superFull{}}} \\
original & 89.1 & 98.5 & 99.4 & 92.1 & 91.7 & 86.3 & 95.0 & 59.3 & 96.8 & 94.9 & 96.0 \\
\duet & 90.1 & 98.5 & 99.4 & 92.1 & 92.1 & 86.3 & 95.5 & 58.1 & 96.7 & 94.8 & 95.9 \\
difference & $+1.0\!\pm\!0.6$ & $0.0\!\pm\!0.7$ & $-0.1\!\pm\!0.1$ & $0.0\!\pm\!0.1$ & $+0.5\!\pm\!0.4$ & $0.0\!\pm\!0.4$ & $+0.5\!\pm\!0.4$ & $-1.2\!\pm\!0.8$ & $0.0\!\pm\!0.2$ & $-0.1\!\pm\!0.2$ & $-0.1\!\pm\!0.2$ \\
\bottomrule
\tabexamclose
\end{table*}

\paragraph{Protocol.}
Each component set and its original model answer the same questions with the same per-request sampling seeds (on MMLU, all but at most 0.4\% of samples). Table~\ref{tab:exam-main} gives the results, and Table~\ref{tab:exam} (Appendix~\ref{app:eval}) adds the stage-1 sets that we examined. Both arms are served by SGLang with their recurrent states in fp32. The throughput measurements of Section~\ref{sec:perf} use the same component sets and state precision.

\paragraph{Exams.}
Every exam runs with lm-eval~\cite{lmeval2024} 0.4.13 and its official grader, with each vendor's recommended sampling configuration; GPQA-Diamond and MMLU use the prompt and answer extraction of OpenAI simple-evals~\cite{openai2024simpleevals}, and GPQA-Diamond also its option permutation. The \examCountWord{} exams~\cite{aime2025,gpqa2023,mmlu2021,hendrycks2021math,math500_2023,ifeval2023,livecodebench2024,chen2021evaluating,evalplus2023,longbenchv2_2024,ruler2024} are GPQA-Diamond, MMLU, AIME~2024+25, MATH-500, IFEval, LiveCodeBench~v6 (problems from August 2024 to April 2025), HumanEval+, LongBench~v2, and RULER at 32K, 64K and 128K.

\paragraph{Repetitions and selection.}
Every exam runs eight repetitions except LongBench~v2 (four), MMLU (one) and each RULER length (one; RULER is greedy), always the first $n$ repetitions. No sample is dropped, and no repetition was rerun because of its outcome. Where a model had several candidate sets, the choice between them rested on repetitions 0--3. On the thirty cells of GPQA-Diamond, AIME, MATH-500, IFEval and LiveCodeBench of all six sets, repetitions 4--7, on which no choice rested, agree with the eight-repetition values within about one of their standard errors.

\paragraph{Scores and answers.}
Across the six sets, on unrounded values, \examWithinTwoSEN{} of the \examCellsTotal{} checkpoint--exam cells (Figure~\ref{fig:glance}a) lie within two standard errors of the original model or above it; Appendix~\ref{app:eval} analyses every cell at least one standard error below. Equal scores can hide answer flips~\cite{dutta2024accuracy}, but \duet flips answers about as often as a second sample of the original model does, and its replies are not longer (Appendix~\ref{app:eval-flips}).

\begin{table}[t]
\centering\small
\caption{The shallow state stays closer to the original model than the official NVFP4 checkpoints do, and it composes with 16-bit storage. Top: decode drift (\S\ref{sec:eval-drift}) in nats per token; grey: stored state size as a share of the original fp32 state (shallow state: right after a compression). $^{s}$: stochastic rounding (Table~\ref{tab:precision} gives both roundings). Bottom: change in the perplexity of the same replies. --: no NVFP4 release (\kimi) or an FP8 release, the checkpoint we use (\glm).}
\label{tab:precision-main}
\begin{tabular}{@{}lrrrrr@{}}
\toprule
 & \shortstack[r]{Qwen3.8-\\Flash-Next} & \shortstack[r]{Nemotron-3.5-\\Lightning} & \shortstack[r]{Kimi-\\Linear} & \shortstack[r]{GLM-5.3-\\Flash} & \shortstack[r]{Nemotron-3.5-\\Super} \\
\midrule
\multicolumn{6}{@{}l}{\emph{original model, dense state}} \\
\quad fp32 (exams) & 0\,\makebox[1.9em][r]{\color{nvgrey}\scriptsize 100\%} & 0\,\makebox[1.9em][r]{\color{nvgrey}\scriptsize 100\%} & 0\,\makebox[1.9em][r]{\color{nvgrey}\scriptsize 100\%} & 0\,\makebox[1.9em][r]{\color{nvgrey}\scriptsize 100\%} & 0\,\makebox[1.9em][r]{\color{nvgrey}\scriptsize 100\%} \\
\quad fp16 & 0.0006\,\makebox[1.9em][r]{\color{nvgrey}\scriptsize 50\%} & 0.0004$^{s}$\,\makebox[1.9em][r]{\color{nvgrey}\scriptsize 50\%} & 0.0002\,\makebox[1.9em][r]{\color{nvgrey}\scriptsize 50\%} & 0.0006\,\makebox[1.9em][r]{\color{nvgrey}\scriptsize 50\%} & 0.0002$^{s}$\,\makebox[1.9em][r]{\color{nvgrey}\scriptsize 50\%} \\
\quad fp8 & 0.0073$^{s}$\,\makebox[1.9em][r]{\color{nvgrey}\scriptsize 25\%} & 0.0487\,\makebox[1.9em][r]{\color{nvgrey}\scriptsize 25\%} & 0.0048$^{s}$\,\makebox[1.9em][r]{\color{nvgrey}\scriptsize 25\%} & 0.0055$^{s}$\,\makebox[1.9em][r]{\color{nvgrey}\scriptsize 25\%} & 0.0745\,\makebox[1.9em][r]{\color{nvgrey}\scriptsize 25\%} \\
\multicolumn{6}{@{}l}{\emph{\duet, shallow state}} \\
\quad fp32 (exams) & 0.0056\,\makebox[1.9em][r]{\color{nvgrey}\scriptsize 26\%} & 0.0060\,\makebox[1.9em][r]{\color{nvgrey}\scriptsize 39\%} & 0.0017\,\makebox[1.9em][r]{\color{nvgrey}\scriptsize 51\%} & 0.0027\,\makebox[1.9em][r]{\color{nvgrey}\scriptsize 51\%} & 0.0070\,\makebox[1.9em][r]{\color{nvgrey}\scriptsize 39\%} \\
\quad fp16 & 0.0056\,\makebox[1.9em][r]{\color{nvgrey}\scriptsize 13\%} & 0.0059$^{s}$\,\makebox[1.9em][r]{\color{nvgrey}\scriptsize 20\%} & 0.0016$^{s}$\,\makebox[1.9em][r]{\color{nvgrey}\scriptsize 25\%} & 0.0021\,\makebox[1.9em][r]{\color{nvgrey}\scriptsize 25\%} & 0.0069\,\makebox[1.9em][r]{\color{nvgrey}\scriptsize 20\%} \\
\quad fp8 & 0.0067\,\makebox[1.9em][r]{\color{nvgrey}\scriptsize 6\%} & 0.0136\,\makebox[1.9em][r]{\color{nvgrey}\scriptsize 10\%} & 0.0023\,\makebox[1.9em][r]{\color{nvgrey}\scriptsize 13\%} & 0.0030\,\makebox[1.9em][r]{\color{nvgrey}\scriptsize 13\%} & 0.0160\,\makebox[1.9em][r]{\color{nvgrey}\scriptsize 10\%} \\
\midrule
official NVFP4 checkpoint & 0.0105 & 0.0209 & -- & -- & 0.0190 \\
\midrule
\multicolumn{6}{@{}l}{\emph{perplexity of the original model's replies, change}} \\
\quad \duet, fp32 (exams) & $+$0.46\% & $+$0.46\% & $+$0.10\% & $+$0.21\% & $+$0.55\% \\
\quad official NVFP4 checkpoint & $+$1.16\% & $+$1.94\% & -- & -- & $+$2.05\% \\
\bottomrule
\end{tabular}

\end{table}

\subsection{How Close the Next-Token Distributions Are}
\label{sec:eval-drift}
The shallow state stays closer to the original model's next-token distributions than NVIDIA's official NVFP4 checkpoints do. Exam scores are coarse, so we compare the distributions directly. The \emph{decode drift} is the KL divergence from the original model's next-token distribution, teacher-forced on the original model's own replies to held-out prompts (48 replies in math, science and code; 1{,}024 decode steps), measured above the noise floor of a second run of the original model; the \emph{perplexity change} is the change in perplexity of the same replies (Appendix~\ref{app:precision}; on \glm, 24 replies and 256 steps, above its full-rank factored path). The drift is the primary measure, as it sees every token's full distribution; the perplexity change, given for the exam-tested form and the NVFP4 checkpoints, expresses it in the standard unit of language-model quality. The exams and every drift measurement keep the recurrent states in fp32 except where a lower precision is the variable. Table~\ref{tab:precision-main} gives every storage form at its better rounding; bf16 appears only in Table~\ref{tab:precision}, since fp16 matches or beats it on every model.

\paragraph{The shallow state alone.}
In fp32, as in the exams, the shallow state drifts \precSeedDriftMin--\precSeedDriftMax{} nats per token on the five models and changes the perplexity of the original model's replies by less than \precPplSeedMax\% (here with the exact prefill; \S\ref{sec:depth} measures the shallow prefill). NVIDIA's official NVFP4 checkpoints of \model, \lightning and \super, whose four-bit weights are an accepted serving choice, drift \nvRulerRatioMin--\nvRulerRatioMax{} times as much (\nvRulerKL{} nats per token on \model) and raise the perplexity of the same replies by \nvPplMin--\nvPplMax\%.

\paragraph{Lower precision composes.}
fp16 factors leave the drift within \precFPSixteenMaxChange{} nats per token of fp32 factors on all five models, so right after a compression the shallow state saves a further \precGainMin--\precGainMax$\times$ on top of a 16-bit state (the newest columns stay in fp32 between compressions). Right after a compression, fp16 factors are about as small as a dense fp8 state or smaller and drift less than it on all five models.

Below 16 bits, a dense state must be requantized to fp8 after every token, and on the Mamba-2 models its drift rises to as much as \precFPEightDenseMax{} nats per token at its better rounding. fp8 factors, quantized only at each compression, drift \precFPEightGainOthersMin--\precFPEightGainOthersMax$\times$ less on \lightning, \kimi, \glm and \super and about as much on \model, at \precGainMin--\precGainMax$\times$ fewer bytes right after a compression.

\begin{figure}[t]
\centering
\includegraphics[width=0.85\textwidth]{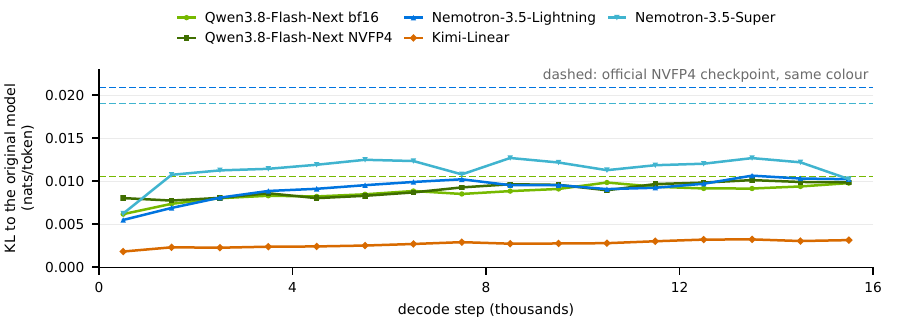}
\caption{The shallow state's drift grows far more slowly than its number of compressions. Decode drift of the released state form (fp32) per 1{,}024 decode steps of long replies of the original model (24 replies on both \model sets and \kimi, 16 on \lightning, 12 on \super; $R=W=16$, 32 on \kimi); dashed: the official NVFP4 checkpoint of the same model over its first 1{,}024 steps (Table~\ref{tab:precision-main}).}
\label{fig:closed-loop}
\end{figure}
\paragraph{Closed-loop drift over long decodes.}
Corollary~\ref{cor:periodic} bounds the error of the open loop, in which the compressed and the exact recurrence see the same inputs; in the closed loop of autoregressive decoding, a compression error also changes later keys and values. We therefore follow the released state form, with fp32 states as in the exams, on both \model sets, \kimi, \lightning and \super over 16K decode steps of long replies, \clComprMin--\clComprMax{} compressions (Figure~\ref{fig:closed-loop}). The drift rises over the first few thousand steps and then grows slowly: from the first to the last 4K steps it grows by \clGrowthMin--\clGrowthMax$\times$ while the number of compressions grows 4$\times$. After 16K steps it is still below the drift of the official NVFP4 checkpoints over their first 1{,}024 steps, on all three models that have one.
\label{endsec:evaluation}

\section{Serving Performance}
\label{sec:perf}

\begin{figure*}[t]
\centering
\includegraphics[width=\textwidth]{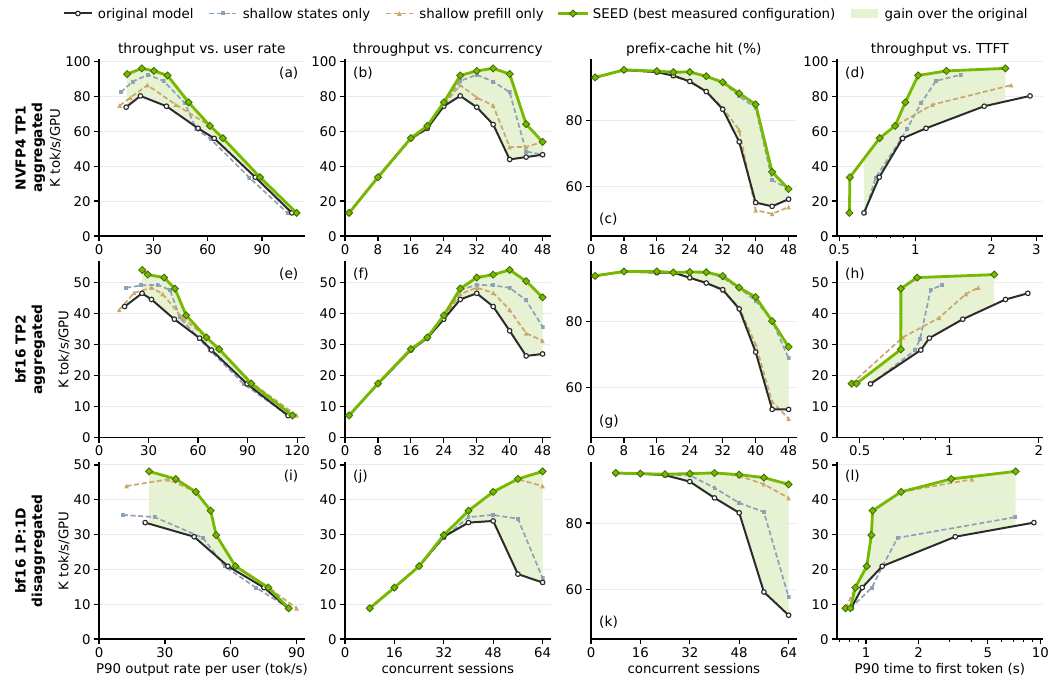}
\caption{\duet{} reaches 1.16--1.42$\times$ the original model's peak throughput per GPU, at 8 to 16 more sessions. Served \model{} with fp32 recurrent states. Rows: aggregated NVFP4 TP1 (a--d), aggregated bf16 TP2 (e--h), disaggregated bf16 1P:1D (i--l). Columns: throughput per GPU against the P90 output rate per user; throughput and prefix-cache hit against concurrency; throughput against the P90 time to first token (Pareto frontiers). Charcoal: the original model; dashed: shallow states alone and shallow prefill alone; green: \duet{}, the best of its three configurations at each concurrency (\S\ref{sec:perf}); shaded: its gain.}
\label{fig:served}
\end{figure*}

\duet's two parts relieve different serving bottlenecks, and together they raise the peak throughput per GPU of served \model to 1.16--1.42$\times$ the original model's (Figure~\ref{fig:served}, Table~\ref{tab:served}). Shallow prefill lowers the prefill's FLOPs. Shallow states shrink the recurrent state, which sets every decode step's state traffic and caps the batch.

\paragraph{Where the savings come from.}
On \model, shallow prefill runs 31 of the 48 layers plus the emitters on every prompt token but the last (33 of 52 on \lightning, \mainKimiDepth{} of 27 on \kimi, 31 of 45 on \glm and 78 of 88 on \super). This lowers the parameter FLOPs per prompt token of \model from \mainPrefillOrig{} to \mainPrefillNew\,GFLOP. For the skipped layers, it also removes the prefill-side reads of their weights and attention caches.

Right after a compression, shallow states reduce the recurrent state of one request from \stateOrigMB{} to \mainStateNewMB\,MB on \model, from \lstateOrigMB{} to \mainLstateNewMB\,MB on \lightning, from \kstateOrigMB{} to \mainKstateNewMB\,MB on \kimi, from \gstateOrigMB{} to \gstateNewMB\,MB on \glm and from 167.8 to 65.5\,MB on \super. Across the five models, the state is \ratioStateMin--\ratioStateMax$\times$ smaller right after a compression and \ratioStateMeanMin--\ratioStateMeanMax$\times$ on average (Table~\ref{\mainMemoryLabel}). A state is read and written at every decode step. In the byte account, a smaller state therefore lowers the per-step state traffic and raises the number of requests that fit in a byte budget. Attention caches remain uncompressed. On these mixture-of-experts checkpoints, the weights dominate a decode step's bytes up to batches of several hundred requests (Appendix~\ref{app:perf-crossover}). The smaller state therefore helps mainly through capacity. Appendix~\ref{app:perf-model} gives the byte and FLOP account from the checkpoints' tensor shapes.

\paragraph{Setup.}
\model is served with SGLang on GB300 GPUs. Aggregated serving uses one instance for both phases: NVFP4 on one GPU (tensor parallelism 1, TP1) or bf16 on two (TP2). Disaggregated bf16 serving uses one two-GPU instance per phase (1P:1D), with 16K-token prefill chunks. Every comparison is paired with the original model on the same hardware. The served component sets are the released \model sets of Section~\ref{sec:eval} (rank $R=\fnR$, code $\fnCode$, state compression every $\fnW$ steps). Both arms keep their recurrent states in fp32, the precision of the models' reference implementations, the default of the serving engine and the precision of the exams; the byte ratio of the shallow and the dense state is the same at any common precision.

\paragraph{Workload and metrics.}
The throughput runs reuse cached prefixes. Each measurement uses the InferenceX (formerly InferenceMAX)~\cite{inferencemax2026} workload, an agentic-coding trace of recorded Claude Code sessions (\texttt{cc-traces-weka-062126-256k}~\cite{cctraces2026}). The AIPerf client~\cite{aiperf2026} replays it in agentic-replay mode for 900 seconds at a fixed concurrency, the number of active sessions. Throughput per GPU counts every output token and every prompt token, whether the prefix cache serves it or the prefill computes it, so a cache hit counts the same in both arms. The P90 end-to-end output rate per user, the rate that 90\% of requests reach, is $1/\mathrm{P90}(\text{end-to-end latency}/\text{output tokens})$. It includes queueing and the time to first token (TTFT). Ratios at an equal per-user rate interpolate each configuration's throughput linearly in this rate between adjacent concurrencies.

\paragraph{Prefix cache.}
In SGLang, a hybrid model's prefix cache keeps a slot for every cached prefix in the same recurrent-state pool that the running requests use~\cite{sglang2024,marconi2025}. When the pool is full, cached slots are evicted, the prefix-cache hit rate falls and the evicted prefixes must be recomputed by the prefill. Smaller states leave room for more cached prefixes and more running requests.

\paragraph{Measurements.}
Each measurement in Figure~\ref{fig:served} is a single 900-second run and includes \duet's own work: the emitters in the prefill and the state compressions in the decode. Up to 16 sessions, where the configurations serve alike, their throughputs agree within 1.3\%. At each concurrency, \duet's curve takes the best of its three measured configurations (shallow states only, shallow prefill only, both) by throughput; in the prefix-cache panels (c, g, k) it takes the highest prefix-cache hit, and in the TTFT panels (d, h, l) the Pareto frontier of all three, the measurements that no other measurement beats in both quantities. This curve lies within 1.1\% of both parts together wherever those were measured; the ratios in the text and in Table~\ref{tab:served} are those of both parts together. The rate panels (a, e, i) draw every measurement with a P90 per-user rate of at least 10 tokens/s, in order of concurrency; the middle columns draw every measurement up to 48 sessions in aggregated and 64 in disaggregated serving.

\paragraph{Throughput.}

Peak to peak, \duet with both parts reaches 1.20$\times$ the original model's throughput per GPU on NVFP4 TP1, 1.16$\times$ on bf16 TP2 and 1.42$\times$ on 1P:1D. At an equal per-user rate of 40 tokens/s (25\,ms per output token, end to end), it delivers 1.23$\times$, 1.26$\times$ and 1.42$\times$ the original model's throughput (Table~\ref{tab:served}). Over per-user rates of 25 to 50 tokens/s, inside the measured range of both arms in all three deployments, the ratio is 1.16--1.26$\times$ in aggregated serving and 1.24--1.49$\times$ in 1P:1D. In aggregated serving, the two peaks come at similar per-user rates: 23.7 tokens/s for \duet and 22.8 for the original model on NVFP4 TP1, and 26.2 and 26.1 on bf16 TP2. In 1P:1D, \duet's peak comes at a far higher per-user rate: 22.8 tokens/s, compared with 9.0 for the original model. Beyond the original model's peak, the gap widens, because its fp32 states fill the state pool in aggregated serving and its prefill instance saturates in 1P:1D. At 40, 44 and 64 sessions, the original model's P90 per-user rate has fallen below 4 tokens/s, and \duet serves 2.1$\times$, 1.9$\times$ and 3.0$\times$ the original model's throughput.

\begin{table}[t]
\centering\small
\setlength{\tabcolsep}{5pt}
\caption{\duet raises the peak throughput per GPU of served \model{} in every deployment (Figure~\ref{fig:served}). Peak: each configuration's highest throughput over its measured concurrencies (tokens/s/GPU), its concurrency and its ratio to the original model's peak. At $C^\star$: the throughput ratio at \duet's peak concurrency (36, 40 and 64 sessions), where the original model is past its own peak (P90 per-user rate 6.8, 6.7 and 1.2 tokens/s). At 40 tok/s: the throughput ratio at an equal P90 per-user rate of 40 tokens/s.}
\label{tab:served}
\colfit{%
\begin{tabular}{@{}lrrrrr@{}}
\toprule
 & \multicolumn{3}{c}{peak} & at $C^\star$ & at 40 tok/s \\
\cmidrule(lr){2-4}\cmidrule(lr){5-5}\cmidrule(l){6-6}
Configuration & sessions & tok/s/GPU & ratio & ratio & ratio \\
\midrule
\multicolumn{6}{@{}l}{\emph{aggregated, NVFP4 TP1}} \\
\quad original & 28 & 80{,}219 & -- & -- & -- \\
\quad shallow states & 32 & 92{,}343 & $1.15\times$ & $1.38\times$ & $1.16\times$ \\
\quad shallow prefill & 28 & 86{,}385 & $1.08\times$ & $1.17\times$ & $1.06\times$ \\
\quad \duet (both) & 36 & 95{,}989 & $1.20\times$ & $1.50\times$ & $1.23\times$ \\
\addlinespace
\multicolumn{6}{@{}l}{\emph{aggregated, bf16 TP2}} \\
\quad original & 32 & 46{,}492 & -- & -- & -- \\
\quad shallow states & 32 & 49{,}235 & $1.06\times$ & $1.40\times$ & $1.18\times$ \\
\quad shallow prefill & 32 & 48{,}301 & $1.04\times$ & $1.20\times$ & $1.11\times$ \\
\quad \duet (both) & 40 & 54{,}047 & $1.16\times$ & $1.57\times$ & $1.26\times$ \\
\addlinespace
\multicolumn{6}{@{}l}{\emph{1P:1D, bf16}} \\
\quad original & 48 & 33{,}906 & -- & -- & -- \\
\quad shallow states & 48 & 35{,}638 & $1.05\times$ & $1.08\times$ & $1.04\times$ \\
\quad shallow prefill & 56 & 45{,}853 & $1.35\times$ & $2.71\times$ & $1.45\times$ \\
\quad \duet (both) & 64 & 48{,}141 & $1.42\times$ & $2.96\times$ & $1.42\times$ \\
\bottomrule
\end{tabular}}
\end{table}

\paragraph{Aggregated serving.}
On GPUs shared by both phases, shallow prefill reduces computation, while shallow states raise request capacity. In aggregated NVFP4 TP1 serving, \duet peaks at 36 sessions, where the original model is past its own peak and serves 6.8 tokens/s per user at P90. At this concurrency, the original model falls to 63.9K tokens/s/GPU, while \duet holds 96.0K. Relative to the original model at the same concurrency, shallow prefill alone raises throughput by 17\%, shallow states alone by 38\% and both together by 50\% (Table~\ref{tab:served}). At 40 sessions, the original model's prefix-cache hit rate has fallen to 55\% and its throughput to 43.8K, while \duet keeps 85\% and 92.8K. At an equal per-user rate of 40 tokens/s, \duet delivers 1.23$\times$ the original model's throughput (89.0K versus 72.3K tokens/s/GPU).

In aggregated bf16 TP2 serving, the original model's throughput collapses once its fp32 states fill the state pool: from 46.5K tokens/s/GPU at 32 sessions to 26.3K at 44. There, its prefix-cache hit rate has fallen to 53\%; \duet's is 80\%. \duet peaks at 54.0K at 40 sessions, 1.57$\times$ the original model's 34.4K at that concurrency. At an equal per-user rate of 40 tokens/s, it delivers 1.26$\times$ the original model's throughput (51.2K versus 40.7K).

\paragraph{Disaggregated serving.}
In bf16 1P:1D serving, the prefill instance is the bottleneck, and shallow prefill relieves it. Peak throughput rises from 33.9K tokens/s/GPU for the original model to 45.9K with shallow prefill alone and 48.1K with \duet, that is, 1.35$\times$ and 1.42$\times$ the original peak. Shallow states alone add 5\% (35.6K; Table~\ref{tab:served}). At equal per-user rates of 25 to 40 tokens/s, \duet delivers 1.42--1.49$\times$ the original model's throughput. At 64 sessions, where the prefill instance is the bottleneck, \duet's smaller states keep more prefixes cached than shallow prefill alone does: its prefix-cache hit rate is 91.7\%, compared with 87.5\%. This adds 9\% (Figure~\ref{fig:served}j), and \duet serves 2.96$\times$ the throughput of the original model, whose hit rate has fallen to 52.2\%. The end-to-end gains come with a lower TTFT, which includes queueing, and a higher prefix-cache hit rate (Figure~\ref{fig:served}).

\paragraph{Low load.}
At low load, the offered load of the replayed trace largely sets the throughput. In aggregated serving, \duet stays within 2\% of the original model's throughput up to 20 sessions and exceeds it by more than 2\% from 24 sessions on. Up to 20 sessions, \duet's P90 output rate per user is 2--3\% above the original model's on bf16 TP2. On NVFP4 TP1, where shallow prefill shortens the time to first token, it lies between $-2\%$ and $+9\%$ of the original model's.
\label{endsec:perf}

\section{Related Work}
\label{sec:related}
\paragraph{Shallow prefill and layer skipping.}
YOCO~\cite{yoco2024} and the causal encoder--decoder of DeepSeek-V4.1-Flash~\cite{deepseekv41} build a shallow prompt stack into new architectures trained from scratch. For trained transformers, SwiftKV~\cite{swiftkv2024} computes the deep keys and values from an earlier layer's output after fine-tuning the deep projections, POP~\cite{pop2026} fills the skipped layers' caches with their original projections of the last kept residual, SPEED~\cite{speed2026} fine-tunes the model to decode without the upper layers' prompt caches, and KV Prediction~\cite{kvprediction2024} predicts the caches with a small auxiliary model. ADEPT~\cite{adept2026} and CALM~\cite{calm2022} exit early per token and fill the skipped layers' keys and values; KV-Direct~\cite{kvdirect2026} recomputes them from a checkpointed residual. For a hybrid model, Puzzle-75B~\cite{puzzle75b2026} prefills with a narrower jointly trained variant that writes the attention caches and the Mamba-2 states. Unlike the new architectures, \duet finds the depth asymmetry in trained models; unlike the methods for transformers, it also writes recurrent states, where an inexact write persists for every later token; and unlike SwiftKV, SPEED and Puzzle-75B, it changes no checkpoint weight. Depth pruning and conversion with distillation~\cite{minitron2024,puzzle2024,mambainllama2024} change the network for every token, and prompt compression~\cite{gist2023,icae2024,llmlingua2023} shortens the prompt itself; \duet keeps every layer and every token in the decode.

\paragraph{Cache and state compression.}
Disaggregated serving ships a prompt's memories between machines~\cite{distserve2024,splitwise2024,mooncake2025}; cache systems page, stream, store and merge them~\cite{vllm2023,sglang2024,cachegen2024,lmcache2025,cacheblend2025}, and Marconi~\cite{marconi2025} brings prefix caching to hybrid models. KV-cache compression~\cite{kivi2024,kvquant2024,snapkv2024,h2o2023} shrinks the output of a full prefill. \duet keeps every attention cache at full precision, so all of these apply to its caches unchanged. For recurrent layers, KVBuffer~\cite{kvbuffer2026}, DeltaLog~\cite{deltalog2026} and ReplaySSM~\cite{replayssm2026} keep the recent writes or layer inputs next to a dense state and fold them in periodically, which cuts the per-step traffic at unchanged outputs but not the state that a request holds; \duet keeps the recent writes next to a rank-$R$ summary and the sink, so the state itself shrinks. \citet{nazari2026rank} analyse the rank of trained linear-attention states and prune their key and query projections, \citet{staterank2026} find both low-rank and high-rank heads in Qwen3-Next, DASC~\cite{dasc2026} keeps the long-horizon heads and channels of stored prefix states, and GHOST~\cite{ghost2026} prunes Mamba-2 state dimensions. \duet keeps every head and factors the active state at run time, around a sink that the update preserves exactly.

\paragraph{Sinks.}
\citet{streamingllm2024} identified the attention sink, \citet{gu2024attentionsink} traced it in part to the softmax normalization, \citet{sun2024massive} linked it to massive activations, \citet{unifiedsinks2026} unify attention and residual sinks, and \citet{gatedattention2025} remove it with a gate in their models. The state sink is its counterpart in recurrent layers: a direction of the state rather than a position in the sequence, which through the head's normalization scales the content as a sink logit does through a softmax denominator. \citet{sinkrec2026} call the steady state of a repeated key in gated delta networks a semantic state sink and mitigate it; Lemma~\ref{prop:decomp} contains this case. \citet{gfssm2024} build a sink into a state-space model, and \citet{mamba2sink2026} use the name ``state sink'' for a $\Delta$ gate that fires on boundary tokens. \citet{loststate2026} find that the final states of a frozen Mamba-130M point in nearly the same direction for unrelated sentences.
\label{endsec:related}

\section{Limitations and Conclusion}
\label{sec:discussion}

\paragraph{Limitations.}
\duet compresses no attention cache. On hybrids whose memory at long context is dominated by their few attention layers, the shallow state's share of the saving shrinks; there \duet composes with KV-cache quantization, eviction and offloading, which act on exactly the memories it leaves untouched. The anchor depth is set per checkpoint from untrained probes rather than learned, and lands at \prefillLayerPctMin--\prefillLayerPctMax\% of the depth on the five models; how shallow a residual stream can determine every deep memory, as a function of architecture and training, is open. The open-loop bound is a worst case set by a head's largest decay, and the measured closed-loop drift grows far more slowly than it allows (\S\ref{sec:eval-drift}); a closed-loop theory of repeated compression would explain why. Our serving gains are measured on mixture-of-experts checkpoints, whose weights dominate a decode step's bytes up to batches of several hundred requests, so the smaller state pays off mainly through capacity (Appendix~\ref{app:perf-crossover}); dense (non-mixture-of-experts) hybrids, with smaller weights, would also see the per-step traffic saving. Three steps follow directly: serving the shallow state in fp16, which keeps its drift (\S\ref{sec:eval-drift}) and halves its bytes again; a prefix cache that stores a prompt's code in place of its deep caches, up to 22\% fewer bytes at bf16 caches (Table~\ref{tab:prefix-forms}); and training beyond 16K-token windows with the periodic compression in the loop.

\paragraph{Conclusion.}
Two properties of trained hybrids locate the compute and HBM that serving them spends beyond what the output requires. Writes are easier than reads, so a prompt's deep memories can be emitted from a shallow anchor layer. A recurrent state is a sink plus a short working memory, a form that the model's own update preserves exactly. \duet turns both into a serving method for frozen checkpoints: it matches the original model's scores on six checkpoints of five production models and raises the throughput per GPU of \model 1.2--1.4$\times$ on an agentic-coding trace. Neither property was designed into these models; both emerge from training, which leaves a hybrid's memories far simpler than the computation and storage that serve them.

\label{end:maintext}

\begingroup\raggedright
\bibliography{refs}

@article{puzzle75b2026,
  title={{Nemotron-Labs-3-Puzzle-75B-A9B}: Compressing Hybrid {MoE} {LLMs}},
  author={Bercovich, Akhiad and Abramovich, Talor and Afrimi, Daniel and Aharon, Shay and Ailon, Nir and Anisimov, Vladimir and Argov, Omer Ullman and Ashkenazi, Maor and Asida, Tomer and Assaf, Nave and Bar Natan, Tomer and Bukharin, Alexander and Chlebus, Grzegorz and Chochowski, Marcin and Chung, Eric and Dabbah, Mohammad and del Mundo, Carlo and Dobrowolska, Ewa and Galil, Ido and Galron, Yaniv and Geifman, Amnon and Geifman, Yonatan and Golan, Izik and Gronskiy, Alex and Grzegorzek, Tomasz and Haber, Netanel and Kadoch, Lior and Karch, Grzegorz and Keren, Tomer and Khattar, Abhinav and Klein, Amir and Konuk, Tugrul and Koren, Roi and Korzekwa, Daniel and Kotek, Shaun and Krommydas, Konstantinos and Levy, Itay and Masad, Ofri and Miron, Yoav and Molchanov, Pavlo and Mor, Shahar and Moshe, Zach and Muralidharan, Saurav and Nabwani, Najeeb and Nushi, Besmira and Patwary, Mostofa and Puny, Omri and Rausch, Johannes and Ronen, Tomer and Sameni, Sepehr and Schen, Itamar and Segal, Elad and Serebrenik, Daniel and Shahaf, Ido and Singhal, Soumye and Sorokin, Daniil and Turuvekere Sreenivas, Sharath and Stepniewska-Dziubinska, Marta and Taghibakhshi, Ali and Tajbakhsh, Nima and Tropp, Oren and Tzur, Dor and Warno, Anna and Wu, Yi-Fu and Zawalski, Michal and Zeng, Jiaqi and Zhang, Yian and Zilberstein, Ran and Zuker, Amit and El-Yaniv, Ran},
  journal={arXiv preprint arXiv:2607.04371},
  year={2026}
}

@article{adept2026,
  title={{ADEPT}: Adaptive Dynamic Early-Exit Process for Transformers},
  author={Yoo, Sangmin and Malla, Srikanth and Choi, Chiho and Lu, Wei D. and Choi, Joon Hee},
  journal={arXiv preprint arXiv:2601.03700},
  year={2026}
}

@inproceedings{pop2026,
  title={{POP}: Prefill-Only Pruning for Efficient Large Model Inference},
  author={He, Junhui and Fu, Zhihui and Wang, Jun and Li, Qingan},
  booktitle={Findings of the Association for Computational Linguistics: ACL 2026},
  pages={5864--5877},
  year={2026}
}

@article{speed2026,
  title={Shallow Prefill, Deep Decoding: Efficient Long-Context Inference via Layer-Asymmetric {KV} Visibility},
  author={Oh, Jungsuk and Jeon, Hyeseo and Ji, Hyunjune and Kong, Kyongmin and Lee, Jay-Yoon},
  journal={arXiv preprint arXiv:2605.06105},
  year={2026}
}

@inproceedings{distserve2024,
  title={{DistServe}: Disaggregating Prefill and Decoding for Goodput-optimized Large Language Model Serving},
  author={Zhong, Yinmin and Liu, Shengyu and Chen, Junda and Hu, Jianbo and Zhu, Yibo and Liu, Xuanzhe and Jin, Xin and Zhang, Hao},
  booktitle={USENIX Symposium on Operating Systems Design and Implementation (OSDI)},
  year={2024}
}

@inproceedings{splitwise2024,
  title={{Splitwise}: Efficient Generative {LLM} Inference Using Phase Splitting},
  author={Patel, Pratyush and Choukse, Esha and Zhang, Chaojie and Shah, Aashaka and Goiri, {\'I}{\~n}igo and Maleki, Saeed and Bianchini, Ricardo},
  booktitle={ACM/IEEE International Symposium on Computer Architecture (ISCA)},
  pages={118--132},
  year={2024}
}

@inproceedings{mooncake2025,
  title={{Mooncake}: Trading More Storage for Less Computation --- A {KVCache}-centric Architecture for Serving {LLM} Chatbot},
  author={Qin, Ruoyu and Li, Zheming and He, Weiran and Cui, Jialei and Ren, Feng and Zhang, Mingxing and Wu, Yongwei and Zheng, Weimin and Xu, Xinran},
  booktitle={USENIX Conference on File and Storage Technologies (FAST)},
  pages={155--170},
  year={2025}
}

@inproceedings{mambainllama2024,
  title={The {Mamba} in the {Llama}: Distilling and Accelerating Hybrid Models},
  author={Wang, Junxiong and Paliotta, Daniele and May, Avner and Rush, Alexander M. and Dao, Tri},
  booktitle={Advances in Neural Information Processing Systems (NeurIPS)},
  year={2024},
  note={arXiv:2408.15237}
}

@inproceedings{puzzle2024,
  title={{Puzzle}: Distillation-Based {NAS} for Inference-Optimized {LLMs}},
  author={Bercovich, Akhiad and Ronen, Tomer and Abramovich, Talor and Ailon, Nir and Assaf, Nave and Dabbah, Mohammed and Galil, Ido and Geifman, Amnon and Geifman, Yonatan and Golan, Izhak and Haber, Netanel and Karpas, Ehud Dov and Koren, Roi and Levy, Itay and Molchanov, Pavlo and Mor, Shahar and Moshe, Zach and Nabwani, Najeeb and Puny, Omri and Rubin, Ran and Schen, Itamar and Shahaf, Ido and Tropp, Oren and Argov, Omer Ullman and Zilberstein, Ran and El-Yaniv, Ran},
  booktitle={International Conference on Machine Learning (ICML)},
  year={2025},
  note={arXiv:2411.19146}
}

@inproceedings{minitron2024,
  title={Compact Language Models via Pruning and Knowledge Distillation},
  author={Muralidharan, Saurav and
    Turuvekere Sreenivas, Sharath and
    Joshi, Raviraj and
    Chochowski, Marcin and
    Patwary, Mostofa and
    Shoeybi, Mohammad and
    Catanzaro, Bryan and
    Kautz, Jan and
    Molchanov, Pavlo},
  booktitle={Advances in Neural Information Processing Systems (NeurIPS)},
  year={2024},
  note={arXiv:2407.14679}
}

@inproceedings{swiftkv2024,
  title={{SwiftKV}: Fast Prefill-Optimized Inference with Knowledge-Preserving Model Transformation},
  author={Qiao, Aurick and Yao, Zhewei and Rajbhandari, Samyam and He, Yuxiong},
  booktitle={Proceedings of the 2025 Conference on Empirical Methods in Natural Language Processing (EMNLP)},
  pages={25734--25753},
  year={2025},
  note={arXiv:2410.03960}
}

@inproceedings{yoco2024,
  title={You Only Cache Once: Decoder-Decoder Architectures for Language Models},
  author={Sun, Yutao and Dong, Li and Zhu, Yi and Huang, Shaohan and Wang, Wenhui and Ma, Shuming and Zhang, Quanlu and Wang, Jianyong and Wei, Furu},
  booktitle={Advances in Neural Information Processing Systems (NeurIPS)},
  year={2024},
  note={arXiv:2405.05254}
}

@techreport{deepseekv41,
  title={{DeepSeek-V4.1-Flash}: Pushing the Limits of {KV} Cache Compression},
  author={{DeepSeek-AI} and
    Xu, Anyi and
    Li, B. and
    Lin, Bangcai and
    Xue, Bing and
    Xian, BingCheng and
    Xu, Bingzheng and
    Wu, Bochao and
    Zhang, Bowei and
    Deng, Boyi and
    Yu, C. C. and
    Jin, Chao and
    Lin, Chaofan and
    Dong, Chen and
    Wang, Chenbing and
    Feng, Chenfan and
    Lu, Chengda and
    Zhao, Chenggang and
    Deng, Chengqi and
    Zhang, Chengyuan and
    Xu, Chenhao and
    Zhao, Chenqi and
    Shao, Chenze and
    Wang, Chuhao and
    Zhang, Chuqi and
    Dai, Damai and
    Yang, Dejian and
    Chen, Deli and
    Huang, Di and
    Wu, Di and
    Li, Donghao and
    Li, Erhang and
    Fu, Eric and
    Zhou, F. and
    Zhou, Fangwei and
    Lin, Fangyun and
    Yuan, Fangzhou and
    Xia, Feiyu and
    Dai, Fucong and
    Hao, Guangbo and
    Li, Guanglin and
    Chen, Guanting and
    Cao, Guoai and
    Fan, Guofan and
    Meng, Guolai and
    Li, Guowei and
    Zhang, Haichuan and
    Ma, Haiyang and
    Shen, Haiyang and
    Li, Han and
    Yu, Han and
    Zhang, Han and
    Deng, Hangyuan and
    Xu, Hanwei and
    Xu, Hanxiang and
    Zhong, Hanxun and
    Guo, Hao and
    Jiang, Hao and
    Li, Hao and
    Qin, Hao and
    Wen, Haodong and
    Liang, Haofen and
    Huang, Haofeng and
    Liu, Haohua and
    Zhang, Haoling and
    Luo, Haoming and
    Yang, Haoran and
    Xu, Haotian and
    Yuan, Haotian and
    Huang, Haoting and
    Luo, Haowen and
    Cai, Haoyang and
    Chen, Haoyu and
    Ji, Haozhe and
    Zhang, Hengran and
    Wang, Hengrui and
    Wu, Hengxu and
    Ding, Honghui and
    Tang, Hongxuan and
    Wang, Huadong and
    Cao, Huanqi and
    Gao, Huazuo and
    Qu, Hui and
    Zeng, Hui and
    Yang, J. and
    Jin, J. H. and
    Zhang, J. H. and
    Zou, J. X. and
    Yu, Jia and
    Zhou, Jiahui and
    Chen, Jiajun and
    Huang, Jialiang and
    Zhao, Jialin and
    Tang, Jiamin and
    Zhou, Jian and
    Tong, Jianan and
    Li, Jianwen and
    Zhu, Jiaqi and
    Wang, Jiarui and
    Ye, Jiasheng and
    Li, Jiashi and
    Xu, Jiaxin and
    Ding, Jiaying and
    Lu, Jibai and
    Hu, Jiewen and
    Yan, Jin and
    Zhai, Jincheng and
    Chen, Jingchang and
    Hu, Jingcheng and
    Zhou, Jingli and
    Xu, Jingsheng and
    Xiang, Jingting and
    Yun, Jingyan and
    Yuan, Jingyang and
    Cheng, Jingyuan and
    Zhu, Jinhua and
    Wang, Jinpeng and
    Chen, Jinyi and
    Hu, Jinyi and
    Yu, Jiping and
    Guo, Jueliang and
    Pei, Junbo and
    Sun, Junbo and
    Jiang, Junguang and
    Qiu, Junjie and
    Zhou, Junkang and
    Liu, Junqi and
    Li, Junren and
    Li, Junxian and
    Song, Junxiao and
    Guo, Junyi and
    Dong, Kai and
    Chen, Kaifeng and
    Gao, Kaige and
    Guan, Kang and
    Yuan, Kangdong and
    Hong, Ke and
    Xu, Ke and
    Zhao, Kefan and
    Ji, Kexin and
    Zhang, Kexin and
    Zhou, Kexing and
    Yu, Kuai and
    Zhang, Lan and
    Wang, Lean and
    Zhang, Lecong and
    Wang, Lei and
    Gao, Letian and
    Zhao, Liang and
    Xu, Liansheng and
    Guo, Lihua and
    Luo, Lingxiao and
    Fu, Lingyue and
    Deng, Litao and
    Wang, Litong and
    Zhang, Liyue and
    Chen, Longhao and
    Chen, Lu and
    Huang, Luotian and
    Ma, Luyao and
    Wang, Luyao and
    Di, M. S. and
    Mei, Max and
    Ye, Menghao and
    Cui, Miao and
    Zhang, Mingchuan and
    Zhang, Minghua and
    Tang, Minghui and
    Zhang, Mingjing and
    Wei, Mingqi and
    Chen, Mingshu and
    Liu, Mingxing and
    Zhou, Mingxu and
    Xu, Mingyu and
    Yang, Mingyu and
    Wang, Mingze and
    Chen, Muyang and
    Shentu, Ni and
    Wang, Ning and
    Ning, Niufang and
    Huang, Panpan and
    Cong, Peixin and
    Wang, Peiyi and
    Xin, Peiyuan and
    Ren, Pengfei and
    Yan, Pengfei and
    Zhang, Pengle and
    Kang, Qi and
    Tang, Qi and
    Wang, Qiancheng and
    Li, Qiang and
    Zhu, Qihao and
    Li, Qingyang and
    Chen, Qinyu and
    Du, Qiushi and
    Guo, Qizhou and
    Xu, Rongxian and
    Ding, Rui and
    Hu, Rui and
    Tian, Rui and
    Yu, Rui and
    Zhu, Ruidong and
    Xu, Ruifan and
    Yang, Ruihan and
    Xia, Ruihang and
    Lu, Ruijie and
    Geng, Ruilin and
    Hong, Ruipeng and
    Ge, Ruiqi and
    Zhang, Ruisong and
    Sun, Ruize and
    Pan, Ruizhe and
    Wang, Runji and
    Chen, Runqian and
    Xu, Runxin and
    Tian, Ruohong and
    Shen, Ruomeng and
    Zhang, Ruoyu and
    X., Ryan and
    Liu, S. H. and
    Lu, Shanghao and
    Zhou, Shangyan and
    Chen, Shanhuang and
    Cai, Shaofei and
    Nie, Shaoheng and
    Chen, Shaoyuan and
    Hu, Shengding and
    Lin, Shengkai and
    Ran, Shengwen and
    Liu, Shengyu and
    Jia, Shengyuan and
    Bai, Shi and
    Feng, Shi and
    Xu, Shicheng and
    Liu, Shichun and
    Hu, Shiqiang and
    Ma, Shirong and
    Wang, Shiyu and
    Feng, Shiyuan and
    Gong, Shufan and
    Lin, Shuhan and
    Yu, Shuiping and
    Zhou, Shunfeng and
    Yang, Shuo and
    Wang, Shuomeng and
    Guo, Shuting and
    Pan, Shuting and
    Yu, Shuying and
    Cao, Sinuo and
    Lin, Siyi and
    Chen, Sizhe and
    Chen, Songyang and
    Zhou, Songyang and
    Ni, Tao and
    Yun, Tao and
    Jin, Tian and
    Pei, Tian and
    Ye, Tian and
    Lin, Tianle and
    Ji, Tianran and
    Cui, Tianyi and
    Yue, Tianyuan and
    Yu, Tingting and
    Xiong, Tongrui and
    Zeng, Wangding and
    Liu, Wei and
    Zhang, Wei and
    Xu, Weibin and
    Zeng, Weihao and
    Zhao, Weilin and
    Liu, Wen and
    Liang, Wenfeng and
    Pang, Wenjie and
    Luo, Wenjing and
    Yao, Wenjing and
    Gao, Wenjun and
    Shao, Wenkai and
    Yang, Wenkai and
    Zhang, Wenli and
    Wang, Wenlu and
    Huang, Wenlve and
    Yan, Wenqian and
    Zhang, Wentao and
    Gao, Xi and
    He, Xiang and
    Li, Xiang and
    Li, Xiangli and
    Wang, Xiangwen and
    Zhang, Xiangying and
    Wei, Xiankui and
    Bi, Xiao and
    Liu, Xiaodong and
    Wang, Xiaohan and
    Qu, Xiaojian and
    Chen, Xiaokang and
    Zhang, Xiaokang and
    Nie, Xiaotao and
    Zou, Xiaoyao and
    Li, Xiaoyuan and
    Guo, Xicheng and
    Chu, Xieting and
    Cheng, Xin and
    Liu, Xin and
    Xie, Xin and
    Xu, Xinbo and
    Liu, Xingchao and
    Liu, Xingchen and
    Yu, Xingkai and
    Li, Xingyou and
    Yao, Xintong and
    Chen, Xinyang and
    Jiang, Xinyong and
    Yang, Xinyu and
    Yang, Xinyu and
    Chen, Xu and
    Wang, Xuanyu and
    Zhong, Xubei and
    Su, Xuecheng and
    Liu, Xuejie and
    Lin, Xuheng and
    Fan, Xujie and
    Zhao, Xuncheng and
    Fu, Xuwei and
    Yan, Y. C. and
    Jiang, Y. H. and
    Wu, Y. T. and
    M., Y. W. and
    Wang, Y. Z. and
    Gao, Yafei and
    Yang, Yang and
    Zhang, Yang and
    Ma, Yanru and
    Huang, Yanwen and
    Li, Yao and
    Li, Yao and
    Meng, Yao and
    Zhao, Yao and
    Sun, Yaofeng and
    Wang, Yaohui and
    Ye, Yaoyang and
    Yin, Yehang and
    Wu, Yexinrui and
    Qian, Yi and
    Tao, Yi and
    Yu, Yi and
    Zhang, Yichao and
    Jiang, Yichen and
    Wang, Yicheng and
    Ding, Yifan and
    Shi, Yifan and
    Peng, Yifeng and
    Zhai, Yifeng and
    Wu, Yijia and
    Xiong, Yiliang and
    Wang, Yilun and
    He, Ying and
    Zhou, Ying and
    Luo, Yingjia and
    Zhong, Yinmin and
    Wang, Yiping and
    Wang, Yisong and
    Zhang, Yixiang and
    Chen, Yixiao and
    Tan, Yixuan and
    Wei, Yixuan and
    Ma, Yiyang and
    Yang, Yiyao and
    Liu, Yiyuan and
    Cai, Yizai and
    Wei, Yizhen and
    Wang, Yizhi and
    Yang, Yonglun and
    Zhuo, Yongqi and
    Guo, Yongqiang and
    Wu, Yongtong and
    Wu, Yu and
    Zhang, Yu and
    Bian, Yuan and
    Cheng, Yuan and
    Ou, Yuan and
    Sun, Yuan and
    Xu, Yuanfan and
    Sun, Yuanhang and
    Li, Yuanhao and
    Liu, Yuchen and
    Yao, Yuchen and
    Han, Yudong and
    Wang, Yuduan and
    Wu, Yuhan and
    Meng, Yuhao and
    Zou, Yuheng and
    Li, YuKun and
    Wang, Yunchuan and
    Xiao, Yunfan and
    Xiong, Yunfan and
    Chen, Yupeng and
    Cao, Yuqian and
    Wang, Yuqian and
    Chen, Yuqing and
    Zhang, Yushun and
    Lin, Yutong and
    Xiao, Yuwei and
    Gu, Yuxian and
    Chen, Yuxiang and
    Huang, Yuxiang and
    Luo, Yuxiang and
    You, Yuxiang and
    Chen, Yuxin and
    Xiang, Yuxin and
    Liu, Yuxuan and
    Zhou, Yuxuan and
    Zhou, Yuyang and
    Guo, Yuzhe and
    Huang, Yuzhen and
    Bai, Yuzhuo and
    Z., Z. Y. and
    Ni, Zanlin and
    Wang, Zehao and
    Zhao, Zehua and
    Ren, Zehui and
    Zhao, Zejun and
    Sha, Zhangli and
    Wang, Zhanying and
    Zhang, Zhaochen and
    Du, Zhaoshuai and
    Fu, Zhe and
    Xu, Zhean and
    Xie, Zhenda and
    Liu, Zheng and
    Zhang, Zhengyan and
    Dong, Zhenhua and
    Hao, Zhewen and
    Wang, Zhibang and
    Gou, Zhibin and
    Ma, Zhicheng and
    Li, Zhihao and
    Shao, Zhihong and
    Huang, Zhihuan and
    Li, Zhijie and
    Lu, Zhirui and
    Huang, Zhixian and
    Chen, Zhixuan and
    Chen, Zhixuan and
    Pan, Zhixuan and
    Wu, Zhiyu and
    Ren, Zhizhou and
    He, Zhu and
    Li, Zhuoshu and
    Zhang, Zhuping and
    Xu, Zian and
    Wang, Zihao and
    Gu, Zihui and
    Zhu, Zijia and
    Zhang, Zili and
    Li, Zilin and
    Hou, Zilong and
    Lyu, Zilong and
    Wang, Ziqiao and
    Xie, Ziwei and
    Zhang, Ziya and
    Gao, Ziyi and
    Pan, Zizheng and
    Li, Zonglin and
    Yao, Zongqing and
    Chen, Zui and
    Wu, Zuofan and
    Ling, Chenchen and
    Hou, Chengyu and
    Chen, Chong and
    Li, D. and
    Qi, Di and
    Ji, Dongjie and
    Wei, Fang and
    Xia, Fanyi and
    Xie, Fei and
    Tan, Feiyi and
    Guo, Hailong and
    Zhai, Haiyan and
    Zhou, Hui and
    Tan, Huihui and
    Li, Huijie and
    Luo, Jia and
    Song, Jia and
    Cai, Jialu and
    Liang, Jian and
    Zhou, Jiangting and
    Gao, Jiaqi and
    Shao, Jiayi and
    Chen, Jie and
    Yang, Jieyu and
    Chen, Jin and
    Zhang, Jingde and
    Zhou, Jingzi and
    Wang, Jinqian and
    Liu, Jinyang and
    Sun, JinZhao and
    Ling, Junhua and
    Zheng, Junmin and
    Yang, Kaicheng and
    Xu, Ke and
    Su, Le and
    Xia, Leyi and
    Ding, Liangfeng and
    Zhuo, Lin and
    Ma, Linwang and
    Zhu, Linyan and
    Cai, Liyu and
    Yao, Luqi and
    Zhang, M. K. and
    Li, Meng and
    Lin, Miao and
    Wang, Miaojun and
    Zhang, Min and
    Li, Mingming and
    Wang, Mingming and
    Yin, Mingze and
    Han, Minmin and
    Cao, Nan and
    Wang, Ning and
    Ma, Ningxin and
    Wang, Panpan and
    Lin, Peihan and
    Sun, Peng and
    Zhang, Peng and
    Ying, Qian and
    Xiang, Qiang and
    Wang, Qiao and
    Mao, Qingmiao and
    Jiang, Qiwei and
    Jin, Rongli and
    Chen, Ruyi and
    Tao, Sha and
    Sun, Shangmian and
    Wu, Shaoqing and
    Zou, Shichao and
    Lei, Si and
    Zhang, Tianyang and
    Sun, Tianyu and
    Yin, Tingting and
    Xiao, W. L. and
    An, Wei and
    Li, Wei and
    Wang, Wei and
    Lin, Weiwei and
    Hou, Wenqing and
    Lin, X. and
    Meng, Xiangfei and
    Huang, Xianzhu and
    Peng, Xiao and
    Li, Xiaoqian and
    Zhang, Xiaoting and
    Sun, Xiaowen and
    Wang, Xiaoxiang and
    Ye, Xiaoyu and
    Zhang, Xinrou and
    Zhang, Xinyu and
    Cao, Xue and
    Chen, Xueyin and
    Zhou, Yanan and
    Xu, Yanhong and
    Xia, Yao and
    Xu, Yao and
    Shao, Yi and
    Zhang, Yihong and
    Ma, Yiling and
    Tang, Ying and
    Lou, Yining and
    Chen, Yiru and
    Piao, Yishi and
    Chen, Yixuan and
    Xiong, Yong and
    Xuan, Yuchen and
    Yang, Yuehan and
    Xu, Yuer and
    Zha, Yukun and
    Ma, Yunxian and
    Lin, Yuping and
    Yan, Yuting and
    Xie, Yutong and
    Sheng, Yuwen and
    Zhu, Yuxuan and
    Zhang, Zekai and
    Ju, Zhe and
    Lin, Zhenzhen and
    Gao, Zheren and
    Sun, Zheyang and
    Yan, Zhigang and
    Wu, Zhongyu and
    Wang, Zi and
    Qu, Zihua and
    Yan, Ziling and
    Wan, Ziyi},
  institution={DeepSeek},
  year={2026},
  note={arXiv:2609.19969}
}

@inproceedings{streamingllm2024,
  title={Efficient Streaming Language Models with Attention Sinks},
  author={Xiao, Guangxuan and Tian, Yuandong and Chen, Beidi and Han, Song and Lewis, Mike},
  booktitle={International Conference on Learning Representations (ICLR)},
  year={2024}
}

@inproceedings{sglang2024,
  title={{SGLang}: Efficient Execution of Structured Language Model Programs},
  author={Zheng, Lianmin and Yin, Liangsheng and Xie, Zhiqiang and Sun, Chuyue and Huang, Jeff and Yu, Cody Hao and Cao, Shiyi and Kozyrakis, Christos and Stoica, Ion and Gonzalez, Joseph E. and Barrett, Clark and Sheng, Ying},
  booktitle={Advances in Neural Information Processing Systems (NeurIPS)},
  year={2024}
}

@inproceedings{vllm2023,
  title={Efficient Memory Management for Large Language Model Serving with {PagedAttention}},
  author={Kwon, Woosuk and Li, Zhuohan and Zhuang, Siyuan and Sheng, Ying and Zheng, Lianmin and Yu, Cody Hao and Gonzalez, Joseph E. and Zhang, Hao and Stoica, Ion},
  booktitle={ACM Symposium on Operating Systems Principles (SOSP)},
  year={2023}
}

@inproceedings{fineweb2024,
  title={The {FineWeb} Datasets: Decanting the Web for the Finest Text Data at Scale},
  author={Penedo, Guilherme and Kydl{\'i}{\v{c}}ek, Hynek and Ben Allal, Loubna and Lozhkov, Anton and Mitchell, Margaret and Raffel, Colin and von Werra, Leandro and Wolf, Thomas},
  booktitle={Advances in Neural Information Processing Systems (NeurIPS) Datasets and Benchmarks},
  year={2024}
}

@inproceedings{gateddeltanet2025,
  title={Gated Delta Networks: Improving {Mamba2} with Delta Rule},
  author={Yang, Songlin and Kautz, Jan and Hatamizadeh, Ali},
  booktitle={International Conference on Learning Representations (ICLR)},
  year={2025}
}

@misc{aime2025,
  title={{AIME} 2024 and 2025 problem sets},
  author={{Mathematical Association of America}},
  howpublished={\url{https://huggingface.co/datasets/HuggingFaceH4/aime_2024}, \url{https://huggingface.co/datasets/math-ai/aime25}},
  year={2025}
}

@inproceedings{mmlu2021,
  title={Measuring Massive Multitask Language Understanding},
  author={Hendrycks, Dan and Burns, Collin and Basart, Steven and Zou, Andy and Mazeika, Mantas and Song, Dawn and Steinhardt, Jacob},
  booktitle={International Conference on Learning Representations (ICLR)},
  year={2021}
}

@inproceedings{cachegen2024,
  title={{CacheGen}: {KV} Cache Compression and Streaming for Fast Large Language Model Serving},
  author={Liu, Yuhan and Li, Hanchen and Cheng, Yihua and Ray, Siddhant and Huang, Yuyang and Zhang, Qizheng and Du, Kuntai and Yao, Jiayi and Lu, Shan and Ananthanarayanan, Ganesh and Maire, Michael and Hoffmann, Henry and Holtzman, Ari and Jiang, Junchen},
  booktitle={ACM SIGCOMM Conference (SIGCOMM)},
  year={2024},
  note={arXiv:2310.07240}
}

@inproceedings{cacheblend2025,
  title={{CacheBlend}: Fast Large Language Model Serving for {RAG} with Cached Knowledge Fusion},
  author={Yao, Jiayi and Li, Hanchen and Liu, Yuhan and Ray, Siddhant and Cheng, Yihua and Zhang, Qizheng and Du, Kuntai and Lu, Shan and Jiang, Junchen},
  booktitle={European Conference on Computer Systems (EuroSys)},
  pages={94--109},
  year={2025},
  note={arXiv:2405.16444}
}

@article{lmcache2025,
  title={{LMCache}: An Efficient {KV} Cache Layer for Enterprise-Scale {LLM} Inference},
  author={Liu, Yuhan and Cheng, Yihua and Yao, Jiayi and An, Yuwei and Chen, Xiaokun and Feng, Shaoting and Huang, Yuyang and Shen, Samuel and Zhang, Rui and Du, Kuntai and Jiang, Junchen},
  journal={arXiv preprint arXiv:2510.09665},
  year={2025}
}

@article{kvprediction2024,
  title={{KV} Prediction for Improved Time to First Token},
  author={Horton, Maxwell and Cao, Qingqing and Sun, Chenfan and Jin, Yanzi and Mehta, Sachin and Rastegari, Mohammad and Nabi, Moin},
  journal={arXiv preprint arXiv:2410.08391},
  year={2024}
}

@inproceedings{gu2024attentionsink,
  title={When Attention Sink Emerges in Language Models: An Empirical View},
  author={Gu, Xiangming and Pang, Tianyu and Du, Chao and Liu, Qian and Zhang, Fengzhuo and Du, Cunxiao and Wang, Ye and Lin, Min},
  booktitle={International Conference on Learning Representations (ICLR)},
  year={2025},
  note={arXiv:2410.10781}
}

@inproceedings{sun2024massive,
  title={Massive Activations in Large Language Models},
  author={Sun, Mingjie and Chen, Xinlei and Kolter, J. Zico and Liu, Zhuang},
  booktitle={Conference on Language Modeling (COLM)},
  year={2024},
  note={arXiv:2402.17762}
}

@inproceedings{gatedattention2025,
  title={Gated Attention for Large Language Models: Non-linearity, Sparsity, and Attention-Sink-Free},
  author={Qiu, Zihan and Wang, Zekun and Zheng, Bo and Huang, Zeyu and Wen, Kaiyue and Yang, Songlin and Men, Rui and Yu, Le and Huang, Fei and Huang, Suozhi and Liu, Dayiheng and Zhou, Jingren and Lin, Junyang},
  booktitle={Advances in Neural Information Processing Systems (NeurIPS)},
  year={2025},
  note={arXiv:2505.06708}
}

@misc{dasc2026,
  title={{DASC}: Decay-Aware State Compression for Hybrid Linear-Attention Serving},
  author={Yu, Yanqi and Sun, Pingwei and Tan, Jianchao and Zhang, Tao and Xie, Yuchen and Cai, Xunliang and Liu, Yao},
  year={2026},
  note={arXiv:2608.30386}
}

@misc{nazari2026rank,
  title={On State Reduction in Linear Attention},
  author={Nazari, Philipp and Rusch, T. Konstantin},
  year={2026},
  note={arXiv:2602.04852}
}

@misc{sinkrec2026,
  author={Zhuang, Zhuang and Wei, Zhipeng and Dai, Ji and Chen, Jie and Pan, Fei and Jiang, Peng and Gai, Kun},
  title={{SinkRec}: Mitigating Semantic State Sink in Long Sequence Recommendation with Memory-Conditioned Gated Delta Networks},
  year={2026},
  note={arXiv:2606.09888}
}

@misc{staterank2026,
  title={State Rank Dynamics in Linear Attention {LLMs}},
  author={Sun, Ao and Zhang, Hongtao and Zhou, Heng and Ma, Yixuan and Qin, Yiran and Su, Tongrui and Liu, Yan and Ma, Zhanyu and Xu, Jun and Gao, Jiuchong and Hao, Jinghua and He, Renqing},
  year={2026},
  note={arXiv:2602.02195}
}

@inproceedings{ghost2026,
  title={{GHOST}: Unmasking Phantom States in {Mamba2} via Grouped Hidden-state Output-aware Selection \& Truncation},
  author={Menezes, Michael and Kyrillidis, Anastasios},
  booktitle={International Conference on Machine Learning (ICML)},
  year={2026},
  note={arXiv:2602.11408}
}

@inproceedings{unifiedsinks2026,
  title={A Unified View of Attention and Residual Sinks: Outlier-Driven Rescaling is Essential for Large Language Model Training},
  author={Qiu, Zihan and Huang, Zeyu and Wen, Kaiyue and Jin, Peng and Zheng, Bo and Zhou, Yuxin and Huang, Haofeng and Wang, Zekun and Li, Xiao and Zhang, Huaqing and Xu, Yang and Lian, Haoran and Zhang, Siqi and Men, Rui and Zhang, Jianwei and Titov, Ivan and Liu, Dayiheng and Zhou, Jingren and Lin, Junyang},
  booktitle={Conference on Language Modeling (COLM)},
  year={2026},
  url={https://openreview.net/forum?id=rwk3hL5lS2},
  note={arXiv:2601.22966}
}

@misc{kvdirect2026,
  author={Qasim, Kaleem Ullah and Zhang, Jiashu and Shaheen, Muhammad Kafeel and Alharith, Razan and Zhang, Heying},
  title={The Residual Stream Is All You Need: On the Redundancy of the {KV} Cache in Transformer Inference},
  year={2026},
  note={arXiv:2603.19664}
}

@inproceedings{gpqa2023,
  title={{GPQA}: A Graduate-Level {Google}-Proof {Q\&A} Benchmark},
  author={Rein, David and Hou, Betty Li and Stickland, Asa Cooper and Petty, Jackson and Pang, Richard Yuanzhe and Dirani, Julien and Michael, Julian and Bowman, Samuel R.},
  booktitle={Conference on Language Modeling (COLM)},
  year={2024},
  note={arXiv:2311.12022}
}

@inproceedings{math500_2023,
  title={Let's Verify Step by Step},
  author={Lightman, Hunter and Kosaraju, Vineet and Burda, Yuri and Edwards, Harrison and Baker, Bowen and Lee, Teddy and Leike, Jan and Schulman, John and Sutskever, Ilya and Cobbe, Karl},
  booktitle={International Conference on Learning Representations (ICLR)},
  year={2024},
  note={arXiv:2305.20050}
}

@article{ifeval2023,
  title={Instruction-Following Evaluation for Large Language Models},
  author={Zhou, Jeffrey and Lu, Tianjian and Mishra, Swaroop and Brahma, Siddhartha and Basu, Sujoy and Luan, Yi and Zhou, Denny and Hou, Le},
  journal={arXiv preprint arXiv:2311.07911},
  year={2023}
}

@inproceedings{ruler2024,
  title={{RULER}: What's the Real Context Size of Your Long-Context Language Models?},
  author={Hsieh, Cheng-Ping and Sun, Simeng and Kriman, Samuel and Acharya, Shantanu and Rekesh, Dima and Jia, Fei and Zhang, Yang and Ginsburg, Boris},
  booktitle={Conference on Language Modeling (COLM)},
  year={2024},
  note={arXiv:2404.06654}
}

@article{mamba2sink2026,
  author={Jiang, Yuhang and Zhang, Bowen},
  title={A Circuit, Not The Circuit: Non-Unique Causal Localisation of the {Mamba-2} State Sink},
  journal={arXiv preprint arXiv:2606.00930},
  year={2026}
}

@article{gfssm2024,
  author={Meng, Tian and Tao, Yang and Yin, Wuliang},
  title={Enhanced Structured State Space Models via Grouped {FIR} Filtering and Attention Sink Mechanisms},
  journal={arXiv preprint arXiv:2408.00244},
  year={2024}
}

@article{loststate2026,
  author={Wagh, Bhagyashree and Singh, Akash},
  title={Lost in State Space: Probing Frozen {Mamba} Representations},
  journal={arXiv preprint arXiv:2605.00253},
  year={2026}
}

@misc{nemotron35super_2026,
  title={{NVIDIA Nemotron 3.5 Super VL 120B-A12B}},
  author={{NVIDIA}},
  howpublished={Model card, \url{https://huggingface.co/nvidia/NVIDIA-Nemotron-3.5-Super-VL-120B-A12B-BF16}},
  year={2026}
}

@misc{nemotron3_2026,
  title={{NVIDIA Nemotron 3.5 Lightning 30B-A3B}},
  author={{NVIDIA}},
  howpublished={Model card, \url{https://huggingface.co/nvidia/NVIDIA-Nemotron-3.5-Lightning-30B-A3B-BF16}},
  year={2026}
}

@inproceedings{mamba2_2024,
  title={Transformers are {SSMs}: Generalized Models and Efficient Algorithms Through Structured State Space Duality},
  author={Dao, Tri and Gu, Albert},
  booktitle={International Conference on Machine Learning (ICML)},
  year={2024}
}

@inproceedings{kivi2024,
  title={{KIVI}: A Tuning-Free Asymmetric 2bit Quantization for {KV} Cache},
  author={Liu, Zirui and Yuan, Jiayi and Jin, Hongye and Zhong, Shaochen and Xu, Zhaozhuo and Braverman, Vladimir and Chen, Beidi and Hu, Xia},
  booktitle={International Conference on Machine Learning (ICML)},
  year={2024},
  note={arXiv:2402.02750}
}

@inproceedings{kvquant2024,
  title={{KVQuant}: Towards 10 Million Context Length {LLM} Inference with {KV} Cache Quantization},
  author={Hooper, Coleman and Kim, Sehoon and Mohammadzadeh, Hiva and Mahoney, Michael W. and Shao, Yakun Sophia and Keutzer, Kurt and Gholami, Amir},
  booktitle={Advances in Neural Information Processing Systems (NeurIPS)},
  year={2024},
  note={arXiv:2401.18079}
}

@inproceedings{snapkv2024,
  title={{SnapKV}: {LLM} Knows What You are Looking for Before Generation},
  author={Li, Yuhong and Huang, Yingbing and Yang, Bowen and Venkitesh, Bharat and Locatelli, Acyr and Ye, Hanchen and Cai, Tianle and Lewis, Patrick and Chen, Deming},
  booktitle={Advances in Neural Information Processing Systems (NeurIPS)},
  year={2024},
  note={arXiv:2404.14469}
}

@inproceedings{h2o2023,
  title={{H$_2$O}: Heavy-Hitter Oracle for Efficient Generative Inference of Large Language Models},
  author={Zhang, Zhenyu and Sheng, Ying and Zhou, Tianyi and Chen, Tianlong and Zheng, Lianmin and Cai, Ruisi and Song, Zhao and Tian, Yuandong and R{\'e}, Christopher and Barrett, Clark and Wang, Zhangyang and Chen, Beidi},
  booktitle={Advances in Neural Information Processing Systems (NeurIPS)},
  year={2023},
  note={arXiv:2306.14048}
}

@inproceedings{gist2023,
  title={Learning to Compress Prompts with Gist Tokens},
  author={Mu, Jesse and Li, Xiang Lisa and Goodman, Noah},
  booktitle={Advances in Neural Information Processing Systems (NeurIPS)},
  year={2023},
  note={arXiv:2304.08467}
}

@inproceedings{icae2024,
  title={In-context Autoencoder for Context Compression in a Large Language Model},
  author={Ge, Tao and Hu, Jing and Wang, Lei and Wang, Xun and Chen, Si-Qing and Wei, Furu},
  booktitle={International Conference on Learning Representations (ICLR)},
  year={2024},
  note={arXiv:2307.06945}
}

@inproceedings{llmlingua2023,
  title={{LLMLingua}: Compressing Prompts for Accelerated Inference of Large Language Models},
  author={Jiang, Huiqiang and Wu, Qianhui and Lin, Chin-Yew and Yang, Yuqing and Qiu, Lili},
  booktitle={Proceedings of the 2023 Conference on Empirical Methods in Natural Language Processing (EMNLP)},
  year={2023},
  note={arXiv:2310.05736}
}

@inproceedings{marconi2025,
  title={{Marconi}: Prefix Caching for the Era of Hybrid {LLMs}},
  author={Pan, Rui and Wang, Zhuang and Jia, Zhen and Karakus, Can and Zancato, Luca and Dao, Tri and Wang, Yida and Netravali, Ravi},
  booktitle={Proceedings of Machine Learning and Systems (MLSys)},
  year={2025},
  note={arXiv:2411.19379}
}

@article{nvfp4_2025,
  title={Pretraining Large Language Models with {NVFP4}},
  author={{NVIDIA} and
    Abecassis, Felix and
    Agrusa, Anjulie and
    Ahn, Dong and
    Alben, Jonah and
    Alborghetti, Stefania and
    Andersch, Michael and
    Arayandi, Sivakumar and
    Bjorlin, Alexis and
    Blakeman, Aaron and
    Briones, Evan and
    Buck, Ian and
    Catanzaro, Bryan and
    Chang, Muya and
    Choi, Jinhang and
    Chrzanowski, Mike and
    Chung, Eric and
    Cui, Victor and
    Dai, Steve and
    Darvish Rouhani, Bita and
    del Mundo, Carlo and
    Donia, Deena and
    Eryilmaz, Burc and
    Estela, Henry and
    Goel, Abhinav and
    Goncharov, Oleg and
    Guvvala, Yugi and
    Hesse, Robert and
    Hewett, Russell and
    Hum, Herbert and
    Kapasi, Ujval and
    Khailany, Brucek and
    Khona, Mikail and
    Knight, Nick and
    Kondratenko, Alex and
    Krashinsky, Ronny and
    Lanir, Ben and
    Layton, Simon and
    Lightstone, Michael and
    Lo, Daniel and
    Micikevicius, Paulius and
    Mishra, Asit and
    Moon, Tim and
    Narayanan, Deepak and
    Ni, Chao and
    Paithankar, Abhijit and
    Pasumarthi, Satish and
    Patel, Ankit and
    Patwary, Mostofa and
    Poojary, Ashwin and
    Prasad, Gargi and
    Priyadarshi, Sweta and
    Qin, Yigong and
    Ren, Xiaowei and
    Rybakov, Oleg and
    Sakr, Charbel and
    Satheesh, Sanjeev and
    Sergienko, Stas and
    Shamis, Pasha and
    Shankar, Kirthi and
    Sharma, Nishant and
    Shoeybi, Mohammad and
    Siu, Michael and
    Smelyanskiy, Misha and
    Stosic, Darko and
    Stosic, Dusan and
    Su, Bor-Yiing and
    Sun, Frank and
    Tajbakhsh, Nima and
    Thomas, Shelby and
    Tredak, Przemek and
    Tsykunov, Evgeny and
    Vaithilingam, Gandhi and
    Vavre, Aditya and
    Venkatesan, Rangharajan and
    Waleffe, Roger and
    Wan, Qiyu and
    Wang, Hexin and
    Wang, Mengdi and
    Wei, Lizzie and
    Wu, Hao and
    Wu, Evan and
    Wyss, Keith and
    Xu, Ning and
    Xue, Jinze and
    Yang, Charlene and
    Zhai, Yujia and
    Zhang, Ruoxi and
    Zhu, Jingyang and
    Zhu, Zhongbo},
  journal={arXiv preprint arXiv:2509.25149},
  year={2025}
}

@article{kimilinear2025,
  title={{Kimi Linear}: An Expressive, Efficient Attention Architecture},
  author={{Kimi Team} and
    Zhang, Yu and
    Lin, Zongyu and
    Yao, Xingcheng and
    Hu, Jiaxi and
    Meng, Fanqing and
    Liu, Chengyin and
    Men, Xin and
    Yang, Songlin and
    Li, Zhiyuan and
    Li, Wentao and
    Lu, Enzhe and
    Liu, Weizhou and
    Chen, Yanru and
    Xu, Weixin and
    Yu, Longhui and
    Wang, Yejie and
    Fan, Yu and
    Zhong, Longguang and
    Yuan, Enming and
    Zhang, Dehao and
    Zhang, Yizhi and
    Liu, T. Y. and
    Wang, Haiming and
    Fang, Shengjun and
    He, Weiran and
    Liu, Shaowei and
    Li, Yiwei and
    Su, Jianlin and
    Qiu, Jiezhong and
    Pang, Bo and
    Yan, Junjie and
    Jiang, Zhejun and
    Huang, Weixiao and
    Yin, Bohong and
    You, Jiacheng and
    Wei, Chu and
    Wang, Zhengtao and
    Hong, Chao and
    Chen, Yutian and
    Chen, Guanduo and
    Wang, Yucheng and
    Zheng, Huabin and
    Wang, Feng and
    Liu, Yibo and
    Dong, Mengnan and
    Zhang, Zheng and
    Pan, Siyuan and
    Wu, Wenhao and
    Wu, Yuhao and
    Guan, Longyu and
    Tao, Jiawen and
    Fu, Guohong and
    Xu, Xinran and
    Wang, Yuzhi and
    Lai, Guokun and
    Wu, Yuxin and
    Zhou, Xinyu and
    Yang, Zhilin and
    Du, Yulun},
  journal={arXiv preprint arXiv:2510.26692},
  year={2025}
}

@article{lmeval2024,
  title={Lessons from the Trenches on Reproducible Evaluation of Language Models},
  author={Biderman, Stella and
    Schoelkopf, Hailey and
    Sutawika, Lintang and
    Gao, Leo and
    Tow, Jonathan and
    Abbasi, Baber and
    Aji, Alham Fikri and
    Ammanamanchi, Pawan Sasanka and
    Black, Sidney and
    Clive, Jordan and
    DiPofi, Anthony and
    Etxaniz, Julen and
    Fattori, Benjamin and
    Forde, Jessica Zosa and
    Foster, Charles and
    Hsu, Jeffrey and
    Jaiswal, Mimansa and
    Lee, Wilson Y. and
    Li, Haonan and
    Lovering, Charles and
    Muennighoff, Niklas and
    Pavlick, Ellie and
    Phang, Jason and
    Skowron, Aviya and
    Tan, Samson and
    Tang, Xiangru and
    Wang, Kevin A. and
    Winata, Genta Indra and
    Yvon, Fran{\c{c}}ois and
    Zou, Andy},
  journal={arXiv preprint arXiv:2405.14782},
  year={2024}
}

@inproceedings{livecodebench2024,
  title     = {{LiveCodeBench}: Holistic and Contamination Free Evaluation of Large Language Models for Code},
  author    = {Jain, Naman and Han, King and Gu, Alex and Li, Wen-Ding and Yan, Fanjia and Zhang, Tianjun and Wang, Sida and Solar-Lezama, Armando and Sen, Koushik and Stoica, Ion},
  booktitle = {International Conference on Learning Representations (ICLR)},
  year      = {2025},
  note      = {arXiv:2403.07974}
}

@inproceedings{agarwal2024gkd,
  title     = {On-Policy Distillation of Language Models: Learning from Self-Generated Mistakes},
  author    = {Agarwal, Rishabh and Vieillard, Nino and Zhou, Yongchao and Stanczyk, Piotr and Ramos, Sabela and Geist, Matthieu and Bachem, Olivier},
  booktitle = {International Conference on Learning Representations (ICLR)},
  year      = {2024}
}

@inproceedings{gu2024minillm,
  title     = {{MiniLLM}: Knowledge Distillation of Large Language Models},
  author    = {Gu, Yuxian and Dong, Li and Wei, Furu and Huang, Minlie},
  booktitle = {International Conference on Learning Representations (ICLR)},
  year      = {2024}
}

@inproceedings{ko2025distillm2,
  title     = {{DistiLLM-2}: A Contrastive Approach Boosts the Distillation of {LLMs}},
  author    = {Ko, Jongwoo and Chen, Tianyi and Kim, Sungnyun and Ding, Tianyu and Liang, Luming and Zharkov, Ilya and Yun, Se-Young},
  booktitle = {International Conference on Machine Learning (ICML)},
  year      = {2025}
}

@inproceedings{smollm2_2025,
  title={{SmolLM2}: When Smol Goes Big -- Data-Centric Training of a Fully Open Small Language Model},
  author={Ben Allal, Loubna and Lozhkov, Anton and Bakouch, Elie and Mart{\'i}n Bl{\'a}zquez, Gabriel and Penedo, Guilherme and Tunstall, Lewis and Marafioti, Andr{\'e}s and Piqueres Lajar{\'i}n, Agust{\'i}n and Kydl{\'i}{\v{c}}ek, Hynek and Srivastav, Vaibhav and Lochner, Joshua and Fahlgren, Caleb and Nguyen, Xuan-Son and Burtenshaw, Ben and Fourrier, Cl{\'e}mentine and Zhao, Haojun and Larcher, Hugo and Morlon, Mathieu and Zakka, Cyril and Raffel, Colin and von Werra, Leandro and Wolf, Thomas},
  booktitle={Conference on Language Modeling (COLM)},
  year={2025},
  note={arXiv:2502.02737}
}

@inproceedings{calm2022,
  title={Confident Adaptive Language Modeling},
  author={Schuster, Tal and Fisch, Adam and Gupta, Jai and Dehghani, Mostafa and Bahri, Dara and Tran, Vinh Q. and Tay, Yi and Metzler, Donald},
  booktitle={Advances in Neural Information Processing Systems (NeurIPS)},
  year={2022}
}

@inproceedings{evalplus2023,
  title={Is Your Code Generated by {ChatGPT} Really Correct? {R}igorous Evaluation of Large Language Models for Code Generation},
  author={Liu, Jiawei and Xia, Chunqiu Steven and Wang, Yuyao and Zhang, Lingming},
  booktitle={Advances in Neural Information Processing Systems (NeurIPS)},
  year={2023}
}

@inproceedings{longbenchv2_2024,
  title={{LongBench v2}: Towards Deeper Understanding and Reasoning on Realistic Long-context Multitasks},
  author={Bai, Yushi and Tu, Shangqing and Zhang, Jiajie and Peng, Hao and Wang, Xiaozhi and Lv, Xin and Cao, Shulin and Xu, Jiazheng and Hou, Lei and Dong, Yuxiao and Tang, Jie and Li, Juanzi},
  booktitle={Proceedings of the 63rd Annual Meeting of the Association for Computational Linguistics (Volume 1: Long Papers)},
  pages={3639--3664},
  year={2025},
  note={arXiv:2412.15204}
}

@misc{inferencemax2026,
  title={{InferenceX} (formerly {InferenceMAX}): Open Source Inference Research Platform},
  author={{SemiAnalysis}},
  year={2026},
  howpublished={\url{https://github.com/SemiAnalysisAI/InferenceX}}
}

@misc{cctraces2026,
  title={{cc-traces-weka-062126-256k}: {WekaTrace} corpus derived from {SemiAnalysis} {Claude Code} proxy traces},
  author={{SemiAnalysis}},
  year={2026},
  howpublished={\url{https://huggingface.co/datasets/semianalysisai/cc-traces-weka-062126-256k}}
}

@misc{aiperf2026,
  title={{AIPerf}},
  author={{NVIDIA}},
  year={2026},
  howpublished={\url{https://github.com/ai-dynamo/aiperf}}
}

@article{elhage2021mathematical,
  title={A Mathematical Framework for Transformer Circuits},
  author={Elhage, Nelson and Nanda, Neel and Olsson, Catherine and Henighan, Tom and Joseph, Nicholas and Mann, Ben and Askell, Amanda and Bai, Yuntao and Chen, Anna and Conerly, Tom and DasSarma, Nova and Drain, Dawn and Ganguli, Deep and Hatfield-Dodds, Zac and Hernandez, Danny and Jones, Andy and Kernion, Jackson and Lovitt, Liane and Ndousse, Kamal and Amodei, Dario and Brown, Tom and Clark, Jack and Kaplan, Jared and McCandlish, Sam and Olah, Chris},
  journal={Transformer Circuits Thread},
  year={2021},
  url={https://transformer-circuits.pub/2021/framework/index.html}
}

@article{qiu2026qwen38next,
  title={On the Design of {Qwen3.8-Next} Architecture: Evaluation, Efficiency, and Training Stability},
  author={Qiu, Zihan and
    Wang, Zekun and
    Li, Xiao and
    Li, Yanpeng and
    Xu, Yang and
    Wang, Yixuan and
    Zhang, Huaqing and
    Men, Rui and
    Mao, Bochao and
    Zhang, Chengruidong and
    Zhou, Fan and
    Luo, Hao and
    Huang, Haofeng and
    Lian, Haoran and
    Huang, Haoyan and
    Chen, Hongqing and
    Zhang, Jianwei and
    Xu, Jing and
    Wang, Junjie and
    Chen, Langshi and
    Wang, Liangyu and
    Jiang, Linlang and
    Yuan, Man and
    Sun, Minmin and
    Jin, Peng and
    Zhang, Siqi and
    Wang, Siyu and
    Ren, Xingzhang and
    Wang, Yakai and
    Zhang, Yi and
    Dong, Yiming and
    Cao, Yizhong and
    Ma, Yubo and
    Mao, Yunfei and
    Zheng, Bo and
    Liu, Dayiheng},
  journal={arXiv preprint arXiv:2608.30320},
  year={2026}
}

@misc{qwen3next2025,
  title={{Qwen3-Next}: Towards Ultimate Training \& Inference Efficiency},
  author={{Qwen Team}},
  year={2025},
  howpublished={Qwen blog, \href{https://qwen.ai/blog?id=4074cca80393150c248e508aa62983f9cb7d27cd}{\nolinkurl{https://qwen.ai/blog?id=4074cca80393150c}\allowbreak\nolinkurl{248e508aa62983f9cb7d27cd}}}
}

@misc{zai2026glm53flash,
  title={{GLM-5.3-Flash}},
  author={{Z.ai}},
  year={2026},
  howpublished={Model card, \url{https://huggingface.co/zai-org/GLM-5.3-Flash}}
}

@article{glm5team2026glm5,
  title={{GLM-5}: From Vibe Coding to Agentic Engineering},
  author={{GLM-5-Team} and
    Zeng, Aohan and
    Lv, Xin and
    Hou, Zhenyu and
    Du, Zhengxiao and
    Zheng, Qinkai and
    Chen, Bin and
    Yin, Da and
    Ge, Chendi and
    Huang, Chenghua and
    Xie, Chengxing and
    Zhu, Chenzheng and
    Yin, Congfeng and
    Wang, Cunxiang and
    Pan, Gengzheng and
    Zeng, Hao and
    Zhang, Haoke and
    Wang, Haoran and
    Chen, Huilong and
    Zhang, Jiajie and
    Jiao, Jian and
    Guo, Jiaqi and
    Wang, Jingsen and
    Du, Jingzhao and
    Wu, Jinzhu and
    Wang, Kedong and
    Li, Lei and
    Fan, Lin and
    Zhong, Lucen and
    Liu, Mingdao and
    Zhao, Mingming and
    Du, Pengfan and
    Dong, Qian and
    Lu, Rui and
    {Shuang-Li} and
    Cao, Shulin and
    Liu, Song and
    Jiang, Ting and
    Chen, Xiaodong and
    Zhang, Xiaohan and
    Huang, Xuancheng and
    Dong, Xuezhen and
    Xu, Yabo and
    Wei, Yao and
    An, Yifan and
    Niu, Yilin and
    Zhu, Yitong and
    Wen, Yuanhao and
    Cen, Yukuo and
    Bai, Yushi and
    Qiao, Zhongpei and
    Wang, Zihan and
    Wang, Zikang and
    Zhu, Zilin and
    Liu, Ziqiang and
    Li, Zixuan and
    Wang, Bojie and
    Wen, Bosi and
    Huang, Can and
    Cai, Changpeng and
    Yu, Chao and
    Li, Chen and
    Hu, Chengwei and
    Zhang, Chenhui and
    Zhang, Dan and
    Lin, Daoyan and
    Yang, Dayong and
    Wang, Di and
    Ai, Ding and
    Zhu, Erle and
    Yi, Fangzhou and
    Chen, Feiyu and
    Wen, Guohong and
    Sun, Hailong and
    Zhao, Haisha and
    Hu, Haiyi and
    Zhang, Hanchen and
    Liu, Hanrui and
    Zhang, Hanyu and
    Peng, Hao and
    Tai, Hao and
    Zhang, Haobo and
    Liu, He and
    Wang, Hongwei and
    Yan, Hongxi and
    Ge, Hongyu and
    Liu, Huan and
    Chu, Huanpeng and
    Zhao, Jia'ni and
    Wang, Jiachen and
    Zhao, Jiajing and
    Ren, Jiamin and
    Wang, Jiapeng and
    Zhang, Jiaxin and
    Gui, Jiayi and
    Zhao, Jiayue and
    Li, Jijie and
    An, Jing and
    Li, Jing and
    Yuan, Jingwei and
    Du, Jinhua and
    Liu, Jinxin and
    Zhi, Junkai and
    Duan, Junwen and
    Zhou, Kaiyue and
    Wei, Kangjian and
    Wang, Ke and
    Luo, Keyun and
    Zhang, Laiqiang and
    Sha, Leigang and
    Xu, Liang and
    Wu, Lindong and
    Ding, Lintao and
    Chen, Lu and
    Li, Minghao and
    Lin, Nianyi and
    Ta, Pan and
    Zou, Qiang and
    Song, Rongjun and
    Yang, Ruiqi and
    Tu, Shangqing and
    Yang, Shangtong and
    Wu, Shaoxiang and
    Zhang, Shengyan and
    Li, Shijie and
    Li, Shuang and
    Fan, Shuyi and
    Qin, Wei and
    Tian, Wei and
    Zhang, Weining and
    Yu, Wenbo and
    Liang, Wenjie and
    Kuang, Xiang and
    Cheng, Xiangmeng and
    Li, Xiangyang and
    Yan, Xiaoquan and
    Hu, Xiaowei and
    Ling, Xiaoying and
    Fan, Xing and
    Xia, Xingye and
    Zhang, Xinyuan and
    Zhang, Xinze and
    Pan, Xirui and
    Zou, Xu and
    Zhang, Xunkai and
    Liu, Yadi and
    Wu, Yandong and
    Li, Yanfu and
    Wang, Yidong and
    Zhu, Yifan and
    Tan, Yijun and
    Zhou, Yilin and
    Pan, Yiming and
    Zhang, Ying and
    Su, Yinpei and
    Geng, Yipeng and
    Yan, Yong and
    Tan, Yonglin and
    Bi, Yuean and
    Shen, Yuhan and
    Yang, Yuhao and
    Li, Yujiang and
    Liu, Yunan and
    Wang, Yunqing and
    Li, Yuntao and
    Wu, Yurong and
    Zhang, Yutao and
    Duan, Yuxi and
    Zhang, Yuxuan and
    Liu, Zezhen and
    Jiang, Zhengtao and
    Yan, Zhenhe and
    Zhang, Zheyu and
    Wei, Zhixiang and
    Chen, Zhuo and
    Feng, Zhuoer and
    Yao, Zijun and
    Chai, Ziwei and
    Wang, Ziyuan and
    Zhang, Zuzhou and
    Xu, Bin and
    Huang, Minlie and
    Wang, Hongning and
    Li, Juanzi and
    Dong, Yuxiao and
    Tang, Jie},
  journal={arXiv preprint arXiv:2602.15763},
  year={2026}
}

@article{nvidia2025nemotronh,
  title={{Nemotron-H}: A Family of Accurate and Efficient Hybrid {Mamba}-{Transformer} Models},
  author={{NVIDIA} and
    Blakeman, Aaron and
    Basant, Aarti and
    Khattar, Abhinav and
    Renduchintala, Adithya and
    Bercovich, Akhiad and
    Ficek, Aleksander and
    Bjorlin, Alexis and
    Taghibakhshi, Ali and
    Deshmukh, Amala Sanjay and
    Mahabaleshwarkar, Ameya Sunil and
    Tao, Andrew and
    Shors, Anna and
    Aithal, Ashwath and
    Poojary, Ashwin and
    Dattagupta, Ayush and
    Buddharaju, Balaram and
    Chen, Bobby and
    Ginsburg, Boris and
    Wang, Boxin and
    Norick, Brandon and
    Butterfield, Brian and
    Catanzaro, Bryan and
    del Mundo, Carlo and
    Dong, Chengyu and
    Harvey, Christine and
    Parisien, Christopher and
    Su, Dan and
    Korzekwa, Daniel and
    Yin, Danny and
    Gitman, Daria and
    Mosallanezhad, David and
    Narayanan, Deepak and
    Fridman, Denys and
    Rekesh, Dima and
    Ma, Ding and
    Pykhtar, Dmytro and
    Ahn, Dong and
    Riach, Duncan and
    Stosic, Dusan and
    Long, Eileen and
    Segal, Elad and
    Evans, Ellie and
    Chung, Eric and
    Galinkin, Erick and
    Bakhturina, Evelina and
    Dobrowolska, Ewa and
    Jia, Fei and
    Liu, Fuxiao and
    Prasad, Gargi and
    Shen, Gerald and
    Liu, Guilin and
    Chen, Guo and
    Qian, Haifeng and
    Ngo, Helen and
    Liu, Hongbin and
    Li, Hui and
    Gitman, Igor and
    Karmanov, Ilia and
    Moshkov, Ivan and
    Golan, Izik and
    Kautz, Jan and
    Polak Scowcroft, Jane and
    Casper, Jared and
    Seppanen, Jarno and
    Lu, Jason and
    Sewall, Jason and
    Zeng, Jiaqi and
    You, Jiaxuan and
    Zhang, Jimmy and
    Zhang, Jing and
    Huang, Jining and
    Xue, Jinze and
    Huang, Jocelyn and
    Conway, Joey and
    Kamalu, John and
    Barker, Jon and
    Cohen, Jonathan and
    Jennings, Joseph and
    Parmar, Jupinder and
    Sapra, Karan and
    Briski, Kari and
    Chumachenko, Kateryna and
    Luna, Katherine and
    Santhanam, Keshav and
    Kong, Kezhi and
    Sivamani, Kirthi and
    Pawelec, Krzysztof and
    Anik, Kumar and
    Li, Kunlun and
    McAfee, Lawrence and
    Derczynski, Leon and
    Pavao, Lindsey and
    Vega, Luis and
    Voegtle, Lukas and
    Bala, Maciej and
    Rodrigues de Melo, Maer and
    Sreedhar, Makesh Narsimhan and
    Chochowski, Marcin and
    Kliegl, Markus and
    Stepniewska-Dziubinska, Marta and
    Le, Matthieu and
    Novikov, Matvei and
    Samadi, Mehrzad and
    Andersch, Michael and
    Evans, Michael and
    Martinez, Miguel and
    Chrzanowski, Mike and
    Ranzinger, Mike and
    Blaz, Mikolaj and
    Smelyanskiy, Misha and
    Fawzy, Mohamed and
    Shoeybi, Mohammad and
    Patwary, Mostofa and
    Lee, Nayeon and
    Tajbakhsh, Nima and
    Xu, Ning and
    Rybakov, Oleg and
    Kuchaiev, Oleksii and
    Delalleau, Olivier and
    Nitski, Osvald and
    Chadha, Parth and
    Shamis, Pasha and
    Micikevicius, Paulius and
    Molchanov, Pavlo and
    Dykas, Peter and
    Fischer, Philipp and
    Aquilanti, Pierre-Yves and
    Bialecki, Piotr and
    Varshney, Prasoon and
    Gundecha, Pritam and
    Tredak, Przemek and
    Karimi, Rabeeh and
    Kandu, Rahul and
    El-Yaniv, Ran and
    Joshi, Raviraj and
    Waleffe, Roger and
    Zhang, Ruoxi and
    Kavanaugh, Sabrina and
    Jain, Sahil and
    Kriman, Samuel and
    Lym, Sangkug and
    Satheesh, Sanjeev and
    Muralidharan, Saurav and
    Narenthiran, Sean and
    Anandaraj, Selvaraj and
    Bak, Seonmyeong and
    Kashirsky, Sergey and
    Han, Seungju and
    Acharya, Shantanu and
    Ghosh, Shaona and
    Turuvekere Sreenivas, Sharath and
    Clay, Sharon and
    Thomas, Shelby and
    Prabhumoye, Shrimai and
    Pachori, Shubham and
    Toshniwal, Shubham and
    Prayaga, Shyamala and
    Jain, Siddhartha and
    Das, Sirshak and
    Kierat, Slawek and
    Majumdar, Somshubra and
    Han, Song and
    Singhal, Soumye and
    Niverty, Sriharsha and
    Alborghetti, Stefania and
    Panguluri, Suseella and
    Bhendigeri, Swetha and
    Akter, Syeda Nahida and
    Migacz, Szymon and
    Shiri, Tal and
    Kong, Terry and
    Roman, Timo and
    Ronen, Tomer and
    Saar, Trisha and
    Konuk, Tugrul and
    Rintamaki, Tuomas and
    Poon, Tyler and
    De, Ushnish and
    Noroozi, Vahid and
    Singh, Varun and
    Korthikanti, Vijay and
    Kurin, Vitaly and
    Ahmad, Wasi Uddin and
    Du, Wei and
    Ping, Wei and
    Dai, Wenliang and
    Byeon, Wonmin and
    Ren, Xiaowei and
    Xu, Yao and
    Choi, Yejin and
    Zhang, Yian and
    Lin, Ying and
    Suhara, Yoshi and
    Yu, Zhiding and
    Li, Zhiqi and
    Li, Zhiyu and
    Zhu, Zhongbo and
    Yang, Zhuolin and
    Chen, Zijia},
  journal={arXiv preprint arXiv:2504.03624},
  year={2025}
}

@article{minimax2025minimax01,
  title={{MiniMax-01}: Scaling Foundation Models with Lightning Attention},
  author={{MiniMax} and
    Li, Aonian and
    Gong, Bangwei and
    Yang, Bo and
    Shan, Boji and
    Liu, Chang and
    Zhu, Cheng and
    Zhang, Chunhao and
    Guo, Congchao and
    Chen, Da and
    Li, Dong and
    Jiao, Enwei and
    Li, Gengxin and
    Zhang, Guojun and
    Sun, Haohai and
    Dong, Houze and
    Zhu, Jiadai and
    Zhuang, Jiaqi and
    Song, Jiayuan and
    Zhu, Jin and
    Han, Jingtao and
    Li, Jingyang and
    Xie, Junbin and
    Xu, Junhao and
    Yan, Junjie and
    Zhang, Kaishun and
    Xiao, Kecheng and
    Kang, Kexi and
    Han, Le and
    Wang, Leyang and
    Yu, Lianfei and
    Feng, Liheng and
    Zheng, Lin and
    Chai, Linbo and
    Xing, Long and
    Ju, Meizhi and
    Chi, Mingyuan and
    Zhang, Mozhi and
    Huang, Peikai and
    Niu, Pengcheng and
    Li, Pengfei and
    Zhao, Pengyu and
    Yang, Qi and
    Xu, Qidi and
    Wang, Qiexiang and
    Wang, Qin and
    Li, Qiuhui and
    Leng, Ruitao and
    Shi, Shengmin and
    Yu, Shuqi and
    Li, Sichen and
    Zhu, Songquan and
    Huang, Tao and
    Liang, Tianrun and
    Sun, Weigao and
    Sun, Weixuan and
    Cheng, Weiyu and
    Li, Wenkai and
    Song, Xiangjun and
    Su, Xiao and
    Han, Xiaodong and
    Zhang, Xinjie and
    Hou, Xinzhu and
    Min, Xu and
    Zou, Xun and
    Shen, Xuyang and
    Gong, Yan and
    Zhu, Yingjie and
    Zhou, Yipeng and
    Zhong, Yiran and
    Hu, Yongyi and
    Fan, Yuanxiang and
    Yu, Yue and
    Yang, Yufeng and
    Li, Yuhao and
    Huang, Yunan and
    Li, Yunji and
    Huang, Yunpeng and
    Xu, Yunzhi and
    Mao, Yuxin and
    Li, Zehan and
    Li, Zekang and
    Tao, Zewei and
    Ying, Zewen and
    Cong, Zhaoyang and
    Qin, Zhen and
    Fan, Zhenhua and
    Yu, Zhihang and
    Jiang, Zhuo and
    Wu, Zijia},
  journal={arXiv preprint arXiv:2501.08313},
  year={2025}
}

@article{deepseek2025v32,
  title={{DeepSeek-V3.2}: Pushing the Frontier of Open Large Language Models},
  author={{DeepSeek-AI} and
    Liu, Aixin and
    Mei, Aoxue and
    Lin, Bangcai and
    Xue, Bing and
    Wang, Bingxuan and
    Xu, Bingzheng and
    Wu, Bochao and
    Zhang, Bowei and
    Lin, Chaofan and
    Dong, Chen and
    Lu, Chengda and
    Zhao, Chenggang and
    Deng, Chengqi and
    Xu, Chenhao and
    Ruan, Chong and
    Dai, Damai and
    Guo, Daya and
    Yang, Dejian and
    Chen, Deli and
    Li, Erhang and
    Zhou, Fangqi and
    Lin, Fangyun and
    Dai, Fucong and
    Hao, Guangbo and
    Chen, Guanting and
    Li, Guowei and
    Zhang, H. and
    Xu, Hanwei and
    Li, Hao and
    Liang, Haofen and
    Wei, Haoran and
    Zhang, Haowei and
    Luo, Haowen and
    Ji, Haozhe and
    Ding, Honghui and
    Tang, Hongxuan and
    Cao, Huanqi and
    Gao, Huazuo and
    Qu, Hui and
    Zeng, Hui and
    Huang, Jialiang and
    Li, Jiashi and
    Xu, Jiaxin and
    Hu, Jiewen and
    Chen, Jingchang and
    Xiang, Jingting and
    Yuan, Jingyang and
    Cheng, Jingyuan and
    Zhu, Jinhua and
    Ran, Jun and
    Jiang, Junguang and
    Qiu, Junjie and
    Li, Junlong and
    Song, Junxiao and
    Dong, Kai and
    Gao, Kaige and
    Guan, Kang and
    Huang, Kexin and
    Zhou, Kexing and
    Huang, Kezhao and
    Yu, Kuai and
    Wang, Lean and
    Zhang, Lecong and
    Wang, Lei and
    Zhao, Liang and
    Yin, Liangsheng and
    Guo, Lihua and
    Luo, Lingxiao and
    Ma, Linwang and
    Wang, Litong and
    Zhang, Liyue and
    Di, M. S. and
    Xu, M. Y and
    Zhang, Mingchuan and
    Zhang, Minghua and
    Tang, Minghui and
    Zhou, Mingxu and
    Huang, Panpan and
    Cong, Peixin and
    Wang, Peiyi and
    Wang, Qiancheng and
    Zhu, Qihao and
    Li, Qingyang and
    Chen, Qinyu and
    Du, Qiushi and
    Xu, Ruiling and
    Ge, Ruiqi and
    Zhang, Ruisong and
    Pan, Ruizhe and
    Wang, Runji and
    Yin, Runqiu and
    Xu, Runxin and
    Shen, Ruomeng and
    Zhang, Ruoyu and
    Liu, S. H. and
    Lu, Shanghao and
    Zhou, Shangyan and
    Chen, Shanhuang and
    Cai, Shaofei and
    Chen, Shaoyuan and
    Hu, Shengding and
    Liu, Shengyu and
    Hu, Shiqiang and
    Ma, Shirong and
    Wang, Shiyu and
    Yu, Shuiping and
    Zhou, Shunfeng and
    Pan, Shuting and
    Zhou, Songyang and
    Ni, Tao and
    Yun, Tao and
    Pei, Tian and
    Ye, Tian and
    Yue, Tianyuan and
    Zeng, Wangding and
    Liu, Wen and
    Liang, Wenfeng and
    Pang, Wenjie and
    Luo, Wenjing and
    Gao, Wenjun and
    Zhang, Wentao and
    Gao, Xi and
    Wang, Xiangwen and
    Bi, Xiao and
    Liu, Xiaodong and
    Wang, Xiaohan and
    Chen, Xiaokang and
    Zhang, Xiaokang and
    Nie, Xiaotao and
    Cheng, Xin and
    Liu, Xin and
    Xie, Xin and
    Liu, Xingchao and
    Yu, Xingkai and
    Li, Xingyou and
    Yang, Xinyu and
    Li, Xinyuan and
    Chen, Xu and
    Su, Xuecheng and
    Pan, Xuehai and
    Lin, Xuheng and
    Fu, Xuwei and
    Wang, Y. Q. and
    Zhang, Yang and
    Xu, Yanhong and
    Ma, Yanru and
    Li, Yao and
    Li, Yao and
    Zhao, Yao and
    Sun, Yaofeng and
    Wang, Yaohui and
    Qian, Yi and
    Yu, Yi and
    Zhang, Yichao and
    Ding, Yifan and
    Shi, Yifan and
    Xiong, Yiliang and
    He, Ying and
    Zhou, Ying and
    Zhong, Yinmin and
    Piao, Yishi and
    Wang, Yisong and
    Chen, Yixiao and
    Tan, Yixuan and
    Wei, Yixuan and
    Ma, Yiyang and
    Liu, Yiyuan and
    Yang, Yonglun and
    Guo, Yongqiang and
    Wu, Yongtong and
    Wu, Yu and
    Cheng, Yuan and
    Ou, Yuan and
    Xu, Yuanfan and
    Wang, Yuduan and
    Gong, Yue and
    Wu, Yuhan and
    Zou, Yuheng and
    Li, Yukun and
    Xiong, Yunfan and
    Luo, Yuxiang and
    You, Yuxiang and
    Liu, Yuxuan and
    Zhou, Yuyang and
    Wu, Z. F. and
    Ren, Z. Z. and
    Zhao, Zehua and
    Ren, Zehui and
    Sha, Zhangli and
    Fu, Zhe and
    Xu, Zhean and
    Xie, Zhenda and
    Zhang, Zhengyan and
    Hao, Zhewen and
    Gou, Zhibin and
    Ma, Zhicheng and
    Yan, Zhigang and
    Shao, Zhihong and
    Huang, Zhixian and
    Wu, Zhiyu and
    Li, Zhuoshu and
    Zhang, Zhuping and
    Xu, Zian and
    Wang, Zihao and
    Gu, Zihui and
    Zhu, Zijia and
    Li, Zilin and
    Zhang, Zipeng and
    Xie, Ziwei and
    Gao, Ziyi and
    Pan, Zizheng and
    Yao, Zongqing and
    Feng, Bei and
    Li, Hui and
    Cai, J. L. and
    Ni, Jiaqi and
    Xu, Lei and
    Li, Meng and
    Tian, Ning and
    Chen, R. J. and
    Jin, R. L. and
    Li, S. S. and
    Zhou, Shuang and
    Sun, Tianyu and
    Li, X. Q. and
    Jin, Xiangyue and
    Shen, Xiaojin and
    Chen, Xiaosha and
    Song, Xinnan and
    Zhou, Xinyi and
    Zhu, Y. X. and
    Huang, Yanping and
    Li, Yaohui and
    Zheng, Yi and
    Zhu, Yuchen and
    Ma, Yunxian and
    Huang, Zhen and
    Xu, Zhipeng and
    Zhang, Zhongyu and
    Ji, Dongjie and
    Liang, Jian and
    Guo, Jianzhong and
    Chen, Jin and
    Xia, Leyi and
    Wang, Miaojun and
    Li, Mingming and
    Zhang, Peng and
    Chen, Ruyi and
    Sun, Shangmian and
    Wu, Shaoqing and
    Ye, Shengfeng and
    Wang, T. and
    Xiao, W. L. and
    An, Wei and
    Wang, Xianzu and
    Sun, Xiaowen and
    Wang, Xiaoxiang and
    Tang, Ying and
    Zha, Yukun and
    Zhang, Zekai and
    Ju, Zhe and
    Zhang, Zhen and
    Qu, Zihua},
  journal={arXiv preprint arXiv:2512.02556},
  year={2025}
}

@article{deepseek2024v2,
  title={{DeepSeek-V2}: A Strong, Economical, and Efficient Mixture-of-Experts Language Model},
  author={{DeepSeek-AI} and
    Liu, Aixin and
    Feng, Bei and
    Wang, Bin and
    Wang, Bingxuan and
    Liu, Bo and
    Zhao, Chenggang and
    Deng, Chengqi and
    Ruan, Chong and
    Dai, Damai and
    Guo, Daya and
    Yang, Dejian and
    Chen, Deli and
    Ji, Dongjie and
    Li, Erhang and
    Lin, Fangyun and
    Luo, Fuli and
    Hao, Guangbo and
    Chen, Guanting and
    Li, Guowei and
    Zhang, H. and
    Xu, Hanwei and
    Yang, Hao and
    Zhang, Haowei and
    Ding, Honghui and
    Xin, Huajian and
    Gao, Huazuo and
    Li, Hui and
    Qu, Hui and
    Cai, J. L. and
    Liang, Jian and
    Guo, Jianzhong and
    Ni, Jiaqi and
    Li, Jiashi and
    Chen, Jin and
    Yuan, Jingyang and
    Qiu, Junjie and
    Song, Junxiao and
    Dong, Kai and
    Gao, Kaige and
    Guan, Kang and
    Wang, Lean and
    Zhang, Lecong and
    Xu, Lei and
    Xia, Leyi and
    Zhao, Liang and
    Zhang, Liyue and
    Li, Meng and
    Wang, Miaojun and
    Zhang, Mingchuan and
    Zhang, Minghua and
    Tang, Minghui and
    Li, Mingming and
    Tian, Ning and
    Huang, Panpan and
    Wang, Peiyi and
    Zhang, Peng and
    Zhu, Qihao and
    Chen, Qinyu and
    Du, Qiushi and
    Chen, R. J. and
    Jin, R. L. and
    Ge, Ruiqi and
    Pan, Ruizhe and
    Xu, Runxin and
    Chen, Ruyi and
    Li, S. S. and
    Lu, Shanghao and
    Zhou, Shangyan and
    Chen, Shanhuang and
    Wu, Shaoqing and
    Ye, Shengfeng and
    Ma, Shirong and
    Wang, Shiyu and
    Zhou, Shuang and
    Yu, Shuiping and
    Zhou, Shunfeng and
    Zheng, Size and
    Wang, T. and
    Pei, Tian and
    Yuan, Tian and
    Sun, Tianyu and
    Xiao, W. L. and
    Zeng, Wangding and
    An, Wei and
    Liu, Wen and
    Liang, Wenfeng and
    Gao, Wenjun and
    Zhang, Wentao and
    Li, X. Q. and
    Jin, Xiangyue and
    Wang, Xianzu and
    Bi, Xiao and
    Liu, Xiaodong and
    Wang, Xiaohan and
    Shen, Xiaojin and
    Chen, Xiaokang and
    Chen, Xiaosha and
    Nie, Xiaotao and
    Sun, Xiaowen and
    Wang, Xiaoxiang and
    Liu, Xin and
    Xie, Xin and
    Yu, Xingkai and
    Song, Xinnan and
    Zhou, Xinyi and
    Yang, Xinyu and
    Lu, Xuan and
    Su, Xuecheng and
    Wu, Y. and
    Li, Y. K. and
    Wei, Y. X. and
    Zhu, Y. X. and
    Xu, Yanhong and
    Huang, Yanping and
    Li, Yao and
    Zhao, Yao and
    Sun, Yaofeng and
    Li, Yaohui and
    Wang, Yaohui and
    Zheng, Yi and
    Zhang, Yichao and
    Xiong, Yiliang and
    Zhao, Yilong and
    He, Ying and
    Tang, Ying and
    Piao, Yishi and
    Dong, Yixin and
    Tan, Yixuan and
    Liu, Yiyuan and
    Wang, Yongji and
    Guo, Yongqiang and
    Zhu, Yuchen and
    Wang, Yuduan and
    Zou, Yuheng and
    Zha, Yukun and
    Ma, Yunxian and
    Yan, Yuting and
    You, Yuxiang and
    Liu, Yuxuan and
    Ren, Z. Z. and
    Ren, Zehui and
    Sha, Zhangli and
    Fu, Zhe and
    Huang, Zhen and
    Zhang, Zhen and
    Xie, Zhenda and
    Hao, Zhewen and
    Shao, Zhihong and
    Wen, Zhiniu and
    Xu, Zhipeng and
    Zhang, Zhongyu and
    Li, Zhuoshu and
    Wang, Zihan and
    Gu, Zihui and
    Li, Zilin and
    Xie, Ziwei},
  journal={arXiv preprint arXiv:2405.04434},
  year={2024}
}

@article{xie2025mhc,
  title={{mHC}: Manifold-Constrained Hyper-Connections},
  author={Xie, Zhenda and Wei, Yixuan and Cao, Huanqi and Zhao, Chenggang and Deng, Chengqi and Li, Jiashi and Dai, Damai and Gao, Huazuo and Chang, Jiang and Yu, Kuai and Zhao, Liang and Zhou, Shangyan and Xu, Zhean and Zhang, Zhengyan and Zeng, Wangding and Hu, Shengding and Wang, Yuqing and Yuan, Jingyang and Wang, Lean and Liang, Wenfeng},
  journal={arXiv preprint arXiv:2512.24880},
  year={2025}
}

@inproceedings{zhu2025hyperconnections,
  title={Hyper-Connections},
  author={Zhu, Defa and Huang, Hongzhi and Huang, Zihao and Zeng, Yutao and Mao, Yunyao and Wu, Banggu and Min, Qiyang and Zhou, Xun},
  booktitle={International Conference on Learning Representations (ICLR)},
  year={2025}
}

@inproceedings{zhang2019rmsnorm,
  title={Root Mean Square Layer Normalization},
  author={Zhang, Biao and Sennrich, Rico},
  booktitle={Advances in Neural Information Processing Systems (NeurIPS)},
  year={2019},
  note={arXiv:1910.07467}
}

@article{bengio2013estimating,
  title={Estimating or Propagating Gradients Through Stochastic Neurons for Conditional Computation},
  author={Bengio, Yoshua and L{\'e}onard, Nicholas and Courville, Aaron},
  journal={arXiv preprint arXiv:1308.3432},
  year={2013}
}

@article{mcnemar1947note,
  title={Note on the sampling error of the difference between correlated proportions or percentages},
  author={McNemar, Quinn},
  journal={Psychometrika},
  volume={12},
  number={2},
  pages={153--157},
  year={1947}
}

@inproceedings{loshchilov2019decoupled,
  title={Decoupled Weight Decay Regularization},
  author={Loshchilov, Ilya and Hutter, Frank},
  booktitle={International Conference on Learning Representations (ICLR)},
  year={2019},
  note={arXiv:1711.05101}
}

@misc{openai2024simpleevals,
  title={simple-evals},
  author={{OpenAI}},
  year={2024},
  howpublished={\url{https://github.com/openai/simple-evals}}
}

@misc{kydlicek2025mathverify,
  title={{Math-Verify}: Math Verification Library},
  author={Kydl{\'i}{\v{c}}ek, Hynek},
  year={2025},
  howpublished={\url{https://github.com/huggingface/math-verify}}
}

@inproceedings{hendrycks2021math,
  title={Measuring Mathematical Problem Solving With the {MATH} Dataset},
  author={Hendrycks, Dan and Burns, Collin and Kadavath, Saurav and Arora, Akul and Basart, Steven and Tang, Eric and Song, Dawn and Steinhardt, Jacob},
  booktitle={Proceedings of the Neural Information Processing Systems Track on Datasets and Benchmarks},
  year={2021},
  note={arXiv:2103.03874}
}

@article{chen2021evaluating,
  title={Evaluating Large Language Models Trained on Code},
  author={Chen, Mark and
    Tworek, Jerry and
    Jun, Heewoo and
    Yuan, Qiming and
    Ponde de Oliveira Pinto, Henrique and
    Kaplan, Jared and
    Edwards, Harri and
    Burda, Yuri and
    Joseph, Nicholas and
    Brockman, Greg and
    Ray, Alex and
    Puri, Raul and
    Krueger, Gretchen and
    Petrov, Michael and
    Khlaaf, Heidy and
    Sastry, Girish and
    Mishkin, Pamela and
    Chan, Brooke and
    Gray, Scott and
    Ryder, Nick and
    Pavlov, Mikhail and
    Power, Alethea and
    Kaiser, Lukasz and
    Bavarian, Mohammad and
    Winter, Clemens and
    Tillet, Philippe and
    Petroski Such, Felipe and
    Cummings, Dave and
    Plappert, Matthias and
    Chantzis, Fotios and
    Barnes, Elizabeth and
    Herbert-Voss, Ariel and
    Guss, William Hebgen and
    Nichol, Alex and
    Paino, Alex and
    Tezak, Nikolas and
    Tang, Jie and
    Babuschkin, Igor and
    Balaji, Suchir and
    Jain, Shantanu and
    Saunders, William and
    Hesse, Christopher and
    Carr, Andrew N. and
    Leike, Jan and
    Achiam, Josh and
    Misra, Vedant and
    Morikawa, Evan and
    Radford, Alec and
    Knight, Matthew and
    Brundage, Miles and
    Murati, Mira and
    Mayer, Katie and
    Welinder, Peter and
    McGrew, Bob and
    Amodei, Dario and
    McCandlish, Sam and
    Sutskever, Ilya and
    Zaremba, Wojciech},
  journal={arXiv preprint arXiv:2107.03374},
  year={2021}
}

@inproceedings{pyatkin2025generalizing,
  title={Generalizing Verifiable Instruction Following},
  author={Pyatkin, Valentina and Malik, Saumya and Graf, Victoria and Ivison, Hamish and Huang, Shengyi and Dasigi, Pradeep and Lambert, Nathan and Hajishirzi, Hannaneh},
  booktitle={Advances in Neural Information Processing Systems (NeurIPS) Datasets and Benchmarks},
  year={2025},
  note={arXiv:2507.02833}
}

@article{moshkov2025aimo2,
  title={{AIMO-2} Winning Solution: Building State-of-the-Art Mathematical Reasoning Models with {OpenMathReasoning} dataset},
  author={Moshkov, Ivan and Hanley, Darragh and Sorokin, Ivan and Toshniwal, Shubham and Henkel, Christof and Schifferer, Benedikt and Du, Wei and Gitman, Igor},
  journal={arXiv preprint arXiv:2504.16891},
  year={2025}
}

@inproceedings{ahmad2025opencodereasoning,
  title={{OpenCodeReasoning}: Advancing Data Distillation for Competitive Coding},
  author={Ahmad, Wasi Uddin and Narenthiran, Sean and Majumdar, Somshubra and Ficek, Aleksander and Jain, Siddhartha and Huang, Jocelyn and Noroozi, Vahid and Ginsburg, Boris},
  booktitle={Conference on Language Modeling (COLM)},
  year={2025},
  note={arXiv:2504.01943}
}

@inproceedings{guha2026openthoughts,
  title={{OpenThoughts}: Data Recipes for Reasoning Models},
  author={Guha, Etash and
    Marten, Ryan and
    Keh, Sedrick and
    Raoof, Negin and
    Smyrnis, Georgios and
    Bansal, Hritik and
    Nezhurina, Marianna and
    Mercat, Jean and
    Vu, Trung and
    Sprague, Zayne and
    Suvarna, Ashima and
    Feuer, Benjamin and
    Chen, Liangyu and
    Khan, Zaid and
    Frankel, Eric and
    Grover, Sachin and
    Choi, Caroline and
    Muennighoff, Niklas and
    Su, Shiye and
    Zhao, Wanjia and
    Yang, John and
    Pimpalgaonkar, Shreyas and
    Sharma, Kartik and
    Ji, Charlie Cheng-Jie and
    Deng, Yichuan and
    Pratt, Sarah and
    Ramanujan, Vivek and
    Saad-Falcon, Jon and
    Acharya, Stutee and
    Li, Jeffrey and
    Dave, Achal and
    Albalak, Alon and
    Arora, Kushal and
    Wulfe, Blake and
    Hegde, Chinmay and
    Durrett, Greg and
    Oh, Sewoong and
    Bansal, Mohit and
    Gabriel, Saadia and
    Grover, Aditya and
    Chang, Kai-Wei and
    Shankar, Vaishaal and
    Gokaslan, Aaron and
    Merrill, Mike A. and
    Hashimoto, Tatsunori and
    Choi, Yejin and
    Jitsev, Jenia and
    Heckel, Reinhard and
    Sathiamoorthy, Maheswaran and
    Dimakis, Alexandros G. and
    Schmidt, Ludwig},
  booktitle={International Conference on Learning Representations (ICLR)},
  year={2026},
  note={arXiv:2506.04178}
}

@inproceedings{lambert2025tulu3,
  title={{T{\"u}lu 3}: Pushing Frontiers in Open Language Model Post-Training},
  author={Lambert, Nathan and Morrison, Jacob and Pyatkin, Valentina and Huang, Shengyi and Ivison, Hamish and Brahman, Faeze and Miranda, Lester James V. and Liu, Alisa and Dziri, Nouha and Lyu, Shane and Gu, Yuling and Malik, Saumya and Graf, Victoria and Hwang, Jena D. and Yang, Jiangjiang and Le Bras, Ronan and Tafjord, Oyvind and Wilhelm, Chris and Soldaini, Luca and Smith, Noah A. and Wang, Yizhong and Dasigi, Pradeep and Hajishirzi, Hannaneh},
  booktitle={Conference on Language Modeling (COLM)},
  year={2025},
  note={arXiv:2411.15124}
}

@misc{openr1,
  title={Open {R1}: A fully open reproduction of {DeepSeek-R1}},
  author={{Hugging Face}},
  year={2025},
  howpublished={\url{https://github.com/huggingface/open-r1}}
}

@misc{nvidia2025osr2,
  title={{OpenScienceReasoning-2}},
  author={{NVIDIA}},
  year={2025},
  howpublished={Dataset card, \url{https://huggingface.co/datasets/nvidia/OpenScienceReasoning-2}}
}

@misc{logitlens2020,
  title={Interpreting {GPT}: The Logit Lens},
  author={{nostalgebraist}},
  year={2020},
  howpublished={LessWrong, \url{https://www.lesswrong.com/posts/AcKRB8wDpdaN6v6ru/interpreting-gpt-the-logit-lens}}
}

@article{qwen3_2025,
  title={{Qwen3} Technical Report},
  author={Yang, An and
    Li, Anfeng and
    Yang, Baosong and
    Zhang, Beichen and
    Hui, Binyuan and
    Zheng, Bo and
    Yu, Bowen and
    Gao, Chang and
    Huang, Chengen and
    Lv, Chenxu and
    Zheng, Chujie and
    Liu, Dayiheng and
    Zhou, Fan and
    Huang, Fei and
    Hu, Feng and
    Ge, Hao and
    Wei, Haoran and
    Lin, Huan and
    Tang, Jialong and
    Yang, Jian and
    Tu, Jianhong and
    Zhang, Jianwei and
    Yang, Jianxin and
    Yang, Jiaxi and
    Zhou, Jing and
    Zhou, Jingren and
    Lin, Junyang and
    Dang, Kai and
    Bao, Keqin and
    Yang, Kexin and
    Yu, Le and
    Deng, Lianghao and
    Li, Mei and
    Xue, Mingfeng and
    Li, Mingze and
    Zhang, Pei and
    Wang, Peng and
    Zhu, Qin and
    Men, Rui and
    Gao, Ruize and
    Liu, Shixuan and
    Luo, Shuang and
    Li, Tianhao and
    Tang, Tianyi and
    Yin, Wenbiao and
    Ren, Xingzhang and
    Wang, Xinyu and
    Zhang, Xinyu and
    Ren, Xuancheng and
    Fan, Yang and
    Su, Yang and
    Zhang, Yichang and
    Zhang, Yinger and
    Wan, Yu and
    Liu, Yuqiong and
    Wang, Zekun and
    Cui, Zeyu and
    Zhang, Zhenru and
    Zhou, Zhipeng and
    Qiu, Zihan},
  journal={arXiv preprint arXiv:2505.09388},
  year={2025}
}

@article{gsm8k2021,
  title={Training Verifiers to Solve Math Word Problems},
  author={Cobbe, Karl and Kosaraju, Vineet and Bavarian, Mohammad and Chen, Mark and Jun, Heewoo and Kaiser, Lukasz and Plappert, Matthias and Tworek, Jerry and Hilton, Jacob and Nakano, Reiichiro and Hesse, Christopher and Schulman, John},
  journal={arXiv preprint arXiv:2110.14168},
  year={2021}
}

@inproceedings{hellaswag2019,
  title={{HellaSwag}: Can a Machine Really Finish Your Sentence?},
  author={Zellers, Rowan and Holtzman, Ari and Bisk, Yonatan and Farhadi, Ali and Choi, Yejin},
  booktitle={Proceedings of the 57th Annual Meeting of the Association for Computational Linguistics},
  pages={4791--4800},
  year={2019},
  note={arXiv:1905.07830}
}

@article{arc2018,
  title={Think you have Solved Question Answering? {Try} {ARC}, the {AI2} Reasoning Challenge},
  author={Clark, Peter and Cowhey, Isaac and Etzioni, Oren and Khot, Tushar and Sabharwal, Ashish and Schoenick, Carissa and Tafjord, Oyvind},
  journal={arXiv preprint arXiv:1803.05457},
  year={2018}
}

@inproceedings{winogrande2020,
  title={{WinoGrande}: An Adversarial {Winograd} Schema Challenge at Scale},
  author={Sakaguchi, Keisuke and Le Bras, Ronan and Bhagavatula, Chandra and Choi, Yejin},
  booktitle={Proceedings of the AAAI Conference on Artificial Intelligence (AAAI)},
  pages={8732--8740},
  year={2020},
  note={arXiv:1907.10641}
}

@misc{qwen35_2026,
  title={{Qwen3.5-4B-Base} and {Qwen3.5-9B-Base}},
  author={{Qwen Team}},
  year={2026},
  howpublished={Model cards, \url{https://huggingface.co/Qwen/Qwen3.5-4B-Base}, \url{https://huggingface.co/Qwen/Qwen3.5-9B-Base}}
}

@misc{glm46v2025,
  title={{GLM-4.6V-Flash}},
  author={{Z.ai}},
  year={2025},
  howpublished={Model card, \url{https://huggingface.co/zai-org/GLM-4.6V-Flash}}
}

@article{kvbuffer2026,
  title={{KVBuffer}: {IO}-aware Serving for Linear Attention},
  author={Zou, Longwei and Zhong, Lin},
  journal={arXiv preprint arXiv:2605.19049},
  year={2026}
}

@article{deltalog2026,
  title={{DeltaLog}: Deferred Materialization of Recurrent States for Linear Attention Decoding},
  author={Lin, Junqing and Sun, Jingwei and Sun, Guangzhong},
  journal={arXiv preprint arXiv:2608.15533},
  year={2026}
}

@article{replayssm2026,
  title={{ReplaySSM}: Cache {SSM} Inputs, Not State},
  author={Liou, Ze-Wei and Nover, Harris and Naik, Amey and Skliar, Andrii and Shovkun, Igor and Shahaf, Ido and Dao, Tri},
  journal={arXiv preprint arXiv:2610.13778},
  year={2026}
}

@inproceedings{dutta2024accuracy,
  title={Accuracy is Not All You Need},
  author={Dutta, Abhinav and Krishnan, Sanjeev and Kwatra, Nipun and Ramjee, Ramachandran},
  booktitle={Advances in Neural Information Processing Systems (NeurIPS)},
  year={2024},
  note={arXiv:2407.09141}
}
\endgroup
\bibliographystyle{seedreport}

\appendix
\renewcommand{\topfraction}{0.95}
\renewcommand{\bottomfraction}{0.9}
\renewcommand{\textfraction}{0.05}
\renewcommand{\floatpagefraction}{0.85}
\renewcommand{\dbltopfraction}{0.95}
\renewcommand{\dblfloatpagefraction}{0.85}
\setcounter{topnumber}{4}
\setcounter{bottomnumber}{4}
\setcounter{totalnumber}{6}
\setcounter{dbltopnumber}{4}
\setlength{\floatsep}{8pt plus 2pt minus 2pt}
\setlength{\textfloatsep}{10pt plus 2pt minus 2pt}
\setlength{\intextsep}{8pt plus 2pt minus 2pt}
\setlength{\dblfloatsep}{8pt plus 2pt minus 2pt}
\setlength{\dbltextfloatsep}{10pt plus 2pt minus 2pt}
\makeatletter
\setlength{\@fptop}{0pt}
\setlength{\@fpsep}{8pt plus 2pt minus 2pt}
\setlength{\@fpbot}{0pt plus 1fil}
\setlength{\@dblfptop}{0pt}
\setlength{\@dblfpsep}{8pt plus 2pt minus 2pt}
\setlength{\@dblfpbot}{0pt plus 1fil}
\makeatother

\makeatletter
\newenvironment{appendixwide}[1]{%
  \ifreport\def\appwidetype{#1}\begin{#1}[htbp]\else\begin{strip}\def\@captype{#1}\fi
  \centering
}{%
  \ifreport\expandafter\end\expandafter{\appwidetype}\else\par\end{strip}\fi
}
\makeatother

\makeatletter
\newenvironment{appendixcolumn}[1]{%
  \def\appendixfloattype{#1}%
  \csname #1\endcsname[htbp]%
}{\csname end\appendixfloattype\endcsname}
\makeatother
\setlength{\stripsep}{6pt plus 1pt minus 1pt}

\ifreport
\fontsize{11}{12}\selectfont
\setlength{\parskip}{0pt plus 1pt}
\makeatletter
\renewcommand{\scriptsize}{\@setfontsize\scriptsize{8}{9}}
\renewcommand{\small}{\@setfontsize\small{10}{11}}
\captionsetup{font=appendixcompact,skip=4pt}
\makeatother
\fi

\makeatletter
\patchcmd{\@addviper}{\csname hold@viper@hook\endcsname}{%
  \check@floats\cuted@@tempdim@a
  \@reqcolroom\dimexpr\ht@viper+\ht\@viper+\dp\@viper+2\stripsep+\cuted@@tempdim@a\relax
  \csname hold@viper@hook\endcsname
}{}{\errmessage{Appendix strip-fit patch did not match}}
\newcommand{\appendixDeferWideBlocks}{%
\patchcmd{\@addviper}{\setbox\@tempboxa\vsplit\@viper to\@colht}{%
  \setbox\@tempboxa\vbox{}%
}{}{\errmessage{Appendix strip-defer patch did not match}}%
}
\makeatother

\appendixMemoryTable
\section{Prefill Depth: Measurements}
\label{app:depth}
This appendix gives the protocols behind \S\ref{sec:depth}. Appendix~\ref{app:rule} turns the measurements into the anchor depth, the code and the state rank.

\ifreport\suppressfloats[t]\fi %
\paragraph{Protocol.}
Unless stated otherwise, a measurement uses 48 held-out 4K windows of web text, each a prompt of about 3.5K tokens followed by 256 teacher-forced continuation tokens. It is scored by the per-token KL divergence $\mathrm{KL}(p\,\|\,\hat p)$ between the original model's next-token distribution $p$ and the modified model's $\hat p$, or by the change in next-token cross-entropy. The KLs of the layer skips, the untrained shallow prefill and the trained sets, and the decode drift of \model, \lightning and \kimi, sum over the original model's 1{,}024 most likely tokens and one bucket for the rest. This is a lower bound on the full KL and close to it: 0.079 against 0.081 for the untrained shallow prefill on \model, 0.093 against 0.094 on \lightning. The early exit's KL sums over the full vocabulary. In the \emph{untrained shallow prefill}, every memory-holding layer at index $k$ or above writes its memory through its own untrained write path from the exact $h^{(k)}$, without a code; its increase in cross-entropy is the \emph{untrained gap} $\mathrm{gap}(k)$ (Figure~\ref{fig:gap}).

\paragraph{Phase-specific layer skips.}
A prompt-side skip removes layer $\ell$'s output for the prompt tokens while the layer still writes its memory; a decode-side skip removes it for the continuation tokens, after an exact prompt. The skips sample every third layer of \model, \lightning and \glm, every fourth of \super and every layer of \kimi, and score continuation steps 1--255 (\glm: 16 windows, steps 1--63). The early exit applies the output head to $h^{(k)}$ on 64 windows (\glm: 32). \super's engine is not deterministic; its values include a run-to-run floor of 0.005.

\ifreport
\begin{figure}[t]
\centering
\includegraphics[width=\textwidth]{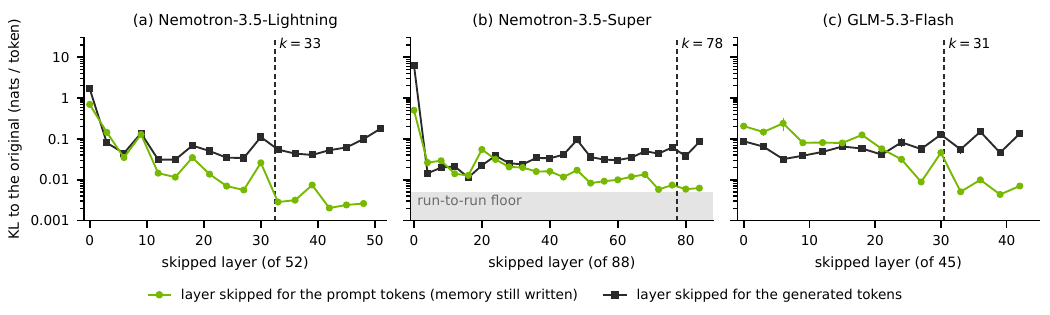}
\caption{The write--read asymmetry of Figure~\ref{fig:prefill-cut}a, b holds on \lightning, \super and \glm. KL from the original model when one layer's output is skipped only for the prompt tokens, its memory still written (green), or only for the generated tokens (black); sampled layers and windows as in Appendix~\ref{app:depth}. \super's values include its engine's run-to-run floor (grey). Dashed: the anchor depth.}
\label{fig:skip-more}
\end{figure}
\fi
\paragraph{Four smaller checkpoints, one of them a transformer.}
The asymmetry is not specific to the hybrids we serve (Figure~\ref{fig:prefill-ladders}). On Qwen3-8B-Base~\cite{qwen3_2025}, Qwen3.5-4B-Base and Qwen3.5-9B-Base~\cite{qwen35_2026} and GLM-4.6V-Flash~\cite{glm46v2025} on text, an untrained shallow prefill at 50--75\% of the layers keeps the direct-answer core within 2\% of the original model and loses mainly needle recall. The core averages GSM8K~\cite{gsm8k2021}, HellaSwag~\cite{hellaswag2019}, ARC-Challenge~\cite{arc2018} and WinoGrande~\cite{winogrande2020} (without GSM8K for the instruction-tuned GLM-4.6V-Flash); needle recall is the minimum over prompts of 2K, 4K and 8K tokens, 64 per length (256 for Qwen3.5-9B-Base at 16 layers). Training the write paths of Qwen3-8B-Base at 22 of 36 layers on 1B tokens, without a code, recovers its needle loss.
\ifreport\begin{appendixwide}{figure}
\centering
\includegraphics[width=\textwidth]{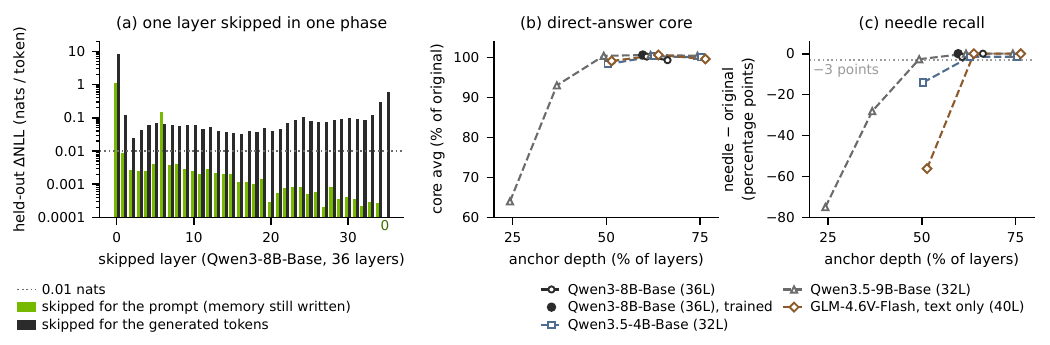}
\par\nopagebreak[4]
\caption{Without training, a shallow prefill keeps a model's direct answers and loses mainly retrieval. (a) Qwen3-8B-Base: held-out NLL change when one layer is skipped only for the prompt tokens (memory still written) or only for the generated tokens; the last layer's prompt-side change is exactly 0 (marked below the axis). (b, c) Untrained shallow prefill at each anchor depth (hollow, dashed); filled: Qwen3-8B-Base at 22 of 36 layers with trained write paths. (b) Core average, \% of the original model. (c) Needle recall minus the original model's; dotted: $-3$ points. A checkpoint's markers share a horizontal shift of at most 1.4 points of depth.}
\label{fig:prefill-ladders}
\end{appendixwide}
\fi

\paragraph{Inexact writes.}
\label{app:depth-writes}
Table~\ref{tab:released} gives the KLs and cache errors of \S\ref{sec:depth}; Table~\ref{tab:staterr} gives the emitted states' errors.
\begin{table}[t]
\centering\footnotesize
\caption{Every released set stays close to the original model although its memories are inexact. Teacher-forced KL to the original model (nats per token) above the floor, the KL of a second run of the original model (nonzero where the engine is not deterministic); windows of Appendix~\ref{app:depth} (\glm: 16 windows, 63 tokens). As served: the emitters read the decoded code and every state is compressed as in the decode. States exact: no compression. Code bypassed: the emitters read the exact $h^{(k)}$ (\glm has no code). Cache error: relative squared error of the emitted attention caches at the first emitted layer, whose original write path itself reads $h^{(k)}$, and beyond it (\super: no attention layer beyond; its code keeps 512 of 4{,}096 coordinates).}
\label{tab:released}
\begin{tabular}{@{}lrrrrrr@{}}
\toprule
 & & \multicolumn{3}{c}{KL above the floor} & \multicolumn{2}{c}{cache error (\%)} \\
\cmidrule(lr){3-5}\cmidrule(l){6-7}
Set & floor & as served & states exact & code bypassed & first emitted layer & beyond \\
\midrule
\model{} bf16 & 0.0033 & 0.0359 & 0.0216 & 0.0411 & 1.4--2.2 & 9.0--66.9 \\
\modelnv{} & 0.0206 & 0.0393 & 0.0224 & 0.0445 & 1.7--2.6 & 10.0--73.5 \\
\lightning{} & 0.0018 & 0.0145 & 0.0100 & 0.0269 & 0.5--4.0 & 11.6--28.6 \\
\kimi{} & 0.0016 & 0.0063 & 0.0038 & 0.0119 & 1.2 & 11.5--67.7 \\
\glm{} & 0.0000 & 0.0330 & 0.0318 & -- & 0.0--0.8 & 2.2--79.5 \\
\super{} & 0.0047 & 0.0099 & 0.0062 & 0.0132 & 3.8--38.0 & -- \\
\bottomrule
\end{tabular}

\end{table}

\appendixDeferWideBlocks
\subsection{The Selection Rule}
\label{app:rule}
Every model's anchor depth, code and state rank follow from one rule, with thresholds shared by all models. This subsection states its criteria (\S\ref{sec:depth-k}) exactly, with their reasons; Table~\ref{tab:depth-rule} lists the measurements and the resulting choices.

\paragraph{The rule.}
The rule reads only the original model, its untrained shallow prefill (Appendix~\ref{app:depth}) and closed-form codes (the principal-component codes that initialize training), never an exam or a trained set; the NVFP4 \model set reuses the bf16 measurements. The \emph{copy probe} hides one record, ``Record $\langle$key$\rangle$ is filed under the identifier $\langle$UUID$\rangle$'', at a random depth of 4K to 16K tokens of held-out web text, and counts the items whose identifier the greedy reply copies exactly (300 items on \model, 1{,}500 on \lightning, \kimi and \super, 600 on \glm).
\begin{enumerate}\setlength\itemsep{0pt}
\item \emph{Depth.} The candidates are the attention-layer boundaries, whose first emitted layer is a softmax-attention layer. $k$ is the shallowest measured boundary at which (D1) the untrained gap is at most 0.075 nats per token and (D2) the untrained shallow prefill copies at least three quarters of the identifiers that the original model copies. D1 gives the prefill three quarters of a budget of a tenth of a nat per token (a perplexity rise of about 10\%) for the untrained shallow prefill and shallow state together; S lets the state add at most a quarter of the prefill's gap, so the sum stays within it ($0.075\cdot5/4\approx0.094$). Boundaries below the first measured one (23 on \model, 19 on \lightning and \glm, 11 on \kimi) are taken to fail D1, as the gap falls with depth at every measured boundary up to $k$; on \super every boundary is measured.
\item \emph{Exact values} (M). $m=256$ if the closed-form code's relative reconstruction error $\|\Delta-\hat\Delta\|/\|\Delta-\mu\|$ at $m=128$ is larger on chat-template tokens than on other tokens of public chat prompts, else $m=128$; $m=0$ without a code. The error is measured at the rank that C1 and C2 select (on \glm at 4608, its largest probed rank within the cap), so $r$ and $m$ are checked jointly.
\item \emph{Code.} Let $d$ be the width of the residual stream. (C3) If the full offset's code takes no more bytes than even the fp8 deep cache, $r=d$. Otherwise $r$ is the smallest probed rank whose closed-form code (C1) shows no significant copy loss against the uncoded shallow prefill and (C2) takes no more bytes than the bf16 deep cache, which a stored code replaces (Table~\ref{tab:prefix-forms}), so a larger code would save nothing. A loss means more identifiers lost than gained with $p<0.10$ in the exact two-sided McNemar test~\cite{mcnemar1947note}, a one-sided level of 0.05 against a loss. (C4) If no rank qualifies, there is no code.
\item \emph{State} (S). $R$ is the smallest of 8, 16 and 32 whose state drift is at most $\mathrm{gap}(k)/4$, else 32; $W=R$. The state drift (\S\ref{sec:depth-k}) is the per-token $\mathrm{KL}(p\,\|\,\tilde p)$ from exact states ($p$) to the shallow form ($\tilde p$), including the original model's run-to-run floor.
\end{enumerate}

\paragraph{The criteria on the five models.}
\emph{Depth} (Figure~\ref{fig:gap}). On \glm, D2 rejects 27, whose gap passes D1 (\ruleGLPrevGap). On \super, D1 rejects 69 (0.091) and accepts 78 (0.065).
\emph{Code.} C1 sets the ranks of \model (8192, the smallest probed passing rank), \lightning (2048) and \super (512). C3 sets \kimi's $r=d$ by bytes alone. On \glm every probed rank loses significantly (2048--4608 and 8192); the unprobed ranks up to the cap of C2 (6144) lie between failing ones, and on every probed grid copying grows with the rank, so C4 removes the code.
\emph{Exact values.} The template clause fires on \model (\ruleFNTemplateErr{} against \ruleFNContentErr) and on \glm (\ruleGLTemplateErr{} against \ruleGLContentErr).
\emph{State} (Figure~\ref{fig:rule-state}). \kimi's rank 16 drifts \ruleKMDriftSixteen{} against its budget \ruleKMBudget, so it gets rank 32. \glm's budget (\ruleGLBudget) lies below its floor (\ruleGLFloor), so the fallback gives 32. An ablation set of \kimi (released code, states trained at rank 8) loses $5.2\pm2.0$ AIME points against the original model when it decodes at rank 16 every 16 steps, and $1.5\pm2.1$ at the criterion's rank 32 every 32 steps (paired by question and seed, eight repetitions).

\begin{appendixwide}{table}
\centering\scriptsize
\setlength{\tabcolsep}{3.5pt}
\caption{Inputs and choices of the selection rule (Appendix~\ref{app:rule}). The \model column serves its bf16 and NVFP4 sets (probes on bf16). Gap: the untrained gap at the attention-layer boundaries, nats per token. Copies: identifiers copied by the untrained shallow prefill / the original model. Closed form: items lost / gained by a closed-form code of rank $r$ against the uncoded shallow prefill ($p$: exact McNemar). Bytes: nominal per token; in parentheses, over the bf16 / fp8 deep cache. Template / other: relative reconstruction error at $m=128$ and the selected rank. Full offset: the code of rank $d$, over the fp8 deep cache. Drift: decode drift at rank $R$, $W=R$, against $\mathrm{gap}(k)/4$.}
\label{tab:depth-rule}
\nopagebreak[4]
\begin{tabular}{@{}l*{5}{>{\raggedright\arraybackslash\everypar{\hangindent=1em\hangafter=1\relax}}p{26.5mm}}@{}}
\toprule
 & \begin{tabular}[b]{@{}l@{}}Qwen3.8-\\Flash-Next\end{tabular} & \begin{tabular}[b]{@{}l@{}}Nemotron-3.5-\\Lightning\end{tabular} & \begin{tabular}[b]{@{}l@{}}Kimi-\\Linear\end{tabular} & \begin{tabular}[b]{@{}l@{}}GLM-5.3-\\Flash\end{tabular} & \begin{tabular}[b]{@{}l@{}}Nemotron-3.5-\\Super\end{tabular} \\
\midrule
\multicolumn{6}{@{}l}{\emph{depth $k$}} \\
gap at boundary & 23:~0.102\par 27:~0.100\par 31:~0.074 & 19:~0.182\par 26:~0.127\par 33:~0.075 & 11:~0.174\par 15:~0.078\par 19:~0.031 & 19:~0.172\par 23:~0.115\par 27:~0.045\par 31:~0.035 & 58:~0.216\par 69:~0.091\par 78:~0.065 \\
copies & 31:~290~/~297 & 33:~1,498~/~1,500 & 19:~1,500~/~1,497 & 27:~269~/~595\par 31:~530~/~595 & 78:~1,494~/~1,496 \\
\addlinespace
\multicolumn{6}{@{}l}{\emph{code $r{+}m$}} \\
closed form & 4096:~227~/~2 ($p<0.001$)\par 6144:~102~/~7 ($p<0.001$)\par 8192:~15~/~7 ($p=0.134$)\par 10240:~4~/~4 ($p=1.000$) & 1024:~17~/~1 ($p<0.001$)\par 1536:~12~/~1 ($p=0.003$)\par 2048:~5~/~1 ($p=0.219$)\par 2688:~2~/~0 ($p=0.500$) & 1024:~0~/~0\par 1536:~0~/~0\par 2048:~0~/~0\par 2304:~0~/~0 & 2048--4608:\ at~least~528~/~0 ($p<0.001$)\par 8192:~144~/~11 ($p<0.001$) & 256:~5~/~0 ($p=0.0625$)\par 512:~3~/~0 ($p=0.250$) \\
bytes & $8192{+}256$:~5,380 (0.509~/~1.019) & $2048{+}128$:~1,540 (0.752~/~1.504) & $2304{+}128$:~1,684 (0.487~/~0.975) & none; $8192{+}256$ would take 5,380 (1.272~/~2.468) & $512{+}128$:~676 (0.660~/~1.320) \\
full offset & $10240{+}256$:~1.237 & $2688{+}128$:~1.855 & $2304{+}128$:~0.975 & $16384{+}256$:~4.582 & $4096{+}128$:~5.258 \\
template / other & 0.180~/~0.153 (1.18) & 0.212~/~0.245 (0.86) & 0.075~/~0.078 (0.97) & 0.488~/~0.384 (1.27) at $r=4608$ & 0.514~/~0.660 (0.78) \\
\addlinespace
\multicolumn{6}{@{}l}{\emph{state $R$, $W$}} \\
drift, $R=8$ & 0.0209 & 0.0281 & 0.0325 & 0.0865 & 0.0408 \\
drift, $R=16$ & 0.0073 & 0.0087 & 0.0096 & 0.0360 & 0.0140 \\
drift, $R=32$ & 0.0031 & 0.0029 & 0.0030 & 0.0214 & 0.0054 \\
$\mathrm{gap}(k)/4$ & 0.0184 & 0.0187 & 0.0077 & 0.0087 & 0.0162 \\
\midrule
\multicolumn{6}{@{}l}{\emph{settings}} \\
$k$ & 31 & 33 & 19 & 31 & 78 \\
$r{+}m$ & $8192{+}256$ & $2048{+}128$ & $2304{+}128$ & no code & $512{+}128$ \\
$R$ / $W$ & 16 / 16 & 16 / 16 & 32 / 32 & 32 / 32 & 16 / 16 \\
\bottomrule
\end{tabular}
\end{appendixwide}

\begin{appendixwide}{figure}
\centering
\includegraphics[width=\textwidth,height=0.5\textheight,keepaspectratio]{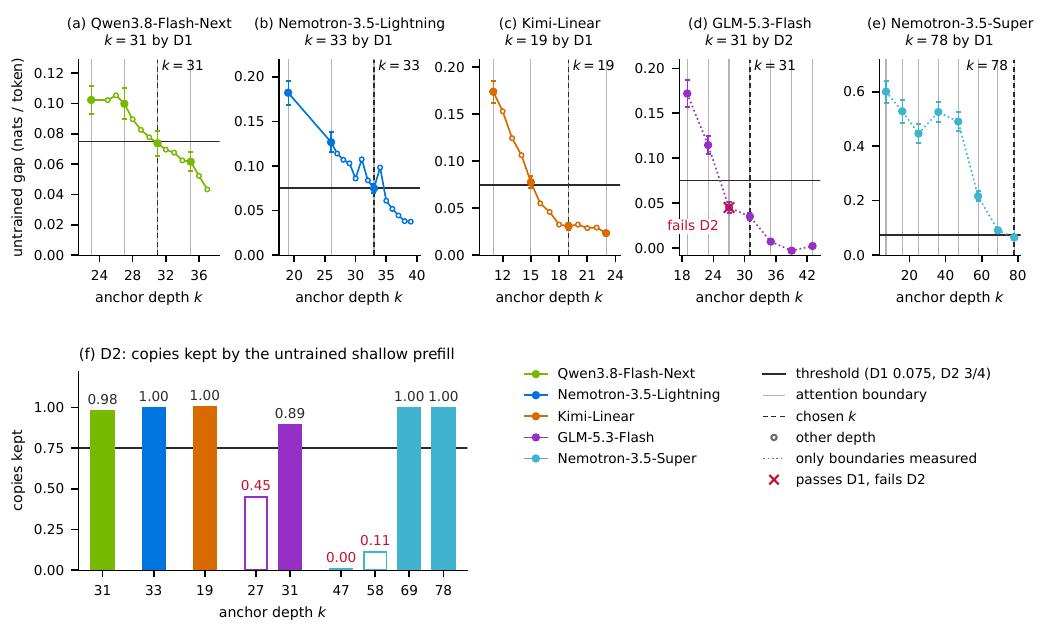}
\par\nopagebreak[4]
\caption{The anchor depth is the shallowest boundary that passes both depth clauses. (a)--(e) The untrained gap at the attention-layer boundaries ($\pm$ standard error, grey lines) and other depths (open markers; \glm and \super: boundaries only); black: the D1 threshold; dashed: the chosen $k$. (f) D2: identifiers copied by the untrained shallow prefill over those copied by the original model; threshold $3/4$. On \glm, 27 passes D1 and fails D2; on \super, every boundary below 78 fails D1.}
\label{fig:gap}
\end{appendixwide}

\paragraph{Why copying is a separate criterion.}
The gap, a code's cost and a closed-form subspace's variance average over tokens or directions, but a 36-character identifier is copied only if each of its many tokens is. A per-token error rate $\eta$ that an average barely registers compounds to $1-(1-\eta)^n\approx n\eta$ over $n$ independent tokens. On \model, the code $4096{+}512$ costs \ruleFNFourKCost{} nats per token, under a quarter of the gap, and holds \ruleFNFourKVar{} of the offset's variance, yet loses \ruleFNFourKLost{} of the \ruleFNCutCopies{} identifiers that the uncoded shallow prefill copies. Copying needs the directions that carry a token's identity, which need not hold much variance. Training keeps the difference: on RULER's UUID needle tasks, trained sets with the code $4096{+}512$ miscopy far more identifiers than the released sets with $8192{+}256$ (Table~\ref{tab:uuid}), and on \glm a preliminary trained set with the code $5632{+}256$ at depth 27 copies \ruleGLPOne{} identifiers, against \ruleGLPTwo{} for a like set without a code ($p=\ruleGLPOneP$): training does not repair a code that miscopies.
\begin{appendixwide}{table}
\centering\footnotesize
\setlength{\tabcolsep}{5pt}
\caption{RULER's two UUID tasks on \model: wrong replies out of 100 documents per task and length, niah\_single\_3 / niah\_multikey\_3 (the original model: none wrong). The rows with the code $4096{+}512$ are an earlier, unreleased bf16 set after each stage.}
\label{tab:uuid}
\begin{tabular}{@{}llr@{\,/\,}lr@{\,/\,}lr@{\,/\,}l@{}}
\toprule
Set & Code & \multicolumn{2}{c}{32K} & \multicolumn{2}{c}{64K} & \multicolumn{2}{c@{}}{128K} \\
\midrule
NVFP4, released & $8192{+}256$ & 1 & 0 & 1 & 1 & 4 & 5 \\
bf16, unreleased, stage 1 & $4096{+}512$ & 21 & 4 & 18 & 4 & 19 & 11 \\
bf16, unreleased, stage 2 & $4096{+}512$ & 22 & 3 & 12 & 5 & 16 & 15 \\
bf16, released & $8192{+}256$ & 1 & 0 & 1 & 2 & 1 & 5 \\
\bottomrule
\end{tabular}
\end{appendixwide}

\paragraph{Why a budget of a quarter of the gap.}
D1 assumes that the KL costs of the shallow prefill and of the compressed states add. To second order, they add when the errors are uncorrelated in the Fisher metric of the output distribution. On three preliminary \glm sets (\ruleAddN{} comparisons), a set's KL with its state form is \ruleAddLo--\ruleAddHi{} of the sum of its KL with exact states and the state form's KL on the original model. The costs combine sub-additively, so assuming additivity errs on the safe side.

\begin{appendixwide}{figure}
\centering
\includegraphics[width=\textwidth,height=0.5\textheight,keepaspectratio]{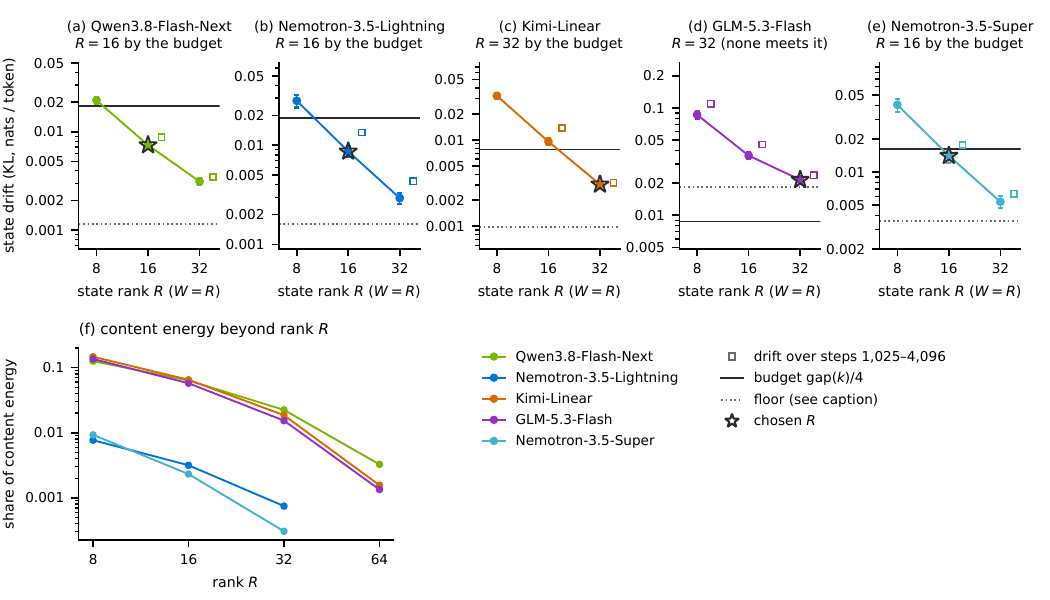}
\par\nopagebreak[4]
\caption{The state clause takes the smallest rank whose state drift fits the budget, else 32. (a)--(e) State drift ($\pm$ standard error) at $R=8,16,32$ with $W=R$; open squares: its mean over decode steps 1{,}025--4{,}096 of 24 replies, drawn at $1.2R$. Black: the budget $\mathrm{gap}(k)/4$. Dotted: the floor, a second run of the original model (on \glm, the exact state scanned in 16-token pieces). Star: the chosen $R$. (f) Share of the content energy ($C=S-a\vbar^{\top}$) beyond rank $R$ in prompt-final states of held-out 4K windows, energy-weighted over heads and layers.}
\label{fig:rule-state}
\end{appendixwide}

\paragraph{The byte account.}
The deep cache takes 10{,}560 / 5{,}280 bytes per token in bf16 / fp8 on \model, 2{,}048 / 1{,}024 on \lightning, 3{,}456 / 1{,}728 on \kimi, 4{,}228 / 2{,}180 on \glm and 1{,}024 / 512 on \super. A code of rank $r$ (a multiple of 16) with $m$ exact values takes nominally $\tfrac{9}{16}r+3m+4$ bytes per token: half a byte per NVFP4 coordinate plus an 8-bit scale per sixteen, a bf16 value and a one-byte index per exact value, and one fp32 scale per token. Each index counts the residual-stream positions skipped since the previous exact value; each gap of 255 or more adds two bytes (on \model at most 78 bytes, 1.5\% of the code), which changes no outcome of C2 or C3.

\paragraph{The exact values and the drift window.}
The template clause guards against an observed failure: a \model set on the NVFP4 checkpoint with a full-rank code and $m=128$ answered 66 of 200 prompts of a RULER multi-key needle task at 8K with an immediate end-of-sequence token, because the code of the answer's template token collapsed; with $m=256$, 4 of 800 such prompts did. In the open-loop sum~\eqref{eq:trunc}, each compression's error fades with the head's decays, so recent compressions dominate; the drift is therefore read late, over steps 961--1{,}024. $W=R$ is a convention, and the ranks stop at 32 because at $R=W=32$ the 64 factor columns of a $128\times128$ head hold as many numbers as the head.

\FloatBarrier

\section{Theory}
\label{app:proofs}\label{app:theory}\label{app:principle}

This appendix proves the four facts on which \duet's shallow state rests. Every recurrent state splits exactly into a sink and a content (Lemma~\ref{prop:decomp}). The sink-plus-factors form is closed under the update, so compression is the only approximation (Lemma~\ref{prop:closed}). In the open loop, the compression errors do not accumulate (Proposition~\ref{prop:trunc} and Corollary~\ref{cor:periodic}). After the head's normalization, the sink acts as a bias and a gain (Lemma~\ref{lem:gain}).

\subsection{One recurrence for three layer types}

Gated DeltaNet, Kimi Delta Attention and Mamba-2 heads update their states by the same linear recurrence, so we state every result once, for
\begin{equation}
S_t=\Phi_tS_{t-1}+\beta_t\kappa_tv_t^{\top},\qquad \Phi_t=(I-\theta\beta_t\kappa_t\kappa_t^{\top})\Lambda_t .
\label{eq:rec}
\end{equation}
Here $S_t\in\Rr^{d_k\times d_v}$, the decay $\Lambda_t$ is diagonal with entries in $(0,1)$, and $\theta\in\{0,1\}$ switches the delta rule's correction on or off. The three layer types set these as follows.
\begin{itemize}
\item Gated DeltaNet~\eqref{eq:gdn}: $\theta=1$ and $\Lambda_t=g_tI$.
\item Kimi Delta Attention: $\theta=1$ and $\Lambda_t=\mathrm{diag}(g_{t,1},\dots,g_{t,d_k})$, one decay per key channel.
\item Mamba-2: $\theta=0$ and $\Lambda_t=\gamma_tI$. Its state is $d_x\times d_B$; in its transpose, the input projection $B_t$, the input $x_t$ and the step size $\delta_t$ take the places of $\kappa_t$, $v_t$ and $\beta_t$, and $d_B$, $d_x$ those of $d_k$, $d_v$.
\end{itemize}

Every step contracts the state. Let $\lambda_t$ be the largest entry of $\Lambda_t$. In the delta rules, $\|\kappa_t\|=1$ and $\beta_t\in(0,1)$, so $I-\beta_t\kappa_t\kappa_t^{\top}$ has spectral norm one. Hence, in all three layer types,
\begin{equation*}
\|\Phi_t\|_2\le\lambda_t .
\end{equation*}

The sink direction $\vbar\neq0$ is the head's mean value (for Mamba-2, its mean input $\bar x$), and $\omega_t$, $\varepsilon_t$, $a_t$ and $C_t$ are as in Lemma~\ref{prop:decomp}. The shallow state stores a sink coefficient $a$ and content factors $U$, $V$,
\begin{equation*}
S=a\vbar^{\top}+UV^{\top},\qquad V^{\top}\vbar=0,
\end{equation*}
and compresses the content $UV^{\top}$ to rank $R$ every $W$ steps.

\subsection{The sink decomposition}

Lemma~\ref{prop:decomp} (Section~\ref{sec:sink}) splits every state exactly into a rank-one sink along $\vbar$ and a content orthogonal to it. Since both parts follow the head's own recurrence, \duet can store and update them separately.

\begin{proof}[Proof of Lemma~\ref{prop:decomp}]
The recurrence is linear in the values, so it suffices to project it onto $\vbar$. Since $v_t^{\top}\vbar=\omega_t\|\vbar\|^2$, right-multiplying the recurrence by $\vbar/\|\vbar\|^2$ gives the recurrence for $a_t$; subtracting its product with $\vbar^{\top}$ from the recurrence for $S_t$ gives the one for $C_t$. For the energies, the two terms of
\begin{equation*}
S_t-z\vbar^{\top}=(a_t-z)\,\vbar^{\top}+C_t
\end{equation*}
are orthogonal, because $\langle y\vbar^{\top},C_t\rangle_F=y^{\top}C_t\vbar=0$ for every $y$. Their squared Frobenius norms therefore add, and $\|(a_t-z)\vbar^{\top}\|_F=\|a_t-z\|\,\|\vbar\|$.
\end{proof}

\subsection{The factored update is exact}

The factored form never needs the dense state. Each step updates the sink coefficient and appends one column to each content factor. The next lemma shows that this update is exact for all three layer types and keeps the content orthogonal to $\vbar$, and that compressing the content leaves the sink untouched. Compression is therefore the shallow state's only approximation.

\begin{lemma}[exact factored update]
\label{prop:closed}
One step of~\eqref{eq:rec} maps a sink plus factors to a sink plus factors with one more column each. Precisely, let $S=a\vbar^{\top}+UV^{\top}$ with $V^{\top}\vbar=0$, and let $v=\omega\vbar+\varepsilon$ with $\varepsilon^{\top}\vbar=0$. Then $\Phi S+\beta\kappa v^{\top}=a^{+}\vbar^{\top}+U^{+}V^{+\top}$, where
\begin{equation}
\begin{split}
a^{+}&=\Lambda a+\beta\,(\omega-\theta\,\kappa^{\top}\Lambda a)\,\kappa,\\
U^{+}&=\big[\,\Lambda U,\;\kappa\,\big],\qquad V^{+}=\big[\,V,\;\beta\,(\varepsilon-\theta\,VU^{\top}\Lambda\kappa)\,\big],
\end{split}
\label{eq:closed}
\end{equation}
and $V^{+\top}\vbar=0$. For $\theta=1$ and $\Lambda=gI$, this is the update~\eqref{eq:append}. Moreover, if $Q$ and $P$ have orthonormal columns in the column space and the row space of the content $C=UV^{\top}$, then both compressions $C'=QQ^{\top}C$ and $C'=CPP^{\top}$ satisfy $C'\vbar=0$, so replacing $C$ by $C'$ leaves the sink coefficient $a=S\vbar/\|\vbar\|^2$ unchanged.
\end{lemma}

\begin{proof}
The step acts on the sink and the content separately, except for the delta rule's correction, which reads the content along $\kappa$ and writes the result into the new column. Since $\Lambda$ is symmetric, $\Phi=\Lambda-\theta\beta\kappa\kappa^{\top}\Lambda$, and substituting $S$ and $v$ gives
\begin{equation*}
\begin{aligned}
\Phi S+\beta\kappa v^{\top}&=(\Phi a+\beta\omega\kappa)\,\vbar^{\top}+\Phi UV^{\top}+\beta\kappa\varepsilon^{\top}\\
&=a^{+}\vbar^{\top}+\Lambda UV^{\top}+\beta\kappa\,(\varepsilon-\theta\,VU^{\top}\Lambda\kappa)^{\top}\\
&=a^{+}\vbar^{\top}+U^{+}V^{+\top}.
\end{aligned}
\end{equation*}
The new column of $V^{+}$ combines $\varepsilon$ and the columns of $V$, all orthogonal to $\vbar$, so $V^{+\top}\vbar=0$. For the compressions, $QQ^{\top}C\vbar=0$ because $C\vbar=0$. The same identity makes $\vbar$ orthogonal to the row space of $C$, so $P^{\top}\vbar=0$ and $CPP^{\top}\vbar=0$. Both compressions stay in factored form,
\begin{equation*}
QQ^{\top}C=Q\,(VU^{\top}Q)^{\top},\qquad CPP^{\top}=(UV^{\top}P)\,P^{\top}. \qedhere
\end{equation*}
\end{proof}

\subsection{Compression errors do not accumulate}

In the open loop, the compressed and the exact recurrence see the same keys, values, decays and write strengths, and differ only by the errors that compressions inject. The proposition below bounds the resulting error and shows that the sink coefficient stays exact; its corollary shows that periodic compressions do not compound.

\begin{proposition}[open-loop error]
\label{prop:trunc}
The error at step $t$ is at most the sum of the injected errors, each damped by the decays applied after it. Precisely, let
\begin{equation*}
\tilde S_0=S_0+\mathcal{E}_0,\qquad \tilde S_t=\Phi_t\tilde S_{t-1}+\beta_t\kappa_tv_t^{\top}+\mathcal{E}_t\quad(t\ge1),
\end{equation*}
with the inputs of~\eqref{eq:rec}. Then
\begin{equation}
\|\tilde S_t-S_t\|_F\le\sum_{s=0}^{t}\Big(\prod_{u=s+1}^{t}\lambda_u\Big)\|\mathcal{E}_s\|_F .
\label{eq:trunc}
\end{equation}
If, moreover, every $\mathcal{E}_s\vbar=0$, then $\tilde S_t\vbar=S_t\vbar$ for all $t$.
\end{proposition}

\begin{proof}
The difference $\mathcal{D}_t=\tilde S_t-S_t$ follows the same recurrence without inputs, driven only by the injected errors: $\mathcal{D}_t=\Phi_t\mathcal{D}_{t-1}+\mathcal{E}_t$ with $\mathcal{D}_0=\mathcal{E}_0$. Unrolling it gives
\begin{equation*}
\mathcal{D}_t=\sum_{s=0}^{t}\Phi_t\cdots\Phi_{s+1}\,\mathcal{E}_s .
\end{equation*}
Since $\|\Phi_u\mathcal{E}\|_F\le\|\Phi_u\|_2\,\|\mathcal{E}\|_F\le\lambda_u\|\mathcal{E}\|_F$, the triangle inequality gives~\eqref{eq:trunc}. If every $\mathcal{E}_s\vbar=0$, each term of the sum vanishes on $\vbar$, so $\mathcal{D}_t\vbar=0$.
\end{proof}

\begin{corollary}[periodic compression]
\label{cor:periodic}
If a compression every $W$ steps injects an error of norm at most $\bar\epsilon$, the error never exceeds $\bar\epsilon$ times the head's decay horizon. Precisely, let $\|\mathcal{E}_s\|_F\le\bar\epsilon$, with $\mathcal{E}_s=0$ unless $W$ divides $s$, and let the mean of $\lambda_u$ over each period $u=mW{+}1,\dots,(m{+}1)W$ be at most $\bar g<1$ (as it is when every $\lambda_u\le\bar g$). Then, for every $t$,
\begin{equation*}
\|\tilde S_t-S_t\|_F\le\frac{\bar\epsilon}{1-\bar g^{\,W}}\le\frac{\bar\epsilon}{1-\bar g} .
\end{equation*}
\end{corollary}

\begin{proof}
Count the compressions before $t$ from the most recent, $n=0,1,\dots$. The decay product that damps the $n$-th one in~\eqref{eq:trunc} spans $n$ full periods and part of one more. By the inequality of arithmetic and geometric means, each full period contributes at most $\bar g^{\,W}$, and the partial period at most one. Hence
\begin{equation*}
\|\tilde S_t-S_t\|_F\le\bar\epsilon\sum_{n\ge0}\bar g^{\,nW}=\frac{\bar\epsilon}{1-\bar g^{\,W}},
\end{equation*}
and $\bar g^{\,W}\le\bar g$ gives the second bound.
\end{proof}

\subsection{After normalization, the sink is a bias and a gain}

Gated DeltaNet and Kimi Delta Attention heads normalize their read-out before the output gate. The lemma below shows that the sink then sets a fixed output direction and, through the normalizer, scales the content, as the sink logit does in a softmax denominator.

\begin{lemma}[the sink as a gain]
\label{lem:gain}
In a Gated DeltaNet or Kimi Delta Attention head, the read-out splits into a part along $\vbar$ and a part orthogonal to it,
\begin{equation*}
o_t=S_t^{\top}q_t=b_t+r_t,\qquad b_t=(q_t^{\top}a_t)\,\vbar,\qquad r_t=C_t^{\top}q_t,\qquad r_t^{\top}\vbar=0 .
\end{equation*}
An RMSNorm with unit weight and constant $\eta\ge0$ therefore returns
\begin{equation}
\rms(o_t)=\frac{\sqrt{d_v}\,(b_t+r_t)}{\big(\|b_t\|^2+\|r_t\|^2+d_v\eta\big)^{1/2}} .
\label{eq:gain}
\end{equation}
If $\eta=0$ and $q_t^{\top}a_t\neq0$, the normalized read-out is a fixed direction plus the content scaled by $1/\|b_t\|$, up to an error quadratic in the ratio $\rho_t=\|r_t\|/\|b_t\|$. With $s=\mathrm{sign}(q_t^{\top}a_t)$ and $\hat v=\vbar/\|\vbar\|$,
\begin{equation*}
\Big\|\rms(o_t)-\sqrt{d_v}\,\Big(s\hat v+\frac{r_t}{\|b_t\|}\Big)\Big\|=\sqrt{d_v}\,\Big(\big(1+\rho_t^2\big)^{1/2}-1\Big)\le\sqrt{d_v}\,\frac{\rho_t^2}{2} .
\end{equation*}
\end{lemma}

\begin{proof}
Since $C_t\vbar=0$, we have $r_t^{\top}\vbar=q_t^{\top}C_t\vbar=0$, so $\|o_t\|^2=\|b_t\|^2+\|r_t\|^2$, which gives~\eqref{eq:gain}. With $\eta=0$, dividing the numerator and the denominator of~\eqref{eq:gain} by $\|b_t\|$, and using $b_t/\|b_t\|=s\hat v$, gives
\begin{equation*}
\rms(o_t)=\sqrt{d_v}\,\frac{x}{\|x\|},\qquad x=s\hat v+\frac{r_t}{\|b_t\|},\qquad \|x\|=\big(1+\rho_t^2\big)^{1/2}.
\end{equation*}
The difference $\sqrt{d_v}\,x\,(1/\|x\|-1)$ thus has norm $\sqrt{d_v}\,(\|x\|-1)$, and $\sqrt{1+\rho^2}\le1+\rho^2/2$.
\end{proof}

\FloatBarrier

\section{Experimental Details and State Probes}
\label{app:details}\label{app:extra}
This appendix gives the model dimensions and the protocols of the state measurements, storage precision and closed-loop drift.
\paragraph{Model dimensions.}
\lightning has 30B parameters (3B active per token), \kimi 48B (3B), \super 120B (12B), \model 125B (6B, plus a 51B-parameter n-gram embedding table read by lookup) and \glm 320B (18B). \model, \glm and \super run with text input only.
\model: hidden size 2560; 512 routed experts (ten active) and one shared; four parallel 2560-dimensional residual streams, mixed and updated at every layer by input-dependent gates. Its sparse-attention layers have 24 query heads and two 256-dimensional key--value (KV) heads, select 2048 tokens per query and keep one indexer key per four tokens, 2112 bytes per token in all.
\lightning: hidden size 2688; attention layers with two 128-dimensional KV heads.
\kimi: hidden size 2304; 256 routed experts (eight active) and one shared. Its MLA layers cache the model's own latent key--value format: a 512-number latent and a 64-number shared key (the rotary part, which \kimi leaves unrotated).
\glm: hidden size 4096 in four residual streams, mixed with input-dependent weights by manifold-constrained hyper-connections (mHC); 288 routed experts (eight active) and one shared, with dense feed-forward layers in the first three layers. Expert, dense feed-forward and most sparse-attention projection weights are fp8 (E4M3, $128\times128$ blocks) with dynamically quantized fp8 activations; the Kimi Delta Attention layers, the indexer and the latent up-projection stay bf16. Each DSA layer has a 32-head indexer whose 128-dimensional keys are pooled over four tokens and cached in fp8 with a 4-byte scale, so the layer caches 1{,}057 bytes per token: the 1{,}024-byte bf16 latent plus 33 bytes of indexer key. The fp8 cache figures of Tables~\ref{tab:depth-rule} and~\ref{tab:prefix-forms} halve the latent only.
\super: hidden size 4096; 512 routed experts (22 active) and one shared, computing in a 1024-dimensional latent space; attention layers with 32 query heads and two 128-dimensional KV heads; Mamba-2 layers with eight groups. We use the language model of its vision-language checkpoint unchanged, without the vision tower and projector.
Above the anchor layer, the emitters write the prompt entries of the attention caches of layers 31, 35, 39, 43 and 47 on \model; 33 and 42 on \lightning; MLA layers 19, 23 and 26 on \kimi; DSA layers 31, 35, 39 and 43 on \glm; and layer 78 on \super.

\paragraph{State probes.}
States are captured in fp32 at the end of the same six held-out web prompts of 3{,}584 tokens (Figure~\ref{fig:sink-heat}). Independent random directions in $\Rr^{128}$, the baseline of Figure~\ref{fig:sink-heat}c, have a median $|\cos|$ of 0.06 (95th percentile 0.18). Figure~\ref{fig:sink-matrix} replays the gated delta rule in fp64 over one prompt without a document boundary. \lightning uses the held-out stream of its own tokenizer and all $23\times64$ Mamba-2 heads; the content of Figure~\ref{fig:sink-heat}d uses $\vbar$ estimated on other text, and the other models use the mean input or mean value that their component sets ship. The median and pooled sink shares are 75\% and \sinkPooledMinPct\% on \model, 64\% and 99.6\% on \lightning, \sinkSUMedPct\% and \sinkPooledMaxPct\% on \super, \sinkKMMedPct\% and \sinkKMPoolPct\% on \kimi, and \sinkGLMedPct\% and \sinkGLPoolPct\% on \glm. On six windows of each model's own math and code replies, every share changes by at most \sinkDomainMaxChange.

\paragraph{Eviction and step-wise policies.}
The eviction experiment acts once, in all 36 layers, on the state after the first 3{,}583 tokens (\duet also compresses before the last prompt token), with $\vbar$ estimated on other text; the last prompt token then runs, and the next 512 tokens are scored with exact state updates. The sink's replacement by its mean over other prompts averages over the other five prompts. The step-wise policies of \S\ref{sec:sink-content} start from the same state and truncate exactly after every token of a 256-token teacher-forced continuation, with $\vbar$ the mean value of the other five prompts. Every policy scores the same 1536 tokens.

\begin{appendixwide}{figure}
\centering
\includegraphics[width=\textwidth,height=0.5\textheight,keepaspectratio]{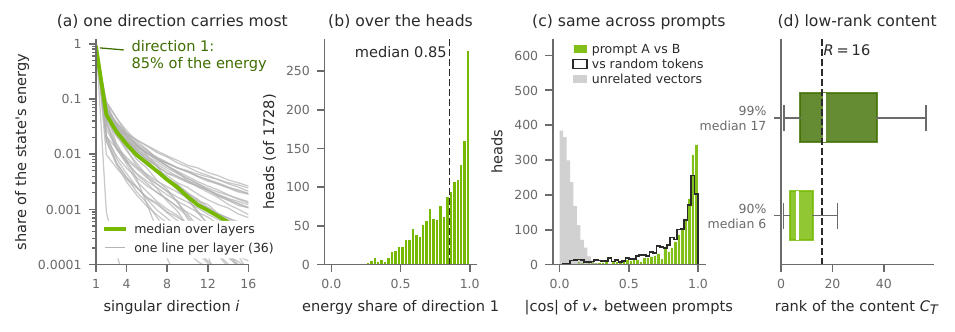}
\par\nopagebreak[4]
\caption{One direction carries most of the recurrent state, and it is nearly the same on every prompt tested (\model, $36\times48$ heads, six web prompts of 3{,}584 tokens). (a) Energy share of the state's $i$-th singular direction; grey: one layer (median over heads and prompts); green: median over layers. (b) The top direction's share over the 1{,}728 heads. (c) $|\cos|$ between a head's top output directions $v_\star$ on two unrelated prompts (green), on a web prompt and a prompt of uniformly random tokens (charcoal outline), and for independent random directions in $\Rr^{128}$ (grey). (d) Rank holding 90\% and 99\% of the energy of the content $C_T$ (Lemma~\ref{prop:decomp}) per head, averaged over the prompts; box 25--75\%, whiskers 5--95\%, bar the median; dashed: \model's $R=\fnR$.}
\label{fig:sink-heat}
\end{appendixwide}

\FloatBarrier

\paragraph{Storage precision.}
\label{app:precision}
Decode drift (Tables~\ref{tab:precision-main} and~\ref{tab:precision}) is the per-token KL divergence from the original model's next-token distribution, teacher-forced over the first 1{,}024 tokens of 48 of its own replies to held-out prompts (16 each of math, science and code). It counts the original model's 1{,}024 most likely tokens plus one bucket for the remaining mass, and is reported above a second run of the original model (0.001--0.004 nats per token), paired over replies. On \glm, whose reference engine takes about 8\,s per decode step, it uses 24 replies over their first 256 tokens and is reported above the full-rank factored path, through which every rounded form runs.

\paragraph{Rounding.}
The original model's dense state is rounded after every token. The shallow state rounds its factors and sink coefficient at every compression (fp8 with one fp32 scale per vector) and keeps the up to $W$ newest columns in fp32 between compressions. Stochastic rounding adds a uniform random integer below the dropped bits to the fp32 bit pattern before truncating, so it is unbiased. Under either rounding, fp16 factors change the shallow state's drift by at most \precFPSixteenMaxChange{} nats per token.

\paragraph{NVFP4 checkpoints.}
The NVFP4 row of Table~\ref{tab:precision-main} decodes the same replies with NVIDIA's official NVFP4 checkpoints of \model, \lightning and \super and their original caches and states, against the bf16 model's reference pass and above its second run. Our reference path runs them with the served arithmetic of their FP8 and NVFP4 layers and SGLang's scale conventions. On four held-out 2K-token windows, its per-token loss matches SGLang's within \nvCheckMaxDiff{} nats on average on \model and \lightning; on \super, the NVFP4 checkpoint raises the loss by \nvSUEffectOurs{} nats in our path and by \nvSUEffectSgl{} in SGLang.

\begin{appendixwide}{table}
\centering\scriptsize\setlength{\tabcolsep}{4pt}
\caption{The shallow state's drift is the same with fp32 and fp16 factors under either rounding; a dense state's depends on the format and, on the Mamba-2 models, on the rounding. Decode drift (nats per token above a second run of the original model) of the original model's dense state, rounded after every token, and of the shallow state, rounded at every compression. The exams use the fp32 shallow state. Last row: size per head and layer in 16 bits, the shallow state's right after a compression.}
\label{tab:precision}
\setlength{\tabcolsep}{4pt}
\begin{tabular}{@{}llrrrrrrrrrr@{}}
\toprule
 & & \multicolumn{2}{c}{\shortstack{Qwen3.8-\\Flash-Next}} & \multicolumn{2}{c}{\shortstack{Nemotron-3.5-\\Lightning}} & \multicolumn{2}{c}{\shortstack{Kimi-\\Linear}} & \multicolumn{2}{c}{\shortstack{GLM-5.3-\\Flash}} & \multicolumn{2}{c}{\shortstack{Nemotron-3.5-\\Super}} \\
\cmidrule(lr){3-4} \cmidrule(lr){5-6} \cmidrule(lr){7-8} \cmidrule(lr){9-10} \cmidrule(lr){11-12}
format & rounding & dense & shallow & dense & shallow & dense & shallow & dense & shallow & dense & shallow \\
\midrule
fp32 & -- & 0 & 0.0056 & 0 & 0.0060 & 0 & 0.0017 & 0 & 0.0027 & 0 & 0.0070 \\
fp16 & nearest & 0.0006 & 0.0056 & 0.0011 & 0.0062 & 0.0002 & 0.0017 & 0.0006 & 0.0021 & 0.0012 & 0.0069 \\
fp16 & stochastic & 0.0006 & 0.0056 & 0.0004 & 0.0059 & 0.0002 & 0.0016 & 0.0010 & 0.0025 & 0.0002 & 0.0070 \\
bf16 & nearest & 0.0008 & 0.0056 & 0.0057 & 0.0061 & 0.0008 & 0.0017 & 0.0008 & 0.0030 & 0.0066 & 0.0072 \\
bf16 & stochastic & 0.0007 & 0.0056 & 0.0044 & 0.0065 & 0.0004 & 0.0017 & 0.0010 & 0.0024 & 0.0048 & 0.0073 \\
fp8 & nearest & 0.0088 & 0.0067 & 0.0487 & 0.0136 & 0.0104 & 0.0023 & 0.0070 & 0.0030 & 0.0745 & 0.0160 \\
fp8 & stochastic & 0.0073 & 0.0076 & 0.1752 & 0.0194 & 0.0048 & 0.0027 & 0.0055 & 0.0037 & 0.1740 & 0.0199 \\
\midrule
KB per head, 16 bits & & 32.8 & 8.4 & 16.4 & 6.4 & 32.8 & 16.6 & 32.8 & 16.6 & 16.4 & 6.4 \\
\bottomrule
\end{tabular}

\end{appendixwide}

\ifreport\paragraph{Closed-loop drift.}\label{app:closed-loop} The long replies of \S\ref{sec:eval-drift} are the original model's replies of at least 16{,}384 tokens (on \lightning and \super from their long-reply sets), scored over their first 16{,}384 decode steps at the released $R$ and $W$, per block of 1{,}024 steps, above a second run of the original model. Only the tokens are teacher-forced: a compression error changes the later keys, values and decays.
\fi

\FloatBarrier

\section{Training}
\label{app:training}
This appendix gives the objectives, data and settings of the two training stages of \S\ref{sec:design-depth}; Table~\ref{tab:train} gives the run metrics.

\paragraph{Both stages.}
$\mu$ and the sink directions stay fixed. The training forward pass differs from inference in two ways (on \glm, three). It truncates the prompt-final states by an exact fp64 eigendecomposition instead of a subspace iteration, and it runs the continuation with the exact recurrence, without the compression every $W$ steps. On \glm it compresses only the emitted layers' prompt-final states, whereas inference compresses every Kimi Delta Attention state. Apart from the checkpoint's own quantization, only the prompt code is quantized (NVFP4, exact values in bf16), in training and in every exam. Both stages use AdamW~\cite{loshchilov2019decoupled} with $\beta=(0.9,0.95)$, $\epsilon=10^{-8}$, no weight decay and gradient clipping at 1.0.

\begin{appendixwide}{table}
\centering\footnotesize
\setlength{\tabcolsep}{4pt}
\caption{Hyperparameters of the released sets (\super: stage 1 only). Component parameters: every exported tensor, including the fixed $\mu$ and sink buffers and rows without gradient (Mamba-2 gate rows: 88M on \lightning, 134.2M on \super; indexer query rows: 6.6M on each \model set). Peak LR: stage 1 / stage 2; second stage-2 rounds about $5\times10^{-6}$. Answer-start windows: the share of the 4K / 16K windows that start at the beginning of an answer.}
\label{tab:training-details}
\begin{tabular}{@{}lrr@{\;/\;}lrccl@{}}
\toprule
 & & \multicolumn{2}{c}{} & & \multicolumn{2}{c}{answer-start (\%)} & \\
\cmidrule(lr){6-7}
Set & \begin{tabular}[b]{@{}r@{}}Component\\parameters (M)\end{tabular} & \multicolumn{2}{c}{Peak LR} & \begin{tabular}[b]{@{}r@{}}16K-window\\steps (\%)\end{tabular} & stage 1 & stage 2 & \begin{tabular}[b]{@{}l@{}}Auxiliary\\alignment\end{tabular} \\
\midrule
\model{} bf16 & 619.0 & $5\times10^{-5}$ & $10^{-5}$ & 30 & 50/30 & 70/60 & -- \\
\modelnv{} & 619.0 & $5\times10^{-5}$ & $10^{-5}$ & 30 & 50/30 & 70/60 & -- \\
\lightning{} & 235.8 & $5\times10^{-5}$ & $10^{-5}$ & 40 & 50/40 & 70/60 & states \\
\kimi{} & 113.8 & $5\times10^{-5}$ & $10^{-5}$ & 30 & 50/30 & 70/60 & -- \\
\glm{} & 708.7 & $5\times10^{-5}$ & $10^{-5}$ & 30 & 50/30 & 70/60 & states, index keys \\
\super{} & 311.3 & $5\times10^{-5}$ & -- & 40 & 50/40 & -- & states \\
\bottomrule
\end{tabular}

\end{appendixwide}
\begin{appendixwide}{table}
\centering\footnotesize
\caption{Data recipe of both stages; windows are 4K/16K tokens. $^\dagger$: 13-gram filtered when built against GPQA, MMLU, MATH-500, AIME and IFEval (stage~2 also LiveCodeBench and IFBench). Table~\ref{tab:training-data} gives each set's shares.}
\label{tab:training-recipe}
\begin{tabular}{@{}cl>{\raggedright\arraybackslash}p{108mm}@{}}
\toprule
Stage & Stream & Sources \\
\midrule
1 & public text & FineWeb-Edu~\cite{fineweb2024}, FineMath~\cite{smollm2_2025}; synthetic retrieval facts in web documents; long web documents \\
1 & reasoning/chat$^\dagger$ & OpenMathReasoning~\cite{moshkov2025aimo2}, OpenScienceReasoning-2~\cite{nvidia2025osr2}, OpenCodeReasoning~\cite{ahmad2025opencodereasoning}, OpenR1-Math~\cite{openr1}, OpenThoughts3~\cite{guha2026openthoughts} and tulu-3-sft-mixture~\cite{lambert2025tulu3} \\
1 & teacher replies$^\dagger$ & Frozen-model responses to public code, mathematics, science and chat prompts; long \lightning and \super responses \\
\addlinespace
2 & paired generation$^\dagger$ & Public prompts (code and science reused from stage 1) answered by the teacher and the current components; forward KL on both, reverse KL on the components' replies \\
2 & retained streams & Stage-1 web, retrieval and public-corpus text$^\dagger$ and stage-1 teacher replies \\
\bottomrule
\end{tabular}

\end{appendixwide}
\begin{appendixwide}{table}
\centering\scriptsize\setlength{\tabcolsep}{4pt}
\caption{The training data of every set. Stage~1: share of the tokens per data group; stage~2: rounds, prompts per round and the domain weights of the paired generations in the first round. $^\dagger$: 13-gram filtered when built against GPQA, MMLU, MATH-500, AIME and IFEval (stage~2 also LiveCodeBench and IFBench).}
\label{tab:training-data}
\begin{tabular}{@{}lrrrrrr@{}}
\toprule
 & \shortstack[r]{Qwen3.8-\\Flash-Next\\bf16} & \shortstack[r]{Qwen3.8-\\Flash-Next\\NVFP4} & \shortstack[r]{Nemotron-3.5-\\Lightning} & \shortstack[r]{Kimi-\\Linear} & \shortstack[r]{GLM-5.3-\\Flash} & \shortstack[r]{Nemotron-3.5-\\Super} \\
\midrule
\multicolumn{7}{@{}l}{\emph{stage 1, share of $6\times10^8$ tokens (\%)}} \\
\quad web text and retrieval documents & 14.1 & 14.1 & 9.9 & 15.6 & 15.6 & 9.9 \\
\quad long web documents & 5.5 & 5.5 & 4.2 & 7.8 & 7.8 & 4.2 \\
\quad reasoning and chat corpora$^\dagger$ & 21.9 & 21.9 & 17.6 & 39.3 & 19.0 & 17.6 \\
\quad frozen model's replies$^\dagger$ & 58.5 & 58.5 & 43.3 & 37.2 & 57.6 & 43.3 \\
\quad frozen model's long replies$^\dagger$ & -- & -- & 25.0 & -- & -- & 25.0 \\
\addlinespace
\multicolumn{7}{@{}l}{\emph{stage 2}} \\
\quad rounds & 2 & 1 & 2 & 1 & 2 & 1 \\
\quad prompts per round$^\dagger$ & 12{,}957 & 12{,}957 & 12{,}965 & 12{,}954 & 12{,}958 & 12{,}957 \\
\quad mathematics weight (\%) & 26 & 27 & 19 & 25 & 24 & 32 \\
\quad science weight (\%) & 23 & 23 & 15 & 18 & 18 & 33 \\
\quad code weight (\%) & 19 & 18 & 14 & 29 & 30 & 11 \\
\quad chat weight (\%) & 14 & 15 & 18 & 11 & 16 & 14 \\
\quad constraint following weight (\%) & 18 & 17 & 33 & 18 & 12 & 10 \\
\bottomrule
\end{tabular}

\end{appendixwide}

\paragraph{Stage 1.}
Stage 1 minimizes the forward next-token KL, the KL of the first 64 continuation positions (the first answer tokens on windows anchored at an answer start) and $0.1\times$ cross-entropy, on $6\times10^8$ tokens with a 50-step warm-up and cosine decay to 2\% of the peak (Tables~\ref{tab:training-details} and~\ref{tab:training-recipe}). Teacher replies supply \shareOwnFN\% of the trained tokens on both \model sets, \shareOwnLN\% on \lightning, \shareOwnKM\% on \kimi, \shareOwnGL\% on \glm and \shareOwnSU\% on \super.

\paragraph{Stage 2.}
A KL on the teacher's text misses the departures that arise when the components condition on their own tokens. Each stage-2 round therefore pairs the teacher's and the current set's replies to public mathematics, chat and output-constraint prompts unseen in stage 1 and to code and science prompts reused from it. The forward KL applies to both arms' replies, the reverse KL to the set's own, which get no cross-entropy; the other windows keep the $0.1\times$ cross-entropy. The visible answer and the 20\% of positions with the highest teacher entropy count twice (weights normalized per window), and the KL of the first 256 continuation positions is added. Paired replies supply 80\% of the tokens and retained stage-1 streams 20\% (on \super, whose round adds 32K-token windows of retained streams, about a third and two thirds), with domain weights set by the measured divergence of the set's own replies. Each round reads $1.5\times10^8$ tokens (on \super $2.1\times10^8$, two fifths of them in 32K-token windows) with a short warm-up and a cosine learning-rate schedule. Stage 2 ran two rounds on bf16 \model, \lightning and \glm and one on \modelnv, \kimi and \super. Every model releases its stage-2 set except \super, whose stage-1 set is released.

\paragraph{Exam data.}
No exam is used as training data. Stage-2 prompts and replies and the retained public-text streams pass a 13-gram filter against GPQA, MMLU, MATH-500, AIME, IFEval, LiveCodeBench and IFBench~\cite{pyatkin2025generalizing}, and stage 2's constraint prompts exclude IFEval's 25 instruction types.

\begin{appendixwide}{table}
\centering\footnotesize
\setlength{\tabcolsep}{6pt}
\caption{Run metrics of the training stages from their histories (\super releases its stage-1 set). Nodes have four GPUs; stage 1 starts from the closed-form initialization and each stage-2 round from its predecessor. Held-out: 32 windows of 4K tokens, scored after token 2048 at the run's training rank.}
\label{tab:train}
\nopagebreak[4]
\begin{tabular}{@{}lrrr@{\,$\times$\,}lrr@{}}
\toprule
 & & & \multicolumn{2}{c}{} & \multicolumn{2}{c@{}}{held-out KL} \\
\cmidrule(l){6-7}
Run & steps & tokens & \multicolumn{2}{c}{nodes $\times$ min} & step 0 & final \\
\midrule
\multicolumn{7}{@{}l}{\emph{\model{} bf16}} \\
\quad stage 1 & 881 & 600M & 16 & 159.7 & 0.0315 & 0.0176 \\
\quad stage 2, round 1 & 221 & 151M & 16 & 58.2 & 0.0174 & 0.0172 \\
\quad stage 2, round 2 & 221 & 151M & 16 & 52.5 & 0.0173 & 0.0171 \\
\addlinespace
\multicolumn{7}{@{}l}{\emph{\modelnv{}}} \\
\quad stage 1 & 881 & 600M & 16 & 208.1 & 0.0587 & 0.0423 \\
\quad stage 2 & 221 & 151M & 16 & 73.7 & 0.0424 & 0.0423 \\
\addlinespace
\multicolumn{7}{@{}l}{\emph{\lightning{}}} \\
\quad stage 1 & 1635 & 600M & 8 & 97.6 & 0.0615 & 0.0085 \\
\quad stage 2, round 1 & 409 & 150M & 8 & 25.7 & 0.0085 & 0.0083 \\
\quad stage 2, round 2 & 409 & 150M & 8 & 25.8 & 0.0083 & 0.0082 \\
\addlinespace
\multicolumn{7}{@{}l}{\emph{\kimi{}}} \\
\quad stage 1 & 1761 & 600M & 8 & 57.6 & 0.0176 & 0.0050 \\
\quad stage 2 & 441 & 150M & 8 & 17.0 & 0.0051 & 0.0049 \\
\addlinespace
\multicolumn{7}{@{}l}{\emph{\glm{}}} \\
\quad stage 1 & 881 & 600M & 16 & 89.5 & 0.0145 & 0.0064 \\
\quad stage 2, round 1 & 221 & 151M & 16 & 24.9 & 0.0064 & 0.0061 \\
\quad stage 2, round 2 & 221 & 151M & 16 & 24.3 & 0.0061 & 0.0061 \\
\addlinespace
\multicolumn{7}{@{}l}{\emph{\super{}}} \\
\quad stage 1 & 1635 & 600M & 8 & 130.5 & 0.0174 & 0.0064 \\
\quad stage 2 & 409 & 214M & 8 & 35.2 & 0.0063 & 0.0060 \\
\bottomrule
\end{tabular}
\end{appendixwide}

\paragraph{Emitted-state alignment.}
\label{app:recipe}
\lightning, \glm and \super add a term on the emitted prompt-final states: their relative squared error against the frozen model's states, on \lightning and \super weighted by how much the layer later reads of an error. Its weight is set at the start of each run to 0.2 times the answer-window KL over the term's own value. \glm adds a like term for the indexer keys and pooling scores that its emitters write, since the hard top-2048 selection gives them no KL gradient. Table~\ref{tab:staterr} gives the emitted states' errors against the frozen model's own states on 48 held-out windows split at mid-window, averaged over the emitted recurrent layers and taken before the stored form (which changes each by at most 0.03); the sets it compares also differ in code or anchor depth.
\begin{appendixwide}{table}
\centering\scriptsize
\setlength{\tabcolsep}{4pt}
\caption{Training moves the emitted prompt-final states towards the frozen model's: the released sets' pooled error is 0.20--0.52, against 0.34--1.29 for the untrained shallow prefill. Frobenius: $\|S_{\mathrm{emit}}-S\|_F^2/\|S\|_F^2$, pooled over a layer's heads (above 1: farther from the frozen model's state than a zero state). State term: trained with the emitted-state alignment term. Read-out: the error as the layer reads it over the next 64 teacher-forced tokens, relative to the state's own read-out. Floor: the frozen model's own state in the stored form.}
\label{tab:staterr}
\nopagebreak[4]
\begin{tabular}{@{}lcrrr@{}}
\toprule
 & & \multicolumn{2}{c}{error} & floor \\
\cmidrule(lr){3-4}\cmidrule(l){5-5}
Set & \begin{tabular}[b]{@{}c@{}}state\\term\end{tabular} & \begin{tabular}[b]{@{}r@{}}Frobenius,\\pooled\end{tabular} & \begin{tabular}[b]{@{}r@{}}read-out,\\64 tokens\end{tabular} & \begin{tabular}[b]{@{}r@{}}pooled /\\read-out\end{tabular} \\
\midrule
\multicolumn{5}{@{}l}{\emph{\model{} bf16}} \\
\quad untrained shallow prefill & -- & 0.615 & 0.076 & 0.021 / 0.001 \\
\quad released set & no & 0.492 & 0.047 &  \\
\addlinespace
\multicolumn{5}{@{}l}{\emph{\modelnv{}}} \\
\quad untrained shallow prefill & -- & 0.616 & 0.075 & 0.021 / 0.001 \\
\quad released set & no & 0.520 & 0.055 &  \\
\addlinespace
\multicolumn{5}{@{}l}{\emph{\lightning{}}} \\
\quad untrained shallow prefill & -- & 1.289 & 0.334 & 0.001 / 0.001 \\
\quad stage-1 set trained with the KL alone (code $1024{+}128$, not released) & no & 0.497 & 0.137 &  \\
\quad released set & yes & 0.294 & 0.100 &  \\
\addlinespace
\multicolumn{5}{@{}l}{\emph{\kimi{}}} \\
\quad untrained shallow prefill & -- & 0.466 & 0.204 & 0.004 / 0.002 \\
\quad released set & no & 0.199 & 0.091 &  \\
\addlinespace
\multicolumn{5}{@{}l}{\emph{\glm{}}} \\
\quad untrained shallow prefill at $k=31$ & -- & 0.639 & 0.141 & 0.024 / 0.005 \\
\quad preliminary set at $k=27$, code $5632{+}256$ & no & 0.576 & 0.110 &  \\
\quad preliminary set at $k=27$, no code & no & 0.628 & 0.122 &  \\
\quad preliminary set at $k=31$, no code (the released configuration) & yes & 0.501 & 0.090 &  \\
\quad released set (16 windows) & yes & 0.399 & 0.074 &  \\
\addlinespace
\multicolumn{5}{@{}l}{\emph{\super{}}} \\
\quad untrained shallow prefill & -- & 0.339 & 0.348 & 0.000 / 0.000 \\
\quad released set & yes & 0.261 & 0.071 &  \\
\bottomrule
\end{tabular}
\end{appendixwide}

\paragraph{Served divergence.}
\label{app:training-served}
Each set answers the model's stage-2 evaluation prompts (public prompts, not exam items) with one seed per prompt, and the frozen model scores every reply token. We report this divergence in $10^{-3}$ nats per token above the frozen model's floor, the divergence of its own replies rescored by prefill. Stage 2 lowers it on \modelnv from \heldNVVThreeAll{} to \heldNVVThreeFiveAll{} (\heldNVPrompts{} prompts) and on \kimi from \heldKMVThreeAll{} to \heldKMVThreeFiveAll{} (\heldKMPrompts{} prompts), while it changes the teacher-forced held-out KL of every set it trains by less than 0.0006 nats per token (Table~\ref{tab:train}).
\FloatBarrier

\section{Evaluation Configuration and Results}
\label{app:eval}
This appendix gives the exam protocol of \S\ref{sec:eval} (Table~\ref{tab:exam-tasks}), the answer flips and reply lengths, a reading of every cell at least one standard error below the original model, and the stage-1 sets.
\begin{appendixcolumn}{table}
\centering\footnotesize
\caption{Task configurations. Sampled exams are paired by question and request seed; RULER is greedy at each length. The official grader gives the result.}
\label{tab:exam-tasks}
\begin{tabular}{@{}lrr>{\raggedright\arraybackslash}p{\ifreport 98mm\fi}@{}}
\toprule
Task & Items & Repetitions & Grading and configuration \\
\midrule
GPQA-Diamond & 198 & 8 & simple-evals~\cite{openai2024simpleevals} prompt, extraction and option permutation \\
AIME 2024+25 & 60 & 8 & prompt asks for a boxed answer; integer gold answers \\
MATH-500 & 500 & 8 & prompt asks for a boxed answer; last boxed answer, Math-Verify~\cite{kydlicek2025mathverify} \\
MMLU & 14,042 & 1 & test split, simple-evals prompt and extraction \\
IFEval & 541 & 8 & prompt-level strict \\
LiveCodeBench v6 & 454 & 8 & problems of 2024-08 to 2025-04; official prompt, extraction and public/private tests \\
HumanEval+ & 164 & 8 & EvalPlus 0.4.0.dev44 grader on the v0.1.10 data, base and plus tests \\
LongBench v2 & 503 & 4 & official zero-shot template and extraction; contexts longer than 120{,}000 tokens (base tokenizer) are middle-truncated \\
RULER & 1,300 & 1 & 13 subtasks at 32K/64K/128K; at most 1024 answer tokens \\
\bottomrule
\end{tabular}
\end{appendixcolumn}
\paragraph{Server and sampling.}
Both phases of a request run in one engine instance without a prefix cache. \model and \modelnv sample with reasoning effort xhigh, temperature 1.0, top-$p$ 0.95 and top-$k$ 20; \lightning and \super with thinking on, temperature 1.0 and top-$p$ 0.95; \kimi, which has no thinking mode, with temperature 1.0 and top-$p$ 1.0; \glm with reasoning effort max, temperature 1.0 and top-$p$ 0.95. RULER runs with thinking off. The generation budget is $\min(229{,}376,\,262{,}144-\text{prompt}-8)$ tokens, with no forced close.

\paragraph{Pairing and significance.}
Scores and standard errors are paired over questions after averaging each item's repetitions; counts use unrounded values. No sample is dropped; under process sharding, MMLU questions that occur twice can get different seeds (at most 0.4\% of samples). A reply whose thinking never closes scores as no answer; one cut in its visible part is graded as written. If every true difference were zero and the standard errors exact, about 1.5 of the \examCellsTotal{} cells would lie more than two standard errors below the original model; \examBelowTwoSEN{} do. Some \duet-arm runs that hit a job time limit were resumed, which regenerates the in-flight replies; on LiveCodeBench this touches 7.7\%, 29.5\% and 3.2\% of the replies of the bf16 and NVFP4 \model sets and \kimi.

\paragraph{Answer flips and reply lengths.}
\label{app:eval-flips}
A flip is a question whose correctness differs between two runs~\cite{dutta2024accuracy}. The flips use the eight repetitions of AIME~2024, AIME~2025, GPQA-Diamond, MATH-500, IFEval and LiveCodeBench for the six sets (\flipCells{} cells). Each cell averages over consecutive pairs $(r,r{+}1)$, comparing the set's repetition $r{+}1$ with the original model's repetition $r$, and the original model's repetitions $r{+}1$ and $r$, so that both comparisons pair different sampling seeds. A set's answer and the original model's differ in correctness \flipSeedPct\% of the time, and two repetitions of the original model \flipOrigPct\%. Reply lengths pair each reply of a set with the original model's reply to the same prompt, repetition and seed, over \lenCells{} cells (GPQA-Diamond, MMLU, LiveCodeBench and the job that runs AIME, MATH-500 and IFEval together; on \super also HumanEval+ and LongBench~v2) and \lenPairs{} reply pairs. The sets write \lenRatio{} times the original model's output tokens; \lenShorter{} cells are shorter by more than two standard errors and \lenLonger{} longer. Fewer replies reach the generation limit (\lenCappedSeed{} against \lenCappedOrig) or end with their thinking unfinished (\lenThinkSeed{} against \lenThinkOrig).

\paragraph{Cells below the original model.}
\label{app:eval-cells}
We read the samples of each of the \examBelowOneSEN{} cells at least one standard error below the original model; a regrade applies to both arms, and the official scores remain the result. Grading accounts for part of the difference in five cells (paired, in points): IFEval on bf16 \model, \examFNIFEvalPair{} officially and \examFNReadIFPair{} with lm-eval's language detector run on lower-cased text, since it does not label fully upper-case English as English, and three other checker quirks fixed; MATH-500 on bf16 \model and \modelnv, \examFNMATHPairTwo{} and \examNVMATHPairTwo{} officially and \examFNReadMathPair{} and \examNVMATHNormPairTwo{} with equivalent notations (on bf16 \model also correct unboxed values) credited; LiveCodeBench on \lightning, \examLNLCBPair{} officially and \fairLNLCBWall{} with wall-clock failures counted as passed; and RULER 64K on \lightning, \examLNRulerSixtyFourPair{} officially and \fairLNRulerQA{} counting every word of a reference. In the other cells, the official grades stand.

\paragraph{Cells above the original model.}
On RULER, the original \model checkpoints sometimes end their turn right after the answer prefix, an empty reply that the official grader scores as wrong; the \duet sets do so less often, which accounts for their higher RULER scores at 32K and 64K.

\paragraph{Stage-1 sets.}
Table~\ref{tab:exam} pairs the stage-1 sets of \modelnv, \lightning and \kimi with the original model on the same questions and seeds; these stage-1 sets did not run HumanEval+ and LongBench~v2.
\begin{appendixwide}{table}
\centering\footnotesize
\setlength{\tabcolsep}{3pt}
\caption{Stage 1 against stage 2: of the 27 released $-$ stage-1 differences, only IFEval on \modelnv{} exceeds two standard errors (on unrounded values). Score, paired: the stage-1 set's score and its paired difference to the original model, mean $\pm$ standard error; released $-$ stage 1: the paired difference between the two sets. Table~\ref{tab:exam-main} gives the original model's and the released sets' scores.}
\label{tab:exam}
\nopagebreak[4]
\begin{tabular*}{\textwidth}{@{\extracolsep{\fill}}lrrrrrrrrr@{}}
\toprule
 & \multicolumn{3}{c}{\modelnv} & \multicolumn{3}{c}{\lightning} & \multicolumn{3}{c}{\kimi} \\
\cmidrule(lr){2-4}\cmidrule(lr){5-7}\cmidrule(l){8-10}
 & \multicolumn{2}{c}{stage-1 set} & released & \multicolumn{2}{c}{stage-1 set} & released & \multicolumn{2}{c}{stage-1 set} & released \\
Exam (questions) & score & paired & $-$ stage 1 & score & paired & $-$ stage 1 & score & paired & $-$ stage 1 \\
\midrule
GPQA-Diamond (198) & 91.5 & $-0.8\pm0.6$ & $-0.3\pm0.6$ & 75.1 & $-1.2\pm1.1$ & $-0.1\pm0.9$ & 67.4 & $-0.2\pm1.2$ & $+0.6\pm1.3$ \\
AIME 2024+25 (60) & 99.6 & $0.0\pm0.4$ & $0.0\pm0.4$ & 87.1 & $-2.5\pm1.4$ & $+1.7\pm1.0$ & 66.0 & $-1.5\pm2.4$ & $+0.8\pm2.6$ \\
MATH-500 (500) & 99.1 & $-0.1\pm0.1$ & $-0.2\pm0.1$ & 98.4 & $+0.2\pm0.2$ & $-0.1\pm0.2$ & 92.5 & $-0.6\pm0.4$ & $-0.1\pm0.4$ \\
MMLU (14{,}042) & 92.4 & $0.0\pm0.2$ & $+0.2\pm0.2$ & 88.0 & $-0.1\pm0.2$ & $-0.1\pm0.2$ & 77.9 & $+0.2\pm0.4$ & $+0.4\pm0.4$ \\
IFEval (541) & 90.7 & $-0.9\pm0.3$ & $+0.9\pm0.3$ & 91.5 & $-0.1\pm0.4$ & $+0.5\pm0.4$ & 71.4 & $-1.8\pm0.6$ & $+0.3\pm0.6$ \\
LiveCodeBench v6 (454) & 89.6 & $+0.6\pm0.4$ & $-0.7\pm0.4$ & 78.0 & $-0.2\pm0.5$ & $-0.4\pm0.5$ & 43.7 & $-0.7\pm0.8$ & $-0.2\pm0.6$ \\
RULER 32K (1{,}300) & 95.8 & $+1.1\pm0.5$ & $+0.3\pm0.2$ & 96.4 & $-0.1\pm0.2$ & $0.0\pm0.1$ & 95.1 & $+0.1\pm0.1$ & $-0.1\pm0.1$ \\
RULER 64K (1{,}300) & 95.3 & $+0.4\pm0.5$ & $+0.2\pm0.2$ & 96.0 & $-0.4\pm0.2$ & $0.0\pm0.2$ & 94.1 & $+0.1\pm0.2$ & $-0.2\pm0.1$ \\
RULER 128K (1{,}300) & 94.4 & $-0.9\pm0.4$ & $+0.2\pm0.3$ & 95.2 & $-0.6\pm0.3$ & $+0.4\pm0.2$ & 94.0 & $-0.5\pm0.2$ & $+0.1\pm0.2$ \\
\bottomrule
\end{tabular*}
\end{appendixwide}

\FloatBarrier

\section{Performance Details}
\label{app:perf-model}\label{app:served}
This appendix gives the byte and FLOP account behind Table~\ref{\mainMemoryLabel} and \S\ref{sec:perf} and the stored forms of a cached prefix.
%
\begin{appendixwide}{table}
\centering\footnotesize
\caption{Analytical byte and FLOP account. $b_s$ and $b_c$ are the state and cache precisions in bytes; counts use checkpoint tensor shapes. For Mamba-2, use the orientation of Appendix~\ref{app:theory}: $d_k=d_B=128$ and $d_v=d_x=64$. Runtime is measured separately.}
\label{tab:performance-definitions}
\begin{tabular}{@{}l>{\raggedright\arraybackslash}p{125mm}@{}}
\toprule
Quantity & Account \\
\midrule
Dense state per head & $b_sd_kd_v$ \\
Shallow state & $b_s[d_k+(d_k+d_v)(R+w)]$ at $w$ steps after a compression, $0\le w\le W$: the sink's coefficient and the factors; the mean direction is shared by all requests and not counted \\
Resident request & $M_{\mathrm r}(T)$ (\S\ref{sec:setting}): the state bytes plus $T$ times the attention-cache bytes per token, plus the short-convolution windows (which Table~\ref{\mainMemoryLabel} leaves out) \\
Decode traffic & cache reads, state reads/writes and the weights touched by the batch; sparse attention reads selected tokens and indexer keys \\
Expected routed experts & $n_e[1-(1-n_a/n_e)^B]$ per layer for the $B$ tokens of a step, with $n_e$ routed experts of which each token activates $n_a$: exact for independent tokens under uniform routing, an upper bound when independent tokens share a skewed routing distribution \\
Prompt computation & two FLOPs per parameter used in layers $0..k{-}1$, code and emitters; no per-token output-head term \\
Compression & once per $W$ decode steps (per generated token in Table~\ref{\mainMemoryLabel}) and once at prompt completion (a per-request cost, not in Table~\ref{\mainMemoryLabel}): fp32 products and reconstruction, fp64 orthonormalization and small eigendecomposition \\
Excluded terms & n-gram lookup table traffic, page tables, padding, activations, attention scores and recurrence scans \\
\bottomrule
\end{tabular}
\end{appendixwide}
\paragraph{Weights against caches and states.}
\label{app:perf-crossover}
With an 8K-token prompt and 1K output tokens, the weights that a decode step reads outweigh the original model's cache and state traffic up to a batch of 512 requests on bf16 \model, \kimi, \glm and \super and of 256 on \modelnv and \lightning; at the next power of two, 1024 and 512, the caches and states overtake the weights (Table~\ref{tab:performance-definitions}).
\paragraph{Prefix storage.}
The original form keeps dense states and every layer's cache. Form (i) of Table~\ref{tab:prefix-forms} keeps every layer's cache (the deep ones as the emitters wrote them) and the shallow states, so a reuse needs no computation. Form (ii) keeps the caches and shallow states of layers $0..k{-}1$ and the prompt code; a reuse decodes the code and reruns the emitters. A code alone is not a further form: its reuse would rerun layers $0..k{-}1$. The smaller form per model and cache precision is a byte account, not an implemented cache policy.
\begin{appendixwide}{table}
\centering\footnotesize
\setlength{\tabcolsep}{5pt}
\caption{With bf16 attention caches, form (ii) of a cached prefix takes up to 22\% fewer bytes than form (i), and its reuse costs 0.22--1.07 GFLOP per prompt token; with fp8 caches the two differ by at most 3\%, except on \lightning, where (i) is 12\% smaller. A cached prefix of 80K tokens in each storage form: bytes per prompt token (the prefix's states and convolution windows spread over its tokens) at bf16 and fp8 attention caches, and the GFLOP per prompt token that its reuse costs. (i)~every layer's cache, the deep ones as the emitters wrote them, and the shallow states; (ii)~the caches and states of layers $0..k{-}1$ and the code, from which the emitters rewrite the deep layers. $^{\dagger}$The smaller of (i) and (ii) in bytes. The NVFP4 set of \model has the bytes and FLOPs of the bf16 set; \glm has no code, so (i) is its only form, and its fp8 figures halve the latent only, its indexer keys being fp8 already.}
\label{tab:prefix-forms}
\begin{tabular}{@{}llrrr@{\hspace{6pt}}}
\toprule
 & & Orig. & (i) & (ii) \\
\midrule
\model & bf16 caches (B) & 26,753 & 25,727 & 20,420\rlap{$^{\dagger}$} \\
 & fp8 caches (B) & 14,081 & 13,055 & 13,028\rlap{$^{\dagger}$} \\
 & reuse (GFLOP) & 0 & 0 & 1.07 \\
\addlinespace
\lightning & bf16 caches (B) & 6,743 & 6,384 & 5,793\rlap{$^{\dagger}$} \\
 & fp8 caches (B) & 3,671 & 3,312\rlap{$^{\dagger}$} & 3,745 \\
 & reuse (GFLOP) & 0 & 0 & 0.46 \\
\addlinespace
\kimi & bf16 caches (B) & 8,594 & 8,342 & 6,500\rlap{$^{\dagger}$} \\
 & fp8 caches (B) & 4,562 & 4,310 & 4,196\rlap{$^{\dagger}$} \\
 & reuse (GFLOP) & 0 & 0 & 0.22 \\
\addlinespace
\glm & bf16 caches (B) & 13,429 & 12,572\rlap{$^{\dagger}$} & -- \\
 & fp8 caches (B) & 7,797 & 6,940\rlap{$^{\dagger}$} & -- \\
 & reuse (GFLOP) & 0 & 0 & -- \\
\addlinespace
\super & bf16 caches (B) & 10,270 & 9,022 & 8,591\rlap{$^{\dagger}$} \\
 & fp8 caches (B) & 6,174 & 4,926\rlap{$^{\dagger}$} & 5,007 \\
 & reuse (GFLOP) & 0 & 0 & 0.62 \\
\bottomrule
\end{tabular}
\end{appendixwide}

\FloatBarrier

\end{document}